%% file: Main.tex
\documentclass[manuscript,screen]{acmart}

\usepackage{graphicx}
\usepackage{nameref}
\usepackage{xcolor}
\usepackage{hyperref}
\usepackage{tikz}
\usetikzlibrary{trees}
\usetikzlibrary{trees,decorations.pathreplacing}
\usepackage{booktabs}
\usepackage{tabularx}
\usepackage{array}
\usepackage{makecell}
\usepackage{multirow}
\usepackage[acronym]{glossaries}
\glsdisablehyper
\input{Data/Acronyms}

\setcopyright{acmlicensed}
\acmJournal{CSUR}
\acmVolume{1}   
\acmNumber{1}    
\acmArticle{1}   
\acmYear{2025}      
\acmMonth{10}        
\acmDOI{XXXXXXX.XXXXXXX}  
\copyrightyear{2025} 

\begin{document}

\title[Mitigating Hallucination in Large Language Models: A Capability-Oriented Survey on RAG, Reasoning, and Agentic Systems]{Mitigating Hallucination in Large Language Models: A Capability-Oriented Survey on RAG, Reasoning, and Agentic Systems}

\author{Yihan Li}
\affiliation{%
  \institution{Electronic Information School, Wuhan University}
  \city{Wuhan}
  \country{China}
}
\affiliation{%
  \institution{Dublin City University}
  \city{Dublin}
  \country{Ireland}
}
\email{yihan.li2@mail.dcu.ie}

\author{Xiyuan Fu}
\affiliation{%
  \institution{School of Public Health, Wuhan University}
  \city{Wuhan}
  \country{China}
}
\email{fuxiyuan@whu.edu.cn}

\author{Ghanshyam Verma}
\affiliation{%
  \institution{Insight Centre for Data Analytics, University of Galway}
  \city{Galway}
  \country{Ireland}
}
\email{ghanshyam.verma@insight-centre.org}

\author{Paul Buitelaar}
\affiliation{%
 \institution{Insight Centre for Data Analytics, University of Galway}
 \city{Galway}
 \country{Ireland}
 }
\email{paul.buitelaar@insight-centre.org}

\author{Mingming Liu}
\affiliation{%
  \institution{Insight Centre for Data Analytics, Dublin City University}
  \city{Dublin}
  \country{Ireland}
}
\email{mingming.liu@dcu.ie}
\authornote{Corresponding author.}

\begin{abstract}
\input{Data/Abstract}
\end{abstract}



\maketitle

\input{Data/1.Introduction}

\input{Data/2.Related_Surveys}

\input{Data/3.Background_and_Definitions}

\input{Data/4.Knowledge-based_Hallucination_and_RAG}

\input{Data/5.Logic-Based_Hallucination_and_Reasoning}

\input{Data/6.Synergy_and_Hybrid_Approaches}

\input{Data/7.Benchmarks}

\input{Data/8.Conclusion_and_Challenges}

\begin{acks}
This work has emanated from research conducted with the financial support of Taighde Éireann — Research Ireland under Grant number \textit{SFI/12/RC/2289\_P2} and co-funded by the European Regional Development Fund in collaboration with the Research Ireland Insight Centre for Data Analytics at Dublin City University.
\end{acks}

\bibliographystyle{ACM-Reference-Format}

\bibliography{References-Pro}

\appendix

\end{document}

%% file: Data/Acronyms.tex
\newacronym{ml}{ML}{Machine Learning} 
\newacronym{ai}{AI}{Artificial Intelligence} \newacronym{gnns}{GNNs}{Graph Neural Networks} \newacronym{cv}{CV}{Computer vision} 
\newacronym{nlp}{NLP}{Natural Language Processing} \newacronym{gcn}{GCN}{Graph Convolution Networks} \newacronym{tl}{TL}{Transfer learning} \newacronym{htl}{HTL}{Heterogeneous transfer learning} 
\newacronym{ssl}{SSL}{Self-supervised learning} \newacronym{cl}{CL}{Contrastive learning} \newacronym{rag}{RAG}{Retrieval-Augmented Generation} 
\newacronym{cot}{CoT}{Chain-of-Thought} \newacronym{llms}{LLMs}{Large Language Models} \newacronym{rlhf}{RLHF}{Reinforcement Learning from Human Feedback} 
\newacronym{aigc}{AIGC}{AI-Generated Content}

%% file: Data/Abstract.tex
Hallucination remains one of the key obstacles to the reliable deployment of large language models (LLMs). Although various mitigation approaches have been proposed, existing studies often analyze different technical paradigms independently, lacking a unified perspective to understand the underlying mechanisms of different approaches and their correspondence with different types of hallucinations.
This survey adopts a capability enhancement perspective to systematically examine hallucination mitigation approaches, focusing on Retrieval-Augmented Generation (RAG), reasoning enhancement, and their integration within agentic systems. Based on their primary mitigation mechanisms, we categorize hallucinations into knowledge-based hallucinations and logic-based hallucinations, analyze how RAG and reasoning enhancement methods respectively improve knowledge acquisition and reasoning reliability, and further discuss the integration mechanisms of retrieval and reasoning capabilities in Agentic Systems for mitigating composite hallucinations. By considering the applicability, mitigation mechanisms, and limitations of different approaches, this survey establishes a unified analytical framework connecting hallucination types, key capability dimensions, and technical paradigms.

%% file: Data/1.Introduction.tex
\section{Introduction}
Hallucination~\cite{huangSurveyHallucinationLarge2025, zhangSirensSongAI2023, jiSurveyHallucinationNatural2023}, defined as the generation of content that appears plausible but is inconsistent with real-world facts or user instructions, has emerged as one of the most critical challenges in the deployment of \gls{llms}~\cite{sahaYouBelieveYour2025}. In high-stakes applications such as medical diagnosis, legal analysis, and scientific research, even minor factual or logical errors can lead to severe consequences~\cite{kimMedicalHallucinationsFoundation2025}, including the spread of misinformation and the erosion of public trust in AI systems~\cite{sahaYouBelieveYour2025}. Existing studies have shown that hallucinations remain widespread across models of all scales~\cite{banerjeeLLMsWillAlways2024}, from trillion-parameter systems such as ChatGPT-4~\cite{openaiGPT4TechnicalReport2024}, and DeepSeek-R1~\cite{liu2024deepseek} to smaller models like LLaMA 7B~\cite{touvronLLaMAOpenEfficient2023} and Mistral~\cite{Jiang2023Mistral7B}, underscoring the urgency and necessity of addressing this issue. Therefore, how to effectively mitigate hallucinations and improve the reliability of LLM generation has become an important research problem.

Existing hallucination mitigation methods~\cite{huangSurveyHallucinationLarge2025, zhangSirensSongAI2023} span multiple stages of the LLM lifecycle, including data processing and alignment strategies~\cite{leeFactualityEnhancedLanguage2023} during pre-training, fine-tuning~\cite{tangMitigatingHallucinatedTranslations2025} and reinforcement learning methods based on human feedback and preferences~\cite{ouyang2022training}, and inference-stage intervention techniques~\cite{chuangDoLaDecodingContrasting2024} such as \gls{rag}~\cite{fanSurveyRAGMeeting2024} and \gls{cot}~\cite{chuNavigateEnigmaticLabyrinth2024} reasoning. Among these approaches, inference-time capability enhancement methods have gradually become an important research direction in hallucination mitigation, as they enhance the knowledge acquisition and reasoning capabilities of models by introducing external knowledge, explicit reasoning processes, and dynamic interaction mechanisms.

Most existing surveys~\cite{zhangSirensSongAI2023, huangSurveyHallucinationLarge2025, jiSurveyHallucinationNatural2023} on hallucinations in LLMs focus on their causes, taxonomies, and lifecycle-based mitigation stages, or analyze isolated strategies at the data, model, or decoding levels. Other surveys investigate specific technical directions, including RAG, reasoning enhancement, or agentic systems that integrate these capabilities~\cite{fanSurveyRAGMeeting2024, peng2025graph, sharma2025retrieval, chengEmpoweringLLMsLogical2025, li2025survey}.
Although these studies provide comprehensive summaries of existing techniques, they mainly focus on the development and evolution of individual technical paradigms. A unified analysis from the perspective of capability enhancement remains relatively limited, particularly regarding how LLMs improve knowledge acquisition, reasoning reliability, and capability coordination, as well as how these enhancement mechanisms correspond to different types of hallucinations.

To fill this gap, this survey adopts a capability enhancement perspective and focuses on three representative technical directions in current hallucination mitigation research: RAG, reasoning enhancement, and their integration within agentic systems. Instead of treating capability enhancement as an independent taxonomy, we employ it as an analytical perspective to investigate two fundamental dimensions affecting LLM generation quality: knowledge acquisition and reliable reasoning. Based on this perspective, we analyze how RAG enhances the ability of LLMs to acquire and utilize external knowledge, how reasoning enhancement improves reasoning reliability in complex tasks, and how agentic systems further coordinate retrieval, reasoning and feedback processes to achieve synergistic capability enhancement. Through this analysis, we establish a unified framework connecting hallucination types, capability enhancement mechanisms, and corresponding technical approaches.

Based on this framework, we further categorize LLM hallucinations into knowledge-based hallucinations and logic-based hallucinations from the perspective of mitigation mechanisms, as illustrated in Fig.~\ref{fig:Hallucination Case}. Knowledge-based hallucinations mainly refer to cases where generated content lacks factual support, conflicts with real-world knowledge, or fails to properly utilize relevant information. In contrast, logic-based hallucinations refer to unreliable generations caused by invalid inference, inconsistent reasoning chains, or inappropriate conclusions under given contextual constraints. It should be noted that this categorization is not intended to cover all possible causes and manifestations of hallucinations. Instead, it serves as a high-level organizational framework for capability-oriented analysis, linking hallucination phenomena with two fundamental capability dimensions: knowledge acquisition and reasoning reliability.
\input{Pictures/hallucination_case}

\input{Tables/Table_1}

\textit{Our contributions.}
With hallucination mitigation as the central objective and capability enhancement as the analytical criterion:
\begin{itemize}
    \item We propose a capability enhancement-oriented analytical framework for hallucination mitigation, establishing the relationship among hallucination types, key capability dimensions, and technical approaches. Provide a unified perspective for comparing and analyzing different mitigation methods.
    \item We systematically review the development of RAG as a pathway for enhancing knowledge acquisition capabilities and analyze how different RAG paradigms improve knowledge acquisition and mitigate knowledge-based hallucinations, together with their corresponding limitations.
    \item We systematically review three major categories of reasoning enhancement approaches from the perspective of improving reasoning reliability. We analyze their mechanisms for enhancing logical derivation and constraint-based reasoning capabilities, as well as their applicability and limitations in mitigating logic-based hallucinations.
    \item We further investigate the integration of retrieval and reasoning capabilities within agentic systems, analyzing how representative paradigms, including reasoning-enhanced retrieval, multi-agent collaboration, and autonomous systems with iterative feedback loops, mitigate composite hallucinations through capability coordination.
\end{itemize}

\textit{Organization of this Survey.} The organization and hierarchical relationship of this survey, covering RAG, reasoning enhancement, and their integration within agentic systems, are illustrated in Fig.~\ref{tab:model_tree}. Specifically, Section~\hyperref[sec:section2]{\ref*{sec:section2}} provides a review and comparison of related surveys. Section~\hyperref[sec:section3]{\ref*{sec:section3}} presents the background and key concepts underpinning this survey. Sections~\hyperref[sec:section4]{\ref*{sec:section4}} and~\hyperref[sec:section5]{\ref*{sec:section5}} respectively review RAG and reasoning enhancement approaches, and analyze their roles in mitigating knowledge-based and logic-based hallucinations from the capability enhancement perspective.
Section~\hyperref[sec:section6]{\ref*{sec:section6}} discusses agentic system paradigms that integrate retrieval and reasoning capabilities, and analyzes how different integration strategies mitigate composite hallucinations through capability coordination. Section~\hyperref[sec:section7]{\ref*{sec:section7}} summarizes existing evaluation benchmarks by hallucination type. Finally, Section~\hyperref[sec:section8]{\ref*{sec:section8}} concludes with challenges and future directions.  

%% file: Pictures/hallucination_case.tex
\begin{figure}[t!]  %
\centering
\includegraphics[width=1\linewidth, trim=50 1200 50 1150, clip]{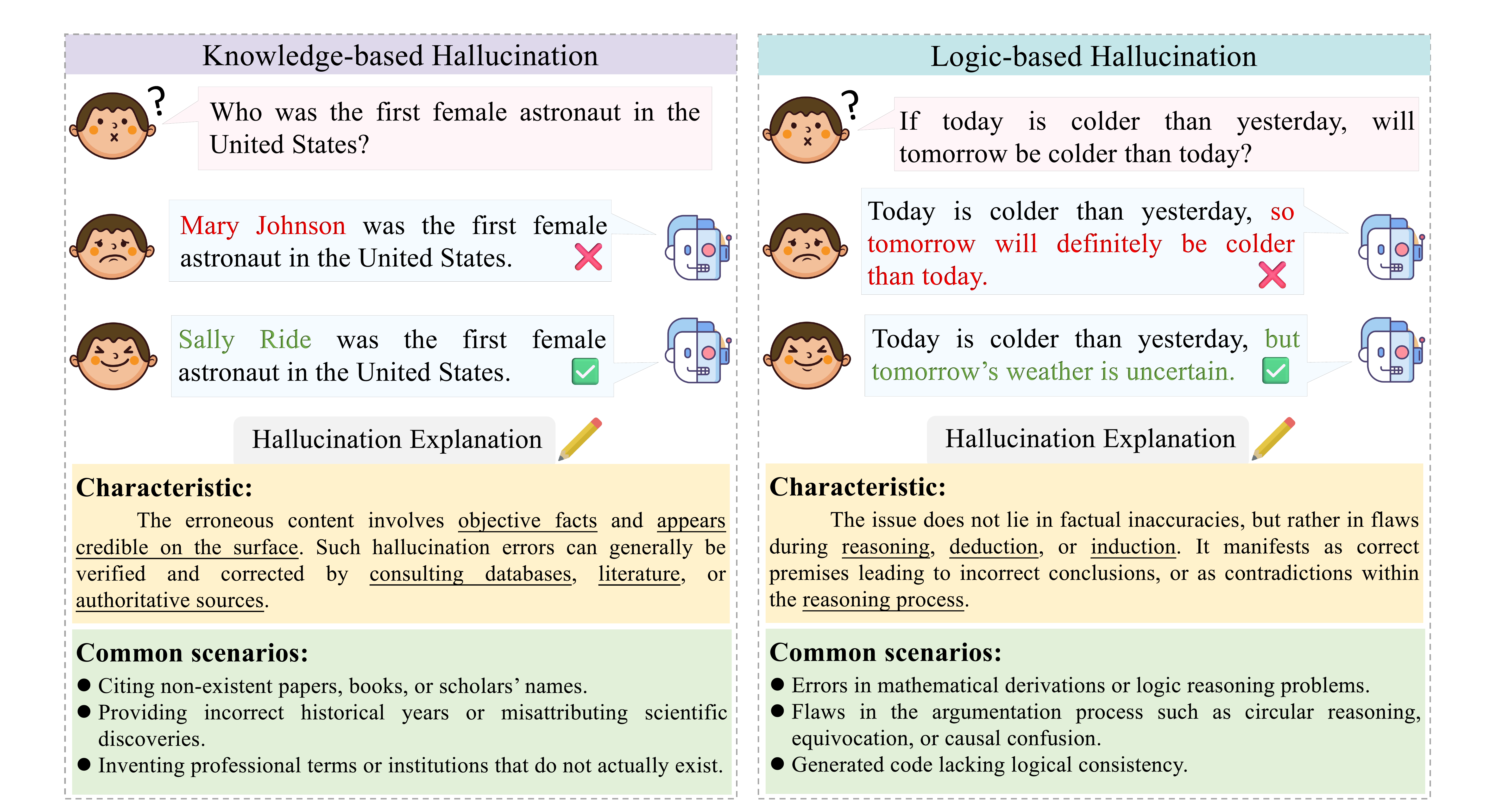}
\caption{Two types of hallucinations in LLM responses}
\label{fig:Hallucination Case}
\Description{Illustrative examples of knowledge-based and logic-based hallucinations in large language models, showing how factual inaccuracies and flawed reasoning processes can lead to misleading or incorrect outputs, along with explanations of their typical characteristics and scenarios.}
\end{figure}

%% file: Tables/Table_1.tex
\begin{figure}[!t]
\centering

\begin{tikzpicture}[
  grow=right,       
  level distance=3cm,   
  sibling distance=3cm, 
  edge from parent path={(\tikzparentnode.east) -- ++(3mm,0) |- (\tikzchildnode.west)}, 
  every node/.style={draw, rectangle, rounded corners, align=left}
]
\node[fill=red!10, line width=0.8pt, draw=red!60](root){\rotatebox{90}{\textbf{Hallucination Mitigation Methods}}}
  child[sibling distance=4cm, level distance=3cm, line width=0.8pt]{node[text width=3.5cm, fill=yellow!15, align=center, draw=yellow]{\textbf{Agentic System} §{\color{purple}\hyperref[sec:section6]{6}}}
    child[sibling distance=1.5cm, level distance=4.5cm, , fill=yellow!10]{node[text width=3.4cm, font=\small, fill=yellow!10, align=center]{Iterative Agentic Systems §{\color{purple}\hyperref[sec:6.2.3]{6.2.3}}}
      child[level distance=4.5cm]{node[text width=4.5cm, font=\small, fill=yellow!5]{Self-Refine~\cite{madaan2023self}, Reflexion~\cite{shinn2023reflexion}, Reflective RAG~\cite{wu2026reflective}}} 
    }
    child[sibling distance=1.5cm, level distance=4.5cm, , fill=yellow!10]{node[text width=3.4cm, font=\small, fill=yellow!10, align=center]{Multi-Agent Systems §{\color{purple}\hyperref[sec:6.2.2]{6.2.2}}}
      child[level distance=4.5cm]{node[text width=4.5cm, font=\small, fill=yellow!5]{AutoGen~\cite{wu2024autogen}, MetaGPT~\cite{hong2024metagpt}, MA-RAG~\cite{nguyenMARAGMultiAgentRetrievalAugmented2025}}}  
    }
    child[sibling distance=1.4cm, level distance=4.5cm, , fill=yellow!10]{node[text width=3.4cm, font=\small, fill=yellow!10, align=center]{Reasoning-Augmented Retrieval Systems §{\color{purple}\hyperref[sec:6.2.1]{6.2.1}}}
      child[level distance=4.5cm]{node[text width=4.5cm, font=\small, fill=yellow!5]{ReAct~\cite{yaoReActSynergizingReasoning2023}, IRCoT~\cite{trivediInterleavingRetrievalChainofThought2023}, SELF-RAG~\cite{asaiSELFRAGLEARNINGRETRIEVE2023}}} 
    }
  }
  child[sibling distance=2cm, level distance=3cm, line width=0.8pt]{node[text width=3.5cm, fill=green!15, align=center, draw=green!60]{\textbf{Reasoning} §{\color{purple}\hyperref[sec:section5]{5}}}
    child[sibling distance=1.6cm, level distance=4.5cm]{node[text width=3.4cm, font=\small, fill=green!10, align=center]{Symbolic Reasoning §{\color{purple}\hyperref[sec:SR]{5.3}}}
      child[sibling distance=2cm]{node[text width=4.5cm, font=\small, fill=green!5]{Faithful CoT~\cite{lyuFaithfulChainofThoughtReasoning2023}, SymbCoT~\cite{xuFaithfulLogicalReasoning2024}, ChatLogic~\cite{wangChatLogicIntegratingLogic2024}, Logic-LM~\cite{panLogicLMEmpoweringLarge2023}}}
    }
    child[sibling distance=1.7cm, level distance=4.5cm]{node[text width=3.4cm, font=\small, fill=green!10, align=center]{Tool-Augmented Reasoning §{\color{purple}\hyperref[sec:TAR]{5.2}}}
      child[sibling distance=2cm, level distance=4.5cm]{node[text width=4.5cm, font=\small, fill=green!5]{ReAct~\cite{yaoReActSynergizingReasoning2023}, Toolformer~\cite{schickToolformerLanguageModels2023}, ToolFiVe~\cite{luToolFiVeEnhancingToolAugmented2025}, PoT~\cite{chenProgramThoughtsPrompting2023}, ANSWERED~\cite{panANSWEREDAdaptiveToolAugmented2024}}}
    }
    child[sibling distance=1.7cm, level distance=4.5cm]{node[text width=3.4cm, font=\small, fill=green!10, align=center]{Chain-of-Thought §{\color{purple}\hyperref[sec:CoT]{5.1}}}
      child[sibling distance=2cm, level distance=4.5cm]{node[text width=4.5cm, font=\small, fill=green!5]{REFINER~\cite{paulREFINERReasoningFeedback2024}, Dynasor-CoT~\cite{fuREASONINGSELFDOUBTMORE2025}, LONGREPS~\cite{zhuChainofThoughtMattersImproving2025}, AdaCoT~\cite{louAdaCoTParetoOptimalAdaptive2025}, D-CoT~\cite{wangDynamicChainofThoughtAdaptive2025}}}
    }
  }
  child[sibling distance=4.5cm, level distance=3cm, line width=0.8pt]{node[text width=3.5cm, fill=blue!15, align=center, draw=blue!60]{\textbf{RAG} §{\color{purple}\hyperref[sec:section4]{4}}}
    child[sibling distance=1.4cm, level distance=4.5cm]{node[text width=3.4cm, font=\small, fill=blue!10, align=center]{Broad Retrieval §{\color{purple}\hyperref[sec:broad retrieval]{4.3}}}
      child[sibling distance=1.6cm, level distance=4.5cm]{node[text width=4.5cm, font=\small, fill=blue!5]{DeepResearcher~\cite{zhengDeepResearcherScalingDeep2025}, WebGLM~\cite{liuWebGLMEfficientWebEnhanced2023}, VisRAG~\cite{yuVisRAGVisionbasedRetrievalaugmented2025}, FoRAG~\cite{caiFoRAGFactualityoptimizedRetrieval2024}, MuRAG~\cite{chenMuRAGMultimodalRetrievalAugmented2022}}}
    }
    child[sibling distance=1.5cm, level distance=4.5cm]{node[text width=3.4cm, font=\small, fill=blue!10, align=center]{Precise Retrieval §{\color{purple}\hyperref[sec:precise retrieval]{4.2}}}
      child[sibling distance=2cm, level distance=4.5cm]{node[text width=4.5cm, font=\small, fill=blue!5]{GraphRAG~\cite{edgeLocalGlobalGraph2025}, GNN-RAG~\cite{mavromatisGNNRAGGraphNeural2024}, HyPA-RAG~\cite{kalraHyPARAGHybridParameter2025}, Hybrid-RAG~\cite{sarmahHybridRAGIntegratingKnowledge2024}}}
    }
    child[sibling distance=1.3cm, level distance=4.5cm]{node[text width=3.4cm, font=\small, fill=blue!10, align=center]{Pipeline Level§{\color{purple}\hyperref[sec:rag pipeline]{4.1}}}
      child[sibling distance=2cm, level distance=4.5cm]{node[text width=4.5cm, font=\small, fill=blue!5]{RQ-RAG~\cite{chanRQRAGLearningRefine2024}, Blended RAG~\cite{sawarkarBlendedRAGImproving2024}, SELF-RAG~\cite{asaiSELFRAGLEARNINGRETRIEVE2023}, LongLLMLingua~\cite{jiangLongLLMLinguaAcceleratingEnhancing2024}}}
    }
  };
\end{tikzpicture}
\caption{Capability-Oriented Organization of Hallucination Mitigation Techniques in LLMs}
\label{tab:model_tree}
\Description{A tree diagram illustrating the main content and structure of this paper, including RAG, Reasoning, Agentic System, along with their related key techniques and representative models.}
\end{figure}

%% file: Data/2.Related_Surveys.tex
\section{Related Surveys}
\label{sec:section2}
To ensure comprehensive and representative coverage, this survey encompasses publications from major academic venues and publishers, including IEEE, ACM, Springer, Elsevier, and leading conference organizations such as AAAI, ACL, NeurIPS, and ICLR, as well as recent works available on arXiv and other reputable preprint platforms (the proportion of preprints is below 15\%). The inclusion of arXiv papers is particularly important given the rapid pace of developments in emerging topics such as hallucination mitigation, RAG, reasoning, agentic systems, and related benchmarks. Most of these preprints consist of high-impact studies released within the past year, ensuring that this survey captures the most recent advances and evolving trends in the field.
Literature selection combines keyword-based retrieval with thematic relevance assessment, emphasizing representative studies that address hallucination control mechanisms, mitigation strategies, and real-world applications.

Within this scope, prior surveys related to this study can be broadly categorized into two main streams: hallucination-oriented surveys and technique-oriented surveys.
The former focus on understanding and mitigating hallucinations in LLMs, while the latter explore retrieval-based or reasoning-based frameworks that are conceptually related to hallucination control.

\textbf{Surveys of Hallucination.}
A substantial body of prior work, including surveys by Zhang et al.~\cite{zhangSirensSongAI2023}, Huang et al.~\cite{huangSurveyHallucinationLarge2025}, and Ji et al.~\cite{jiSurveyHallucinationNatural2023}, has systematically summarized the causes, mitigation strategies, and evaluation methods of hallucinations in LLMs, providing valuable taxonomies for understanding this phenomenon. But they do not examine the impact of reasoning ability on hallucination mitigation and lack evaluation of the interactions among different hallucination mitigation methods.
Bai et al.~\cite{baiHallucinationMultimodalLarge2025} and Sahoo et al.~\cite{sahooComprehensiveSurveyHallucination2024} further extended this discussion to multi-modal hallucination, analyzing how misinformation propagates across modalities. Lin et al.~\cite{linLLMbasedAgentsSuffer2025} conducted a systematic survey on LLM-based agents, summarizing the types, causes, and detection mechanisms of hallucinations within LLM Agents.

Existing hallucination-oriented surveys have established a relatively systematic research landscape covering hallucination taxonomy, formation mechanisms, detection and evaluation, and mitigation techniques. However, they generally examine mitigation methods as relatively independent technical solutions, with limited attention to the interactions among retrieval, reasoning, and agentic mechanisms. In particular, the relationships between different capability deficiencies and hallucination types remain insufficiently explored.

\textbf{Surveys of RAG, Reasoning and Agents.}
Beyond surveys specifically focusing on LLM hallucinations, existing studies have also systematically reviewed RAG, reasoning enhancement, and agentic systems. Fan et al.~\cite{fanSurveyRAGMeeting2024} and Sharma et al.~\cite{sharma2025retrieval} provided comprehensive overviews of the fundamental architectures, key techniques, and development trends of RAG, while Peng et al.~\cite{peng2025graph} and Zhang et al.~\cite{zhang2025survey} further focused on GraphRAG and conducted an in-depth discussion of the construction, retrieval, and generation mechanisms of graph-structured knowledge. Zhang et al.~\cite{zhangHallucinationMitigationRetrievalAugmented2025} specifically investigated hallucination issues within RAG systems. Meanwhile, Ke et al.~\cite{ke2025survey}, Chu et al.~\cite{chuNavigateEnigmaticLabyrinth2024}, and Cheng et al.~\cite{chengEmpoweringLLMsLogical2025} systematically reviewed LLM reasoning techniques, covering the organization of reasoning processes, capability enhancement methods, and evaluation approaches.
As retrieval and reasoning techniques have become increasingly integrated, recent surveys have begun to examine their coordination within agentic systems. Li et al.~\cite{li2025survey} summarized RAG-Reasoning systems from the perspective of the bidirectional integration between retrieval and reasoning. Liang et al.~\cite{liang2025reasoning} primarily investigated Agentic RAG frameworks in which reasoning processes dynamically control retrieval behaviors. Singh et al.,~\cite{singh2025agentic} from the broader perspective of agentic RAG systems, systematically analyzed the organization of functional components such as planning, retrieval, tool use, and feedback. Wei et al.~\cite{wei2026agentic} provide a comprehensive survey on agentic reasoning in LLMs, reviewing the evolution of reasoning paradigms and the development of agentic frameworks.

These technique-oriented surveys provide important foundations for understanding RAG, reasoning enhancement, and agentic systems. However, they primarily focus on technical architectures and mechanisms, without systematically analyzing how different approaches enhance LLM capabilities and mitigate various hallucination types. Different from existing surveys, this work adopts a capability enhancement perspective and establishes connections among hallucination types, capability deficiencies, and mitigation techniques by jointly analyzing RAG, reasoning enhancement, and their integration within agentic systems.

%% file: Data/3.Background_and_Definitions.tex
\section{Background}
\label{sec:section3}

\input{Data/3.Background_and_Related_Surveys/3.1LLMs}

\input{Data/3.Background_and_Related_Surveys/3.2}

\input{Data/3.Background_and_Related_Surveys/3.3}


%% file: Data/3.Background_and_Related_Surveys/3.1LLMs.tex
\subsection{Large Language Models}
Large Language models (LLMs)~\cite{kumar2024llmsurvey, zhaoSurveyLargeLanguage2025} have emerged as one of the most prominent paradigms in deep learning. Initially known for their exceptional performance in \gls{nlp} tasks, \gls{llms} have since demonstrated a remarkable capacity to generalize beyond language, influencing domains such as code generation~\cite{ouyangLLMBoxChocolates2023}, reasoning~\cite{weiChainofThoughtPromptingElicits2023}, and even scientific discovery~\cite{liEmpoweringMoleculeDiscovery2024}.

Technically, LLMs are typically built on the Transformer decoder architecture, generating text through autoregressive next-token prediction over large-scale corpora~\cite{vaswaniAttentionAllYou}. This prediction mechanism, grounded in statistical correlation, endows the model with powerful generative and generalization capabilities. However, it also introduces inherent randomness and uncertainty~\cite{ouyangLLMBoxChocolates2023}, which are widely regarded as major sources of hallucination~\cite{banerjeeLLMsWillAlways2024}. 

Meanwhile, the emergent abilities~\cite{weiEmergentAbilitiesLarge2022} and internal reasoning mechanisms of \gls{llms} remain insufficiently understood~\cite{mirzadehGSMSymbolicUnderstandingLimitations2025}. The opacity of their generation process makes hallucination particularly challenging to control and highlights the need for systematic research on hallucination mitigation.

%% file: Data/3.Background_and_Related_Surveys/3.2.tex
\subsection{Hallucination Definition and Mitigation Methods}
\textit{Definition.}
Hallucination is not a new problem in \gls{nlp}. It is first formally mentioned in abstractive document summarization~\cite{maynezFaithfulnessFactualityAbstractive2020}, where the models are highly prone to hallucinate content that is unfaithful to the input document~\cite{jiSurveyHallucinationNatural2023}. In the realm of \gls{llms}, the hallucination problem typically refers to the phenomenon in which a model generates content that appears plausible, but is factually incorrect, logically inconsistent, or misaligned with the intent of the user~\cite{huangSurveyHallucinationLarge2025, zhangSirensSongAI2023}. This phenomenon seriously affects the reliability and controllability of \gls{llms}, especially in high-stakes domains such as medicine, law, and finance, where high accuracy, consistency, verifiability, and traceability are critical~\cite{kimMedicalHallucinationsFoundation2025, dahlLargeLegalFictions2024}.

With the rapid spread of LLMs, hallucinations have become a key obstacle for the real-world use and trustworthiness of \gls{llms}~\cite{tangMitigatingHallucinatedTranslations2025, sahaYouBelieveYour2025}. They undermine user trust in the outputs and constrain the potential of LLMs to provide expert knowledge and perform deep reasoning. In addition, as LLMs become larger, more widely used, and more open-ended, hallucinations are showing new patterns: they are more diverse, harder to detect, and can change over time~\cite{sahaYouBelieveYour2025, farquhar2024detecting}.

\textit{Mitigation.}
Research has explored how to detect and reduce hallucinations to better understand and manage this problem~\cite{farquhar2024detecting, mishraFinegrainedHallucinationDetection2024}. 
Existing hallucination mitigation methods, apart from RAG and reasoning enhancement, include cleaning and inspecting pre-training data, improving pre-training~\cite{leeFactualityEnhancedLanguage2023} and fine-tuning~\cite{tangMitigatingHallucinatedTranslations2025}, using reinforcement learning~\cite{ouyang2022training} for better alignment, and applying post-hoc verification or rewriting~\cite{huangSurveyHallucinationLarge2025, zhangSirensSongAI2023}. The purpose of these methods is to improve the quality of both training and generation in existing training approaches and models. Other methods aim to reduce the occurrence of hallucinations by optimizing the model architecture~\cite{liBatGPTBidirectionalAutoregessive2023, liuExposingAttentionGlitches2023} or the answer generation process~\cite{farquhar2024detecting, liContrastiveDecodingOpenended2023, azariaInternalStateLLM2023, chuangDoLaDecodingContrasting2024}.

The hallucination mitigation techniques mentioned above have certain practical value, but they are not the focus of this survey. Instead, we primarily concentrate on RAG and reasoning-based approaches as well as their integration for the following reasons: (1) They mitigate hallucinations by enhancing the capabilities of LLMs, rather than suppressing them; (2) They preserve scalability and creativity, which are crucial for open-domain and exploratory tasks; (3) They demonstrate high deployability, making them adaptable to a wide range of real-world applications; (4) Their practical utility has already been validated in the market, and the combination of these two approaches holds great potential for mitigating all types of hallucinations, offering broad prospects for application.

Recent work shows that eliminating hallucinations completely is nearly impossible because they are an inherent feature of generative models~\cite{banerjeeLLMsWillAlways2024}. Interestingly, the same mechanisms that cause hallucinations also drive \gls{llms}' creativity and originality~\cite{jiangSurveyLargeLanguage2024}. Fully removing hallucinations might also suppress this creative potential. Therefore, it is widely accepted that mitigation, not complete removal, is the realistic goal.

%% file: Data/3.Background_and_Related_Surveys/3.3.tex
\subsection{RAG, Reasoning, and Agentic Systems in LLMs}
\textit{Retrieval-Augmented Generation (RAG)}, in the context of LLMs, refers to retrieving information from external knowledge sources during the inference stage to support content generation~\cite{fanSurveyRAGMeeting2024}. It has been widely adopted because it can flexibly integrate the latest or domain-specific knowledge without the need for frequent and large-scale updates to model parameters. Initially, RAG was primarily regarded as a means to fill knowledge gaps left from the pre-training stage, but subsequent research has shown that it has broader value: correcting internal model errors and biases, enabling rapid and modular knowledge updates, improving answer traceability, and enhancing contextual consistency. These extended capabilities make RAG a powerful and versatile solution for mitigating hallucinations in knowledge-intensive tasks.

\textit{Reasoning} refers to the ability of \gls{llms} to dynamically interpret complex instructions, decompose them into sub-goals, construct coherent and rigorous logical chains, and follow structured steps to accomplish tasks~\cite{chengEmpoweringLLMsLogical2025}. This capability is not merely reflected in the length of the generated content, but more importantly in the model’s capacity for structured thinking, logical planning, and adaptive decision-making in solving multi-stage problems. In recent research of LLMs, three representative forms of reasoning have emerged: Chain-of-Thought (CoT)~\cite{weiChainofThoughtPromptingElicits2023, chuNavigateEnigmaticLabyrinth2024}, which is a prompt-based method that guides models to generate intermediate reasoning steps for improved logical coherence; Tool-augmented reasoning~\cite{schickToolformerLanguageModels2023, yaoReActSynergizingReasoning2023}, which leverages external tools such as calculators, search engines, or APIs to enhance problem-solving accuracy; and Symbolic reasoning~\cite{panLogicLMEmpoweringLarge2023, wangChatLogicIntegratingLogic2024}, which transforms natural language into symbolic representations for verifiable, logic-based computation. These methods enhance the reasoning capability of LLMs from different perspectives and thus serve as key approaches for mitigating logic-based hallucinations.

\textit{Agent} is generally defined as a goal-oriented entity with a certain degree of autonomy, which can perceive environmental states, select corresponding actions through internal decision-making processes, and achieve predefined objectives through interactions with the environment~\cite{conte2012rational,martinelli2025taxonomy}. Early studies mainly focused on theoretical modeling of Agent from perspectives such as rationality, autonomy, and internal decision-making mechanisms. For example, Conte et al. emphasized rational goal-oriented behavior and highlighted that Agent makes decisions and takes actions based on cognitive states, such as beliefs, goals, and intentions~\cite{conte2012rational}. In recent years, related studies have further developed taxonomy frameworks from the perspectives of Agent capabilities and types. For instance, Martinelli systematically categorizes different forms of Agent based on levels of proactivity and types of entities~\cite{martinelli2025taxonomy}.
With the development of LLMs, Agentic Systems generally refer to system frameworks centered on LLMs that integrate mechanisms such as task planning, memory, tool use, action execution, and environmental feedback to accomplish complex tasks autonomously or semi-autonomously~\cite{li2025review}. Unlike conventional language models that directly generate responses from input prompts, Agentic Systems can organize execution processes around task objectives and dynamically adjust subsequent actions according to intermediate states.
This survey primarily focuses on a class of frameworks within the broader category of Agentic Systems that integrate external knowledge retrieval with multi-step reasoning capabilities~\cite{li2025survey, liang2025reasoning}. By coordinating retrieval and reasoning processes and incorporating planning, collaboration, or feedback mechanisms, these systems jointly enhance the model’s knowledge acquisition and logical problem-solving capabilities. They therefore provide an important technical pathway for mitigating composite hallucinations jointly caused by insufficient knowledge and reasoning failures.

%% file: Data/4.Knowledge-based_Hallucination_and_RAG.tex
\section{Knowledge-based Hallucination and RAG}
\label{sec:section4}
\input{Pictures/rag_pipeline}
Knowledge-based hallucinations arise from inaccuracies in a model’s internal knowledge or insufficient external information. Mitigation therefore focuses on supplementing the model with accurate and relevant external knowledge. Among various approaches, RAG has emerged as the most effective and widely studied framework for enhancing factual consistency and reliability in knowledge-intensive applications~\cite{mitchellFastModelEditing2022, caoEditingFactualKnowledge2021}.

This section focuses on how RAG supplements external knowledge to mitigate knowledge-based hallucinations. We first outline the overall RAG pipeline, summarizing its core mechanisms and development trends. We then divide RAG techniques into precise retrieval and broad retrieval, systematically analyzing key techniques from different application perspectives and enabling hallucination mitigation methods to better adapt to different application domains.

\input{Data/4/4.1RAG_Pipeline}

\input{Data/4/4.2Precise_Retrieval}

\input{Data/4/4.3Broad_Retrieval}

\input{Data/4/4.4}

%% file: Pictures/rag_pipeline.tex
\begin{figure*}[t!]  %
\centering
\includegraphics[width=1\linewidth, trim=200 150 270 200, clip]{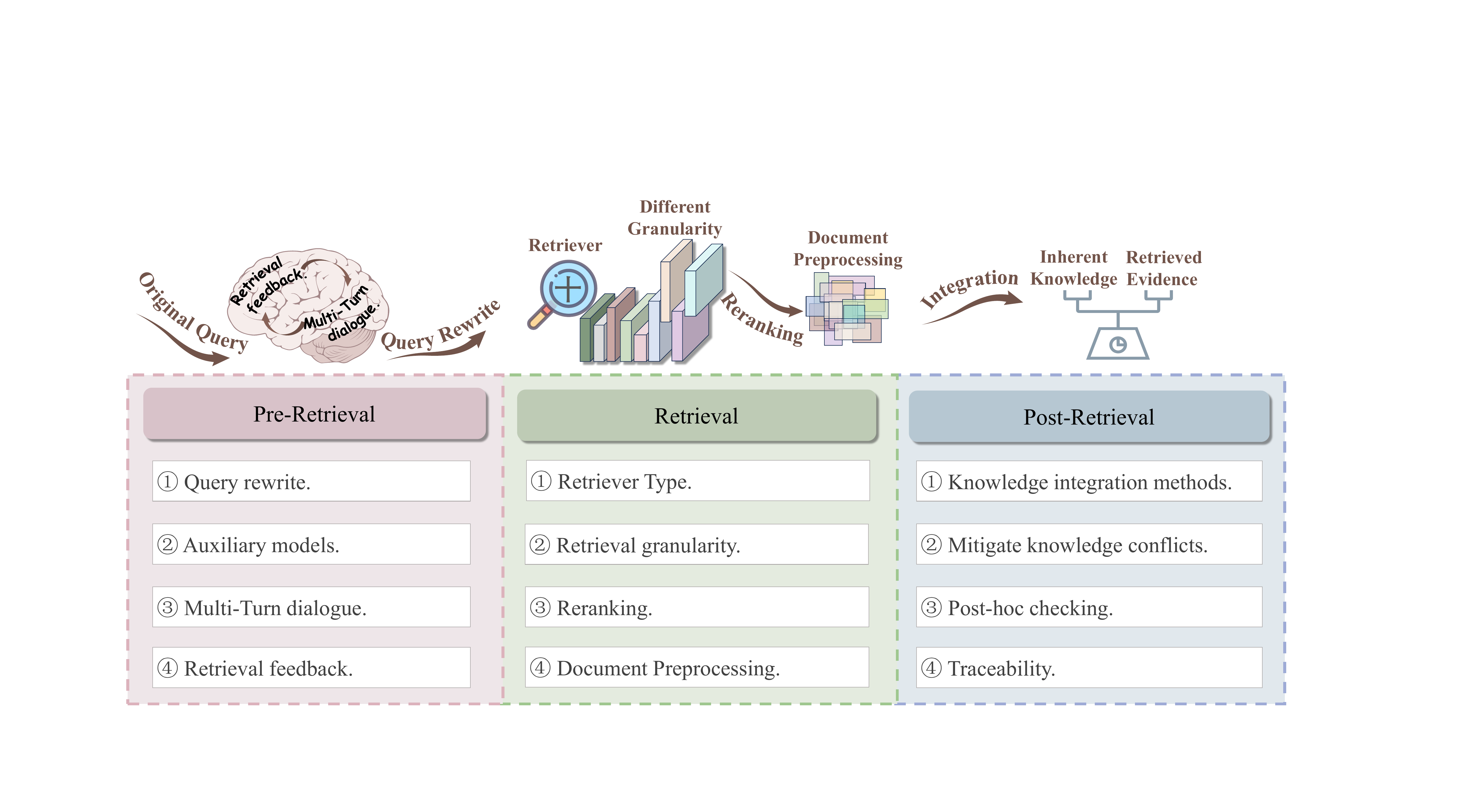}
\caption{Overview of the RAG pipeline}
\label{fig:RAG Pipeline}
\Description{A detailed illustration of the RAG pipeline. The diagram outlines pre-retrieval, retrieval, and post-retrieval stages, highlighting key techniques such as query rewriting, multi-turn dialogue, retrieval feedback, reranking, and knowledge integration.}
\end{figure*}

%% file: Data/4/4.1RAG_Pipeline.tex
\subsection{RAG Pipeline and Key Techniques}
\label{sec:rag pipeline}
Organized along the RAG retrieval pipeline, which consists of pre-retrieval, retrieval, and post-retrieval stages, this category focuses on optimizing the internal procedures of RAG systems, including query processing, retrieval strategy selection, and retrieved context processing, as illustrated in Fig.~\ref{fig:RAG Pipeline}. We summarize recent technical advances in RAG pipeline that enhance the overall effectiveness of RAG and analyze how these techniques improve knowledge retrieval and utilization, thereby mitigating hallucinations in general tasks.

\subsubsection{Pre-retrieval}
The pre-retrieval stage centers on understanding the user’s intent beyond surface-level keyword matching. In RAG, intent understanding means aligning the retriever and \gls{llms} to accurately capture information needs: what is asked? what type of content is required? and how it should be structured or sourced. A strong grasp of intent enables more targeted retrieval, improving relevance and reducing hallucinations. Four key methods to enhance intent understanding are summarized in Fig.~\ref{fig:Intent Understanding}.

\paragraph{Query rewrite}
\label{query rewrite}
Query rewriting is one of the fundamental approaches to improve retrieval-oriented intent understanding. Since user queries are often ambiguous, incomplete, or expressed differently from the knowledge required for retrieval, inaccurate query formulation may lead to irrelevant evidence retrieval and even hallucinations~\cite{watson2025there}. Therefore, rewriting methods transform the original query into a more explicit and retrieval-effective representation, improving the alignment between user intent and external knowledge. Ma et al.~\cite{maQueryRewritingRetrievalAugmented2023} proposed a query rewriting approach to bridge the gap between the input text and the knowledge required for retrieval. This rewriting process reformulates the original query into a more retrieval-effective form, improving the relevance and factual alignment of the retrieved results.
Building on this, RQ-RAG~\cite{chanRQRAGLearningRefine2024} focuses on training the model to decompose and disambiguate complex queries, improving its ability to identify complex user intent.

\input{Pictures/intent_understanding}

\paragraph{Auxiliary models}
Beyond directly modifying the query, some approaches introduce auxiliary models to infer missing information requirements before retrieval. These lightweight models serve as intermediate components that analyze the current knowledge state and identify what additional evidence is needed. For example, Tan et al.~\cite{tanSmallModelsBig2024} propose SlimPLM, which first generates preliminary answers and then uses them to identify missing knowledge for retrieval.

\paragraph{Multi-Turn dialogue}
In multi-turn conversations, user intent is often distributed across multiple dialogue turns rather than explicitly expressed in a single query\cite{qianExplicitQueryRewriting2022}. Therefore, incorporating historical context becomes important for reconstructing complete information needs. 
Liu et al.~\cite{liuLARALinguisticAdaptiveRetrievalAugmentation2024} propose LARA to leverage the dialogue history to better understand user intent and reconstruct more effective queries, resulting in more accurate retrieval. 

\paragraph{Retrieval feedback}
While the above methods mainly improve intent understanding before retrieval, recent approaches further refine retrieval decisions by leveraging feedback from retrieved results or generated content. This feedback helps identify missing information and adjust the original query or retrieval requirements. Representative models include RA-ISF~\cite{liuRAISFLearningAnswer2024} and KiRAG~\cite{fangKiRAGKnowledgeDrivenIterative2025}. RA-ISF adopts an iterative process that incorporates feedback from each retrieval step into decision-making, and decomposes sub-questions for further retrieval when necessary. KiRAG performs step-by-step iterative retrieval over the input query: after each retrieval step, it formulates a new query based on the current reasoning chain to acquire the knowledge needed for the next step.

\subsubsection{Retrieval}
After accurately identifying user intent, the process moves to the retrieval stage. This stage focuses on the retriever’s ability to locate relevant knowledge efficiently and precisely based on the query, ensuring that the retrieved information is accurate, comprehensive, and concise. It serves as a core metric in the design of retrievers within the RAG framework. We have identified and summarized several factors that affect the performance of document retrieval:

\paragraph{Retriever Type}
Currently, commonly used retrievers can be roughly divided into three categories based on their retrieval principles: sparse retrievers, dense retrievers, and hybrid (sparse + dense) retrievers. Table~\ref{tab:Retrievers} categorizes and lists retrievers in terms of their working mechanisms, representative models, and application scenarios. 
Sparse retrievers rely on keyword matching and offer strong interpretability and fast response times. However, they struggle with handling semantic variations in queries. Dense retrievers use encoders to map both queries and documents into a shared semantic space, enabling better capture of semantic relationships. 
Hybrid retrievers integrate sparse and dense methods, combining the efficiency and interpretability of sparse retrievers with the semantic strength of dense retrievers. By leveraging both lexical and semantic signals through score fusion or reranking, hybrid approaches often outperform either method alone, especially in complex open-domain and LLM-related tasks.
\input{Tables/Retrievers}
Arivazhagan et al.~\cite{arivazhaganHybridHierarchicalRetrieval2023} demonstrated that retrieval with hybrid retriever achieves superior performance in accurately mitigating hallucinations, outperforming both purely sparse and purely dense retrieval methods. Hybrid examples include Blended RAG~\cite{sawarkarBlendedRAGImproving2024}, CG-RAG~\cite{huCGRAGResearchQuestion2025} and HyPA-RAG~\cite{kalraHyPARAGHybridParameter2025}. The strong performance of these models demonstrates that hybrid retrievers represent an important direction for future development in retrieval selection.

\paragraph{Retrieval granularity}
Retrieval granularity defines the minimal content unit used for retrieval, such as document, chunk, sentence, or token. Coarse-grained retrieval offers broader context but risks introducing noise, while fine-grained retrieval improves precision at higher computational cost. To balance these trade-offs, recent methods combine multiple granularities.
MoG~\cite{zhongMixofGranularityOptimizeChunking2025} and KET-RAG~\cite{huangKETRAGCostEfficientMultiGranular2025} organize knowledge sources into multi-granularity structures, allowing the model to dynamically select appropriate granularity based on task context. Notably, KET-RAG addresses the coarse-grained limitation of traditional Graph-RAG~\cite{edgeLocalGlobalGraph2025} by capturing finer entity-level relationships for more effective retrieval.

\paragraph{Reranking}
\label{Reranking}
Document reranking aims to optimize the selection and ordering of retrieved results within the context window, often prioritizing the most relevant content at the beginning or end of the input to maximize its impact on generation~\cite{liuLostMiddleHow2023}.
Existing reranking methods can be grouped into three categories: feature-based models (e.g., BM25~\cite{robertson2009probabilistic}, TF-IDF~\cite{saltonTermweightingApproachesAutomatic1988}), learning-to-rank models~\cite{liu2009learning} (pointwise, pairwise, listwise)~\cite{burgesLearningRankUsing2005, burges2006learning}, and neural rerankers based on pre-trained LMs (e.g., BERT~\cite{devlinBERTPretrainingDeep2019}). While neural methods achieve superior semantic matching, most approaches still struggle with complex queries, cross-document reasoning, and multi-turn understanding.
In recent years, LLM-based and reinforcement learning-based reranking methods have emerged, equipping rerankers with stronger contextual understanding and reasoning capabilities. 
LLM4Ranking~\cite{liuLLM4RankingEasytouseFramework2025} proposes a unified LLM-based reranking framework that leverages both open- and closed-source models to enhance generalization and reasoning through external LLM-based semantic modeling and ranking.
Rank-R1~\cite{zhuangRankR1EnhancingReasoning2025} introduces a reinforcement learning-based reranker with task-aware reasoning abilities, capable of dynamically adjusting ranking strategies based on the query.
In RAG systems, reranking also serves as a filtering mechanism to eliminate task-irrelevant information. Advances in reranking techniques help prevent context contamination by unrelated content, thereby reducing the risk of hallucinations caused by distracting inputs~\cite{changMAINRAGMultiAgentFiltering}.

\paragraph{Document Preprocessing}
In addition to document reranking, modifying or compressing retrieved documents before generation has emerged as an important preprocessing step in RAG systems. Unlike reranking methods that mainly adjust document-level relevance, document refinement approaches further improve the usability of retrieved evidence by removing redundancy and highlighting informative content.
One line of research focuses on information compression, aiming to reduce irrelevant content and retain information most beneficial for generation. Zhou et al.~\cite{zhouTrustRAGEnhancingRobustness2025} propose TrustRAG, a robust framework that detects and masks "hallucination-inducing" expressions within retrieved segments. 
Jiang et al.~\cite{jiangLongLLMLinguaAcceleratingEnhancing2024} proposed LongLLMLingua, which significantly reduces the length of retrieved documents and increases the density of key information through a question-aware coarse-to-fine prompt compression and reordering strategy.
Another line of research performs fine-grained evidence extraction by selecting answer-relevant sentences or spans from retrieved documents. For example,
DSLR~\cite{hwangDSLRDocumentRefinement2024} uses a sentence-level selector to extract query-relevant sentences, then reorganize them into a coherent paragraph according to their order in the original text.
These methods aim to improve the information density of retrieved content, reduce input length, and thereby control computational cost, reduce redundancy, and mitigate hallucination risk results from irrelevant information.

\subsubsection{Post-retrieval}
The core task of the post-retrieval stage is to effectively integrate the retrieved external knowledge with the model’s internal knowledge, enabling the generator to provide accurate answers based on the combined information. The generated content should remain semantically aligned with the user’s intent and faithfully reflect the key information contained in the retrieved results.


\paragraph{Knowledge integration methods}
To effectively integrate external knowledge into the generation process of \gls{llms}, three mainstream strategies have been proposed~\cite{fanSurveyRAGMeeting2024}: input-level, intermediate-layer, and output-level integration. These approaches differ in coupling depth, computational cost, and effectiveness in mitigating hallucinations.
Input-level integration, which concatenates retrieved documents with the query as input, remains the most widely adopted due to its simplicity and compatibility with existing models~\cite{ramInContextRetrievalAugmentedLanguage2023}. Intermediate-layer integration introduces interaction between retrieved knowledge and internal representations, offering deeper semantic alignment and better factual consistency~\cite{wuReFusionImprovingNatural2024}. Output-level integration adjusts the model’s output distribution based on retrieval results, providing modularity but limited semantic control.
Overall, input-level integration serves as a practical baseline, while intermediate-layer methods are increasingly favored in recent studies for their superior performance in reducing hallucinations~\cite{fanSurveyRAGMeeting2024}.

\paragraph{Mitigate knowledge conflicts}
To enhance the model’s use of retrieved knowledge, prompt engineering and instruction tuning are often employed to explicitly guide retrieval-grounded reasoning. When conflicts arise between retrieved and internal knowledge, recent methods introduce reliability assessment rather than relying solely on one source.
RE-RAG~\cite{kimRERAGImprovingOpenDomain2024} assigns confidence scores to retrieved documents, while C-RAG~\cite{ranaldiElicitingCriticalReasoning2024} promotes critical reasoning through contrastive explanations. These approaches improve adaptability and robustness in handling noisy or conflicting information.

\paragraph{Post-hoc checking}
Post-hoc consistency checking has emerged as another line of research. This involves aligning the generated output with the retrieved evidence at the semantic level to evaluate whether the model genuinely grounded its response in the provided documents~\cite{asaiSELFRAGLEARNINGRETRIEVE2023}. Such analysis can be used to assess answer faithfulness and information utilization, ultimately serving as a tool for identifying potential hallucinations.

\paragraph{Traceability}
To improve the traceability of RAG-generated content, enhance the transparency of external knowledge usage, and facilitate the detection and correction of potential hallucinations, researchers have proposed a range of mechanisms aimed at making the link between output and source documents more explicit. For instance, Tariq et al.~\cite{guttikondaExplainableAIRetrievalAugmented2025} proposed an interactive interpretability mechanism for RAG outputs and source knowledge, allowing users to view how retrieved passages are used, understand the basis for generation, and query the actual source segments associated with the output without requiring technical expertise. 


%% file: Pictures/intent_understanding.tex
\begin{figure*}[t!]  %
\centering
\includegraphics[width=1\linewidth, trim=150 1200 150 1200, clip]{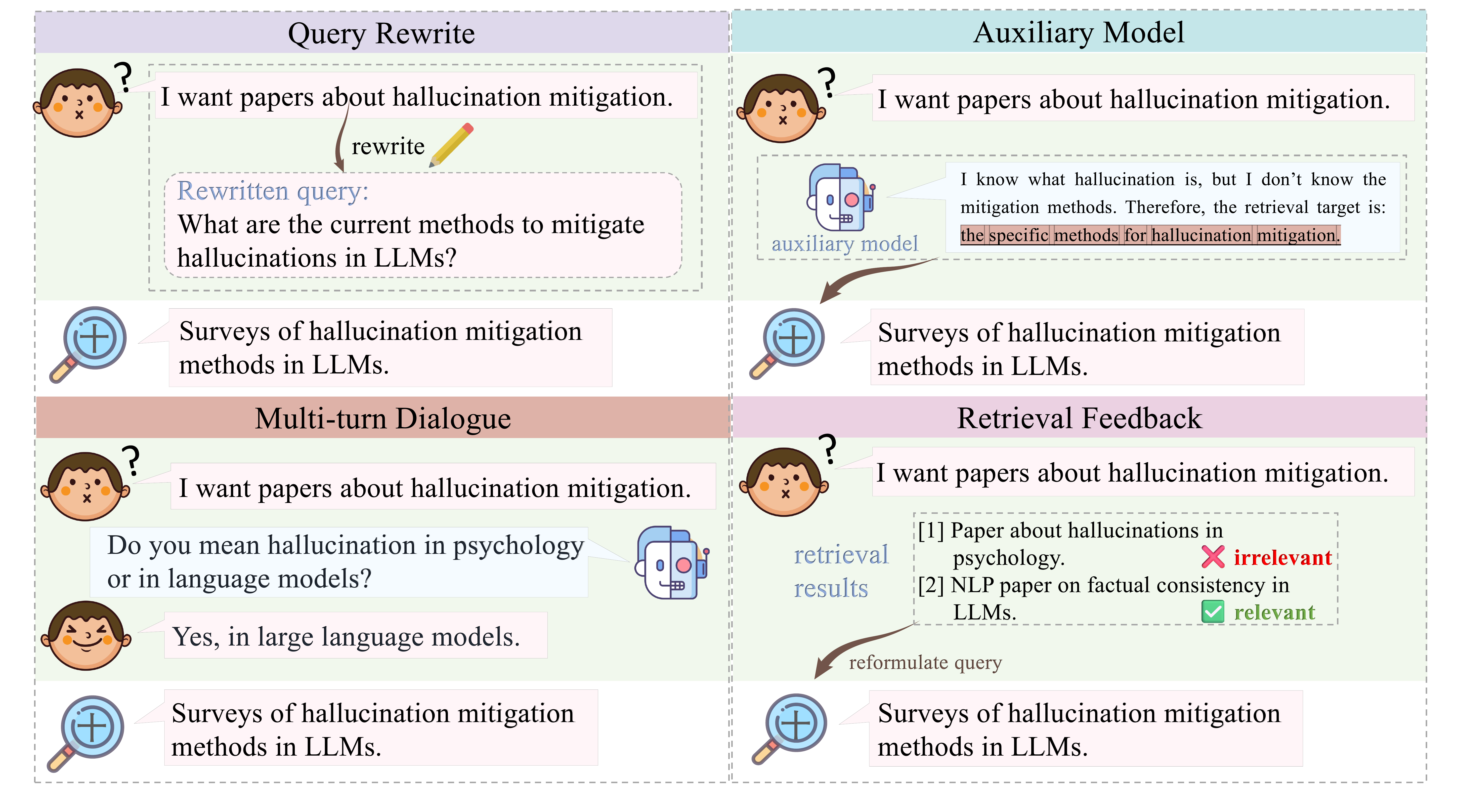}
\caption{Illustrative examples of four methods for enhancing intent understanding}
\label{fig:Intent Understanding}
\Description{An illustration of intent understanding in Retrieval-Agumented Systems, demonstrating how user intent is clarified through query rewriting, auxiliary model feedback, multi-turn dialogue and retrieval feedback to refine ambiguous queries and improve the relevance of retrieved results.}
\end{figure*}

%% file: Tables/Retrievers.tex
\begin{table}[!ht]
\caption{Mainstream Retriever Categories (Sparse vs Dense vs Hybrid)}
\label{tab:Retrievers}
\renewcommand{\arraystretch}{1.3}
\centering
\small
\begin{tabular}{@{}
  >{\centering\arraybackslash}p{0.12\textwidth}
  >{\raggedright\arraybackslash}p{0.29\textwidth}
  >{\raggedright\arraybackslash}p{0.20\textwidth}
  >{\raggedright\arraybackslash}p{0.29\textwidth}
  @{}}
\toprule
Category & Key Mechanism & Representative Models & Typical Use Cases \\
\midrule

Sparse Retrievers &
Inverted index; term matching; no deep vectors &
BM25~\cite{lewisRetrievalAugmentedGenerationKnowledgeIntensive2021}, TF-IDF~\cite{saltonTermweightingApproachesAutomatic1988}, SPLADE~\cite{formalSPLADESparseLexical2021} &
Document retrieval, first-stage recall, resource-limited settings \\

\midrule

Dense Retrievers &
Dense vector encoding; similarity via approximate nearest neighbor search &
DPR\cite{karpukhinDensePassageRetrieval2020}, Contriever\cite{izacardUnsupervisedDenseInformation2022}, ANCE\cite{xiongApproximateNearestNeighbor2020a} & Open-domain QA, semantic similarity search, cross-lingual retrieval \\

\midrule

Hybrid Retrievers &
Combine sparse and dense signals; token-level interaction at retrieval &
ColBERT\cite{khattabColBERTEfficientEffective2020} / ColBERTv2\cite{santhanamColBERTv2EffectiveEfficient2022} &
Retrieval + reranking pipelines, high-precision QA retrieval \\

\bottomrule
\end{tabular}

\end{table}

%% file: Data/4/4.2Precise_Retrieval.tex
\subsection{Precise Retrieval}
\label{sec:precise retrieval}

The development of precise retrieval, which focuses on improving retrieval accuracy through structured and explicitly organized knowledge sources, is driven by the retrieval challenges LLMs face in domain-specific question answering. Such tasks often rely on fixed, structured, and large-scale knowledge sources that contain long-tail information insufficiently covered during pretraining.
To address this problem, researchers have explored several approaches that enhance fine-grained knowledge retrieval by exploiting structured knowledge representations, including Graph-Augmented RAG, Knowledge Graph-based RAG (KG-RAG) and Hybrid RAG~\cite{peng2025graph}.

\subsubsection{Graph-Augmented RAG}
Graph-Augmented RAG is designed to address the challenges of information redundancy, lack of global context, and missing inter-document relationships in large-scale unstructured corpora~\cite{peng2025graph}.
Recent studies have explored organizing document collections into graph structures, where sentences, paragraphs, or topical units are extracted as nodes, and edges are established based on semantic similarity, contextual co-occurrence, or citation relationships.
For example, Edge et al.~\cite{edgeLocalGlobalGraph2025} proposed GraphRAG, which constructs document semantic segments into a graph and performs hierarchical summarization through community detection, significantly improving RAG’s performance on query-focused summarization (QFS) tasks over large corpora.
Hu et al.~\cite{huCGRAGResearchQuestion2025} developed CG-RAG (Citation Graph RAG), which incorporates a citation graph to enhance retrieval by enabling LLMs to perform semantic aggregation, evidence tracing, and cross-paper reasoning based on citation relationships, thereby improving both accuracy and interpretability in scientific question answering.

\input{Tables/Structured_Data}

\subsubsection{Knowledge Graph}
In application domains where knowledge is relatively stable and well-bounded, such as medicine, law, and public policy, knowledge sources can be structured and standardized through the extraction of key facts, definition of entity relationships, and construction of knowledge graphs (KG) (Table~\ref{tab:Knowledge Graph})~\cite{jiSurveyKnowledgeGraphs2021, zhongComprehensiveSurveyAutomatic2023}.
Compared with unstructured documents, KGs contain explicit fields, entity relationships, and hierarchical structures that eliminate redundancy and provide a more organized representation of information. By applying knowledge-graph-based RAG (KG-RAG), LLMs can more accurately identify and extract key information, reducing hallucinations caused by semantic ambiguity and achieving superior performance in domain-specific applications.
For example, Li et al.~\cite{liDALKDynamicCoAugmentation2024} applied KG-RAG to medical question answering on Alzheimer’s disease. Kalra et al.~\cite{kalraHyPARAGHybridParameter2025} used legal-entity and statute relations for structured retrieval in legal and policy QA.

KGs also support deeper logical reasoning and are well suited to multi-hop questions and complex contexts. GNN-RAG~\cite{mavromatisGNNRAGGraphNeural2024} scores candidate nodes and extracts paths on KG subgraphs with graph neural networks, then verbalizes these paths into the RAG context to provide structured evidence, improving multi-hop QA.
Path-based retrieval further enhances traceability and interpretability.
In addition, Sun et al.~\cite{sunThinkonGraphDeepResponsible2024} proposed the ToG (Think-on-Graph) framework, a method that performs reasoning directly on KGs. It enables LLMs to actively explore reasoning paths within the KG and produce verifiable reasoning trajectories, similar to the behavior of intelligent agents. This approach further extends the potential of KGs in enhancing deep reasoning for LLMs.


\subsubsection{Hybrid RAG}
Hybrid RAG refers to a dual-channel retrieval mechanism that integrates KG retrieval and vector-based retrieval. Originally proposed in HybridRAG~\cite{sarmahHybridRAGIntegratingKnowledge2024}, this approach performs both retrieval types simultaneously and concatenates the retrieved text chunks with the corresponding KG subgraphs. It aims to combine the structural and interpretable strengths of KG retrieval with the semantic coverage and representational richness of vector retrieval. Through this integration, the model maintains broad semantic understanding while improving factual consistency and traceability, leading to higher retrieval accuracy~\cite{maThinkonGraph20Deep2025}.
The core of Hybrid RAG lies in the design of its retrieval fusion mechanism. Compared with traditional single-stage fusion approaches, Ma et al.~\cite{maThinkonGraph20Deep2025} proposed an iterative retrieval fusion framework (Think-on-Graph 2.0), which alternately performs graph-based and text-based retrieval to progressively expand knowledge coverage and semantic depth, thereby achieving more comprehensive integration and higher reasoning reliability. Xu et al.~\cite{xuNodeRAGStructuringGraphbased2025} further advanced this idea by deeply integrating KG and vector retrieval into a heterogeneous semantic graph composed of entity and document nodes, enabling unified retrieval within a single space and achieving more efficient and interpretable multi-level knowledge aggregation and generation.

%% file: Tables/Structured_Data.tex
\begin{table*}[!ht]
\renewcommand{\arraystretch}{1.3}
\centering
\caption{Comparison between Knowledge Graph and Unstructured Documents}
\label{tab:Knowledge Graph}
\begin{tabular}{|
  >{\centering\arraybackslash}m{0.2\textwidth} |
  >{\centering\arraybackslash}p{0.35\textwidth} |
  >{\centering\arraybackslash}p{0.35\textwidth} |
}
\hline
\textbf{Dimension} & \textbf{Knowledge Graph} & \textbf{Unstructured Documents} \\
\hline
\raisebox{-1.5\height}{Mechanism} & Represents entities and their relationships as nodes and edges, supporting structured, semantic retrieval and reasoning & Stores information in free-form natural language text, requiring NLP or embedding-based methods for retrieval \\
\hline
\raisebox{-1.5\height}{Retrieval Method} & Graph-based retrieval \par Path query \par Neighbor expansion
& Sparse retrieval \par Dense(semantic) retrieval \par Hybrid retrieval \\
\hline
\raisebox{-1.5\height}{Application Scenarios} & Multi-hop question answering \par Recommender systems \par Intelligent reasoning 
& Search engine document retrieval \par Sources like news, academic papers, reports \\
\hline
\end{tabular}

\end{table*}

%% file: Data/4/4.3Broad_Retrieval.tex
\subsection{Broad Retrieval}
\label{sec:broad retrieval}
As \gls{llms} are increasingly applied to knowledge-intensive and multi-domain tasks, their reliance on external information grows accordingly. Traditional precise retrieval, based on structured corpora, faces limitations in coverage and adaptability.
In contrast, broad retrieval (as illustrated in Fig.~\ref{fig:Broad Retrieval}) refers to approaches that expand the accessible knowledge space by retrieving information from large-scale, heterogeneous, and dynamic sources, such as web pages, social media, and multi-modal content. Building an efficient and trustworthy broad retrieval system has therefore become essential for mitigating knowledge-based hallucinations.

The main challenge of broad retrieval lies in the diversity of information sources and the variation in content granularity.  To address these issues, this section introduces representative advancements in broad retrieval from three key perspectives: Cross-Domain Generalization, Long-Context Comprehension and AI-generated Content Identification.

\input{Pictures/broad_retrieval}

\subsubsection{Cross-Domain Generalization}
Cross-Domain Generalization requires systems to perform information extraction across multiple knowledge domains and contexts. In real-world applications, particularly in open-domain question answering, multi-turn dialogue generation, and public opinion analysis, the necessary knowledge is often fragmented and heterogeneous, distributed across diverse sources. For instance, answering a question about the evolution of healthcare policy may require simultaneously drawing from news articles, government announcements, academic papers, and medical images or videos. This demands that the retrieval module not only recognizes the domain intent of a query, but also performs semantic alignment and integrative reasoning across sources and modalities, all without relying on predefined domain boundaries. Failure to do so can lead to fragmented or disconnected responses.

\paragraph{Web Search}
The internet, as a multi-source and real-time information carrier, has become an indispensable external knowledge source for mitigating knowledge-based hallucinations. WebGPT~\cite{nakanoWebGPTBrowserassistedQuestionanswering2022} and  IALM (Internet-Augmented Language Models)~\cite{lazaridouInternetaugmentedLanguageModels2022} are early variants of RAG systems that incorporate real web search results as external knowledge into the generation process of LLMs. WebGLM~\cite{liuWebGLMEfficientWebEnhanced2023} further improves WebGPT in accuracy and efficiency.
Recently deployed mainstream LLM applications, such as ChatGPT4.0~\cite{openaiGPT4TechnicalReport2024}, Gemini~\cite{team2023gemini} and Deepseek~\cite{liu2024deepseek} have also incorporated live web search to supplement their responses with up-to-date information, despite their underlying web searching and integrating mechanisms being mostly proprietary.

Academic research has also produced open, reproducible frameworks. Schick et al.~\cite{schickToolformerLanguageModels2023} proposed Toolformer, which can learn to decide when and how to call search engines during inference. WebWalker~\cite{wuWebWalkerBenchmarkingLLMs2025} adopts a multi-agent framework that enhances the retrieval system’s ability to navigate deeply through web content, improving both the depth and quality of information gathering.
DeepResearcher~\cite{zhengDeepResearcherScalingDeep2025} employs reinforcement learning to enable end-to-end training, allowing the model to autonomously plan retrieval and verify information in open, dynamic, and noisy real-world web environments. This significantly enhances the performance of LLMs in real-world web-based question answering tasks.
Considering that Web-based knowledge is often difficult to verify, and may contain inaccurate or misleading information, FoRAG (Factuality-optimized RAG)~\cite{caiFoRAGFactualityoptimizedRetrieval2024} employs a dual-granularity \gls{rlhf} framework to optimize the factual accuracy of answer generation in web-enhanced long-form question answering (LFQA). 
Furthermore, HtmlRAG~\cite{tanHtmlRAGHTMLBetter2025} improves traditional web \gls{rag} pipelines by introducing structure-aware preprocessing that directly models HTML content, rather than relying on plain-text extraction. This approach preserves the hierarchical structure and semantic relationships of web pages, enhancing the faithfulness, traceability, and overall robustness of generation in complex web environments.

These systems collectively demonstrate that integrating web search into RAG pipelines not only expands the model's accessible knowledge space but also offers a powerful means of enhancing factual consistency, particularly in fast-changing domains such as news, public health, and technology. However, they also introduce new challenges, including source reliability assessment, information overload, and retrieval latency, which remain active areas of research.

\paragraph{Multi-Modal RAG}
Multi-modal RAG~\cite{zhaoRetrievingMultimodalInformation2023} extends the traditional RAG paradigm beyond text-only inputs by incorporating heterogeneous modalities such as visual, auditory, or structural information into both retrieval and generation processes. This approach enables models to ground their responses in richer evidence and perform reasoning across different information sources, bridging perception and language understanding.
MuRAG~\cite{chenMuRAGMultimodalRetrievalAugmented2022} is the first framework to systematically extend retrieval from pure text to a text–image multi-modal setting, demonstrating that multi-modal retrieval can significantly enhance the performance of RAG systems. VisRAG~\cite{yuVisRAGVisionbasedRetrievalaugmented2025} also explore image-based retrieval techniques, proposing directly embedding images into documents. HtmlRAG~\cite{tanHtmlRAGHTMLBetter2025} retrieves web information directly from HTML data containing webpage structures.

These methods targeting multi-modal knowledge sources expand the input scope of RAG and enrich the accessible information for the model. However, the complexity of modality alignment introduces significant challenges. Semantic gaps across modalities can lead to modality ambiguity and misleading information, which not only compromise retrieval quality but may also introduce new hallucinations, thereby offsetting the benefits of enhanced knowledge coverage~\cite{wassermanREALMMRAGRealWorldMultiModal2025}.

\subsubsection{Long-Context Comprehension}
Another critical capability is the system’s ability to handle long-text inputs. Prior research has shown that models often suffer from the “lost-in-the-middle” phenomenon~~\cite{liuLostMiddleHow2023} when processing overly long inputs—exhibiting significantly reduced attention to information located in the middle of the input sequence. This attention dilution phenomenon significantly weakens the model’s ability to utilize retrieved content effectively, leading to reduced faithfulness in its responses.
Many models have been optimized for long-text reading. 
Some models adopt sparse attention~\cite{ yuanNativeSparseAttention2025, xuXAttentionBlockSparse2025} or linear attention~\cite{huangImprovingContextualFaithfulness2025} mechanisms to optimize attention computation, some improve reading performance by processing long texts recursively or in chunks~\cite{liGraphReaderBuildingGraphbased2024, leeHumanInspiredReadingAgent2024}; others fine-tune the model to enhance its performance on long-text inputs~\cite{huangImprovingContextualFaithfulness2025}.
Overall, iterative chunk-based processing provides stronger dynamic generalization when handling long texts. This approach, unlike attention modification or fine-tuning, preserves the model's sensitivity to short texts. In addition, it can be integrated into retrieval post-processing procedures. For example, chunking can be performed during reranking, which reduces computational cost and shows promising potential for real-world applications.

Enhancing long-context reading ability enables models to make better use of retrieved information, mitigating hallucinations caused by insufficient knowledge grounding. It also improves the comprehension of complex, lengthy queries, further reducing the likelihood of hallucinations.

\subsubsection{AI-generated Content Identification}
With the rapid expansion of \gls{aigc}~\cite{caoComprehensiveSurveyAIGenerated2023}, such as auto-written articles and images, the risk of \gls{llms} retrieving unreliable AI-generated content has increased, potentially contaminating web knowledge and degrading model quality~\cite{shumailovAIModelsCollapse2024}. To prevent this, RAG systems should be equipped with mechanisms to detect and exclude \gls{aigc} during retrieval.
Recent studies have therefore explored several approaches to identify and filter AI-generated content during retrieval. Among them, watermarking embeds~\cite{zhaoSoKWatermarkingAIGenerated2025} imperceptible patterns (e.g., token selection or probability distributions) into generated text for later verification, offering strong traceability but requiring cooperation from the generator and remaining vulnerable to paraphrasing or translation. Statistical detection methods, such as GLTR~\cite{gehrmannGLTRStatisticalDetection2019} and DetectGPT~\cite{mitchellDetectGPTZeroShotMachineGenerated2023}, identify AI-generated text by analyzing lexical diversity, syntactic patterns, repetition rates, and perplexity distributions. Another practical direction is model-based detection, where classifiers (e.g., RoBERTa-based models~\cite{rakibmollahDetectionFakeNews2023}) are trained to distinguish human- and machine-written content.
As \gls{aigc} becomes more pervasive, future RAG frameworks should integrate such detection modules to filter potentially hallucination-inducing content at the source, ensuring reliability in retrieval and generation.

%% file: Pictures/broad_retrieval.tex
\begin{figure}[t!]  %
\centering
\includegraphics[width=1\linewidth, trim=0 1500 0 1050, clip]{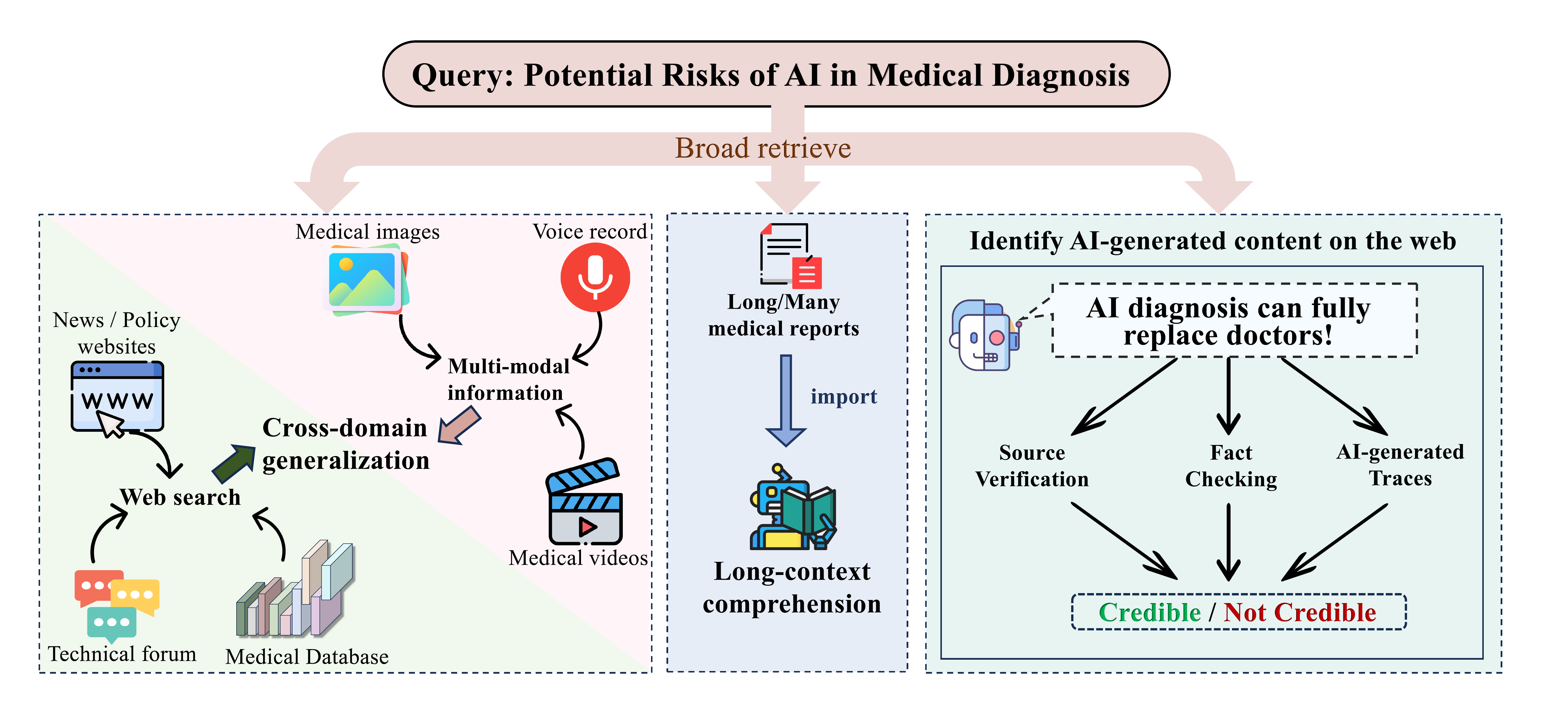}
\caption{Demonstration of a typical broad retrieval process}
\label{fig:Broad Retrieval}
\Description{A conceptual diagram illustrating the broad retrieval process that integrates multi-modal, web, and domain-specific sources, as well as the identification of AI-generated content, for reliable question answering}
\end{figure}

%% file: Data/4/4.4.tex
\subsection{Capability-Oriented Comparison of RAG Approaches}









From the perspective of mitigating knowledge-based hallucinations, the core role of RAG is to enhance the ability of LLMs to acquire and utilize external knowledge~\cite{sharma2025retrieval}. Existing RAG approaches primarily improve knowledge augmentation capabilities through three directions: retrieval decision optimization, evidence grounding enhancement, and knowledge space expansion.

RAG pipeline optimization methods mainly focus on decision-making capabilities during the process of introducing external knowledge. By improving retrieval triggering mechanisms, retrieval planning strategies, and knowledge integration approaches, these methods enhance the ability of LLMs to determine when and how to utilize external knowledge. Therefore, they are particularly advantageous for tasks requiring dynamic retrieval decisions and multi-turn interactions. However, when task requirements are explicit and knowledge sources are stable, complex retrieval planning mechanisms may introduce additional computational overhead.

Precise retrieval methods mainly improve the capability of evidence localization and utilization within a limited knowledge space. Through techniques such as query rewriting, hybrid retrieval, reranking, and structured retrieval, these approaches enhance the alignment between retrieved evidence and generated outputs. Therefore, they are suitable for domain-specific tasks with well-defined knowledge boundaries, reliable sources, and high requirements for evidence consistency, such as enterprise knowledge bases and scientific literature retrieval. However, in open-domain tasks, their performance may be constrained by limited knowledge coverage and query understanding capability.

Broad retrieval methods further expand the scope of knowledge accessible to LLMs by extending fixed corpora to dynamic, multi-source, and heterogeneous information environments. These approaches are suitable for tasks involving frequently updated knowledge or cross-domain information integration. However, expanding knowledge sources also introduces new reliability challenges, including noise filtering, evidence evaluation, and retrieval efficiency.

The three RAG design paradigms enhance different aspects of the knowledge acquisition process, including decision reliability, evidence utilization capability, and knowledge coverage. Their effectiveness depends on balancing retrieval dynamics, knowledge reliability, and coverage requirements.

%% file: Data/5.Logic-Based_Hallucination_and_Reasoning.tex


\section{Logic-based Hallucination and Reasoning}
\label{sec:section5}

Logic-based hallucinations refer to cases where the generated content may be factually correct, but the final output lacks reliability due to incorrect inferences during the reasoning process, inconsistent conclusions, or unreasonable deductions caused by improper contextual understanding. It should be noted that natural language reasoning relies not only on explicit logical relationships but also on contextual information, implicit assumptions, and pragmatic understanding. Therefore, logic-based hallucinations in this work do not simply correspond to formal logical errors; instead, they primarily concern whether models can produce reliable, consistent, and reasonable reasoning results under a given context. Unlike factual hallucinations, which are typically caused by missing or incorrect knowledge, logic-based hallucinations mainly arise from limitations in information integration, reasoning execution, and implicit semantic understanding.

This section investigates reasoning augmentation approaches, particularly the development of techniques such as Chain-of-Thought (CoT), Tool-Augmented Reasoning, and Symbolic Reasoning. The overall relationship among these approaches and their underlying enhancement mechanisms is illustrated in Fig.~\ref{fig:Reasoning Process}. These methods improve the ability of language models to identify, generate, and correct reliable reasoning processes by explicitly expanding reasoning steps, introducing external verification mechanisms, and imposing structured constraints.

\input{Pictures/reasoning_process}

\input{Data/5/5.1Chain-of-Thought_Prompting}

\input{Data/5/5.2Tool-Augmented_Reasoning}

\input{Data/5/5.3Symbolic_reasoning}

\input{Data/5/5.4}

%% file: Pictures/reasoning_process.tex
\begin{figure*}[t]  %
\centering
\includegraphics[width=1\linewidth, trim=100 70 50 100, clip]{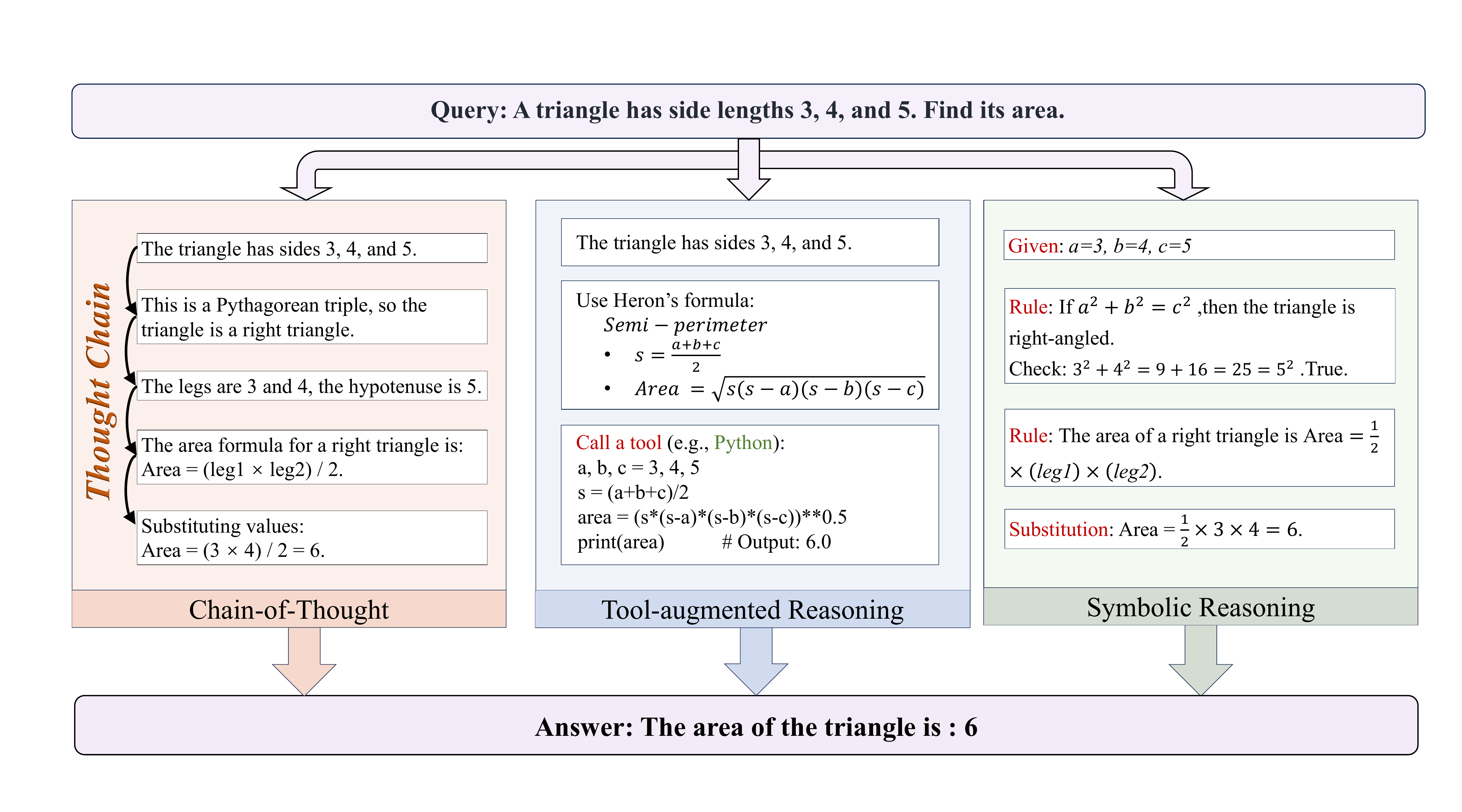}
\caption{Implementation process of three representative reasoning enhancement methods: Chain-of-Thought, Tool-augmented Reasoning and Symbolic Reasoning}
\label{fig:Reasoning Process}
\Description{Illustration of different reasoning processes in large language models, including chain-of-thought reasoning, symbolic reasoning, and tool-augmented reasoning. The diagram demonstrates how step-by-step logical deduction, mathematical formulation, and external tool usage collaboratively enhance reasoning accuracy and reliability.}
\end{figure*}

%% file: Data/5/5.1Chain-of-Thought_Prompting.tex
\subsection{Chain-of-Thought}
\label{sec:CoT}

Chain-of-Thought (CoT)~\cite{chuNavigateEnigmaticLabyrinth2024, weiChainofThoughtPromptingElicits2023} is an inference-time reasoning enhancement paradigm that improves the reliability of \gls{llms} by explicitly representing intermediate reasoning states during problem solving. Different from direct answer generation, which requires the model to implicitly infer complex relationships within a single prediction process, CoT guides the model to decompose a complex problem into a sequence of intermediate reasoning steps through structured prompting. Therefore, the primary capability enhanced by CoT is the organization and decomposition of reasoning processes, enabling models to construct more coherent reasoning trajectories and reducing logical hallucinations caused by skipped steps, implicit assumptions, and inconsistent reasoning transitions.

Due to its simplicity and low deployment cost, CoT has become one of the most widely adopted approaches for improving LLM reasoning capabilities. The method was initially introduced through few-shot prompting~\cite{weiChainofThoughtPromptingElicits2023} and was later extended to zero-shot settings, where simple instructions such as “Let’s think step by step”~\cite{kojimaLargeLanguageModels2023} or adaptive reasoning prompts such as “Wait”~\cite{muennighoffS1SimpleTesttime2025} were shown to effectively elicit multi-step reasoning behaviors. By generating explicit intermediate reasoning chains, CoT provides a structured representation of the model’s inference process, improving both reasoning transparency and the ability to identify potential logical inconsistencies.
Building upon this idea, subsequent studies have further enhanced CoT by introducing additional decomposition and verification mechanisms. For example, Ling et al.~\cite{lingDeductiveVerificationChainofThought2023} proposed Natural Program, which transforms CoT reasoning into executable intermediate processes and performs logical verification over individual reasoning steps. Such extensions improve the faithfulness of reasoning trajectories by combining explicit reasoning decomposition with step-level validation, thereby further reducing logical hallucinations caused by unreliable internal inference.
Xu et al.~\cite{xuFaithfulLogicalReasoning2024} propose a symbolic \gls{cot} framework that also ensures the reliability of the logical chain. It first converts natural language into symbolic representations and then applies symbolic logic rules to solve problems, leveraging a rule-based system to execute and verify each reasoning step sequentially.

In recent years, \gls{cot} has continued to evolve and expand to improve its reasoning robustness against hallucinations. For example, the Self-Consistency~\cite{wangSelfConsistencyImprovesChain2023} strategy generates multiple \gls{cot} paths for the same question and selects the answer by majority voting, effectively reducing randomness and instability in single-path generation. 
Some approaches~\cite{xuFaithfulLogicalReasoning2024, lyuFaithfulChainofThoughtReasoning2023} replace natural language reasoning with symbolic reasoning, enabling the reasoning chain to exhibit stronger logical consistency and verifiability.
Zhu et al.~\cite{zhuChainofThoughtMattersImproving2025} introduces a reasoning path supervision mechanism to guide \gls{llms} in learning complete and coherent reasoning chains within long-context environments. By doing so, it enhances the model’s reasoning ability and factual faithfulness in long-text tasks. The proposed method effectively mitigates hallucination issues arising from information loss or step-skipping during long-context processing.

Although \gls{cot} has demonstrated significant advantages across various reasoning tasks, existing studies have noted that its effectiveness is task-sensitive and does not guarantee performance improvement in all scenarios~\cite{liuMindYourStep2025}. To address this limitation, many approaches have been proposed to equip \gls{llms} with the ability to autonomously decide whether to invoke \gls{cot} and to dynamically determine the appropriate length or style of the reasoning chain~\cite{wangDynamicChainofThoughtAdaptive2025}.
This helps prevent unnecessary long reasoning paths that may introduce logical hallucinations.

Overall, CoT is not merely a prompting strategy, but also represents a paradigm shift in language model generation—from “black-box responses” to “auditable reasoning chains”.

%% file: Data/5/5.2Tool-Augmented_Reasoning.tex
\subsection{Tool-Augmented Reasoning}
\label{sec:TAR}
When handling tasks involving precise computation, fact verification, or structured logical reasoning, \gls{llms} may produce hallucinations due to unreliable intermediate reasoning steps, especially in complex tasks that require explicit execution and verification, such as code generation, mathematical problem solving, and multi-step reasoning. To improve reasoning reliability, the Tool-augmented Reasoning paradigm has been proposed in recent years~\cite{schickToolformerLanguageModels2023, yaoReActSynergizingReasoning2023}.
Different from approaches that only improve the organization of reasoning processes, Tool-augmented Reasoning focuses on enhancing the reliability of critical reasoning steps by introducing external tools or modules to execute, verify, or supplement intermediate operations. Therefore, this paradigm can be viewed as a capability enhancement for critical-step execution, where the model coordinates natural language reasoning with reliable external feedback.

Unlike traditional end-to-end text generation, this method transforms reasoning into a collaborative process of \textit{“language generation + tool invocation.”}~\cite{schickToolformerLanguageModels2023}
The model is not only responsible for generating natural language output but also for planning when to call upon external modules. For example, in code generation tasks, it may invoke a program executor to verify the runnability of its output~\cite{maSciAgentToolaugmentedLanguage2024}, while in mathematical problem solving, it can convert natural language into symbolic expressions for computation~\cite{xuFaithfulLogicalReasoning2024, lyuFaithfulChainofThoughtReasoning2023}. 
Tool-augmented \gls{llms} have been applied to many domains beside code and math.
Lu et al.~\cite{luTARTOpenSourceToolAugmented2025} proposed TART, which handles table question answering by dynamically selecting table operation functions relevant to the current question and generating a corresponding reasoning plan. 
Enhancing reasoning by incorporating external knowledge bases is now a common approach, including web search engines~\cite{trivediInterleavingRetrievalChainofThought2023, chenChatCoTToolAugmentedChainofThought2023} and knowledge graphs~\cite{sunThinkonGraphDeepResponsible2024, maThinkonGraph20Deep2025}. These approaches explicitly decompose the reasoning process into multiple intermediate steps, incorporating external information at each stage to dynamically enrich the context, thereby providing more comprehensive background support and factual grounding for subsequent reasoning.

As this field evolves, dynamic tool augmentation has emerged as a key research frontier. Unlike static tool chains, dynamic approaches such as ToolFive~\cite{luToolFiVeEnhancingToolAugmented2025}, ANSWERED~\cite{panANSWEREDAdaptiveToolAugmented2024} and SciAgent~\cite{maSciAgentToolaugmentedLanguage2024} endow LLMs with decision-making capabilities that allow them to determine whether, when, and which tools to invoke based on the specific task and contextual cues. More advanced systems~\cite{gaoEfficientToolUse2025, chenAdvancingToolAugmentedLarge2024, zhangEvaluatingImprovingToolAugmented2023} even support reflective evaluation—enabling the model to reassess its initial outputs, incorporate intermediate feedback, and adaptively reconfigure tool usage to arrive at improved answers. This highly flexible and interactive reasoning workflow greatly enhances model adaptability and robustness, offering a promising direction for mitigating hallucinations in complex, multi-stage tasks.

%% file: Data/5/5.3Symbolic_reasoning.tex
\subsection{Symbolic Reasoning}
\label{sec:SR}
Contemporary research has increasingly explored the integration of symbolic reasoning~\cite{panLogicLMEmpoweringLarge2023, wangChatLogicIntegratingLogic2024} with \gls{llms} to improve the logical consistency and reliability of model reasoning processes. Different from approaches that primarily focus on organizing reasoning trajectories or introducing external execution tools, symbolic reasoning enhances reasoning reliability by imposing explicit symbolic representations and formal logical constraints on intermediate reasoning steps. By transforming natural language descriptions into structured symbolic forms, this paradigm enables reasoning relationships between different steps to be explicitly represented, checked, and verified.

The core idea is to leverage \gls{llms} as semantic interpreters that translate natural language inputs and intermediate reasoning states into symbolic representations, while symbolic solvers or rule-based systems perform formal deduction and consistency checking. Through this interaction, the flexibility and language understanding capabilities of \gls{llms} can be combined with the strict logical validity of symbolic systems. Consequently, symbolic reasoning reduces logical hallucinations caused by inconsistent reasoning chains, invalid inference patterns, or violations of logical constraints, providing a more reliable framework for complex reasoning tasks.

Specifically, Wang et al. proposed ChatLogic~\cite{wangChatLogicIntegratingLogic2024}, which employs a LLM to convert natural language questions into symbolic logic. A logic programming engine then carries out multi-step deductive reasoning, and the results are translated back into natural language. In this approach, the reasoning process is executed by the logic programming engine, ensuring high reliability. However, it is limited in the types of reasoning tasks it can handle, resulting in a narrow coverage of reasoning scenarios.
Similarly, Logic-LM~\cite{panLogicLMEmpoweringLarge2023} also utilizes a LLM to convert natural language questions into symbolic representations, and then invokes a symbolic reasoner to solve the problem.
Xu et al. proposed SymbCoT~\cite{xuFaithfulLogicalReasoning2024}, which translates natural language into symbolic representations and then performs CoT reasoning using an LLM. Finally, a verifier is employed to examine the reasoning process. This approach directly leverages the reasoning capability of the \gls{llms}, offering greater adaptability, while the subsequent verification step ensures the logical correctness of the reasoning process.

The integration of LLMs with symbolic reasoning embodies the concept of neuro-symbolic AI~\cite{deraedt2021neural, nawaz2025review} and represents an important direction for improving the logical reliability of LLMs, reflecting a broader trend toward building hybrid reasoning architectures that combine symbolic logic with neural language models. By assigning tasks with high logical complexity to symbolic engines and relying on LLMs for natural language understanding and generation, such systems achieve greater reliability, interpretability, and robustness in complex reasoning tasks. This research direction not only offers a promising pathway for mitigating logic-based hallucinations in large models but also reintroduces verifiable reasoning capabilities into modern AI systems, laying the groundwork for safer and more trustworthy applications in high-stakes domains such as law, science, and education.

%% file: Data/5/5.4.tex
\subsection{Capability-Oriented Comparison of Reasoning Approaches}








From the perspective of mitigating logic-based hallucinations, the core role of reasoning enhancement approaches is to improve the reasoning reliability of LLMs during the generation process. Existing methods can be broadly categorized into three directions: reasoning process decomposition, external interaction and verification, and formal reasoning constraints, which correspond to improvements in reasoning organization capability, external verification capability, and logical constraint capability, respectively.

Chain-of-Thought (CoT)-based methods mainly enhance the organization capability of reasoning processes by explicitly generating intermediate reasoning steps. These approaches decompose complex tasks into multiple sequential steps and improve reasoning transparency, making them effective for tasks with clear decomposition structures, such as mathematical reasoning and logical reasoning. However, since the reasoning process still relies on model-generated intermediate steps, incorrect initial reasoning paths may lead to error propagation, and independent verification mechanisms are often lacking.

Tool-augmented reasoning methods improve verification capability by introducing external tools and interaction mechanisms. By invoking modules such as calculators, search engines, code interpreters, or specialized solvers, these approaches enable external components to handle parts of the computation, information retrieval, and verification processes. They are particularly suitable for tasks requiring precise computation, dynamic information acquisition, or environmental feedback. However, their effectiveness depends on tool selection, invocation strategies, and the reliability of external results, and inappropriate tool usage may introduce additional errors.

Symbolic reasoning methods enhance logical consistency through explicit symbolic representations and formal constraints. By transforming natural language problems into logical rules, program code, or mathematical expressions, these approaches reduce uncertainty in neural generation processes and demonstrate advantages in tasks with well-defined rules and strict constraints. However, for open-domain problems, knowledge formalization and the conversion between natural language and symbolic representations remain challenging.

The three categories of reasoning enhancement approaches improve the logical reasoning capability of LLMs from three complementary perspectives: reasoning organization, external verification, and formal constraints. These approaches are not mutually exclusive alternatives but instead address different task structures, verification requirements, and external resource conditions.

%% file: Data/6.Synergy_and_Hybrid_Approaches.tex
\section{Composite Hallucination and Agentic Systems}
\label{sec:section6}

\input{Pictures/general_framework}

\input{Data/6/6.1Motivation}

\input{Data/6/6.2Agentic_Systems}

\input{Data/6/6.3}

%% file: Pictures/general_framework.tex
\begin{figure}[t!]  %
\centering
\includegraphics[width=1\linewidth, trim=250 100 150 100, clip]{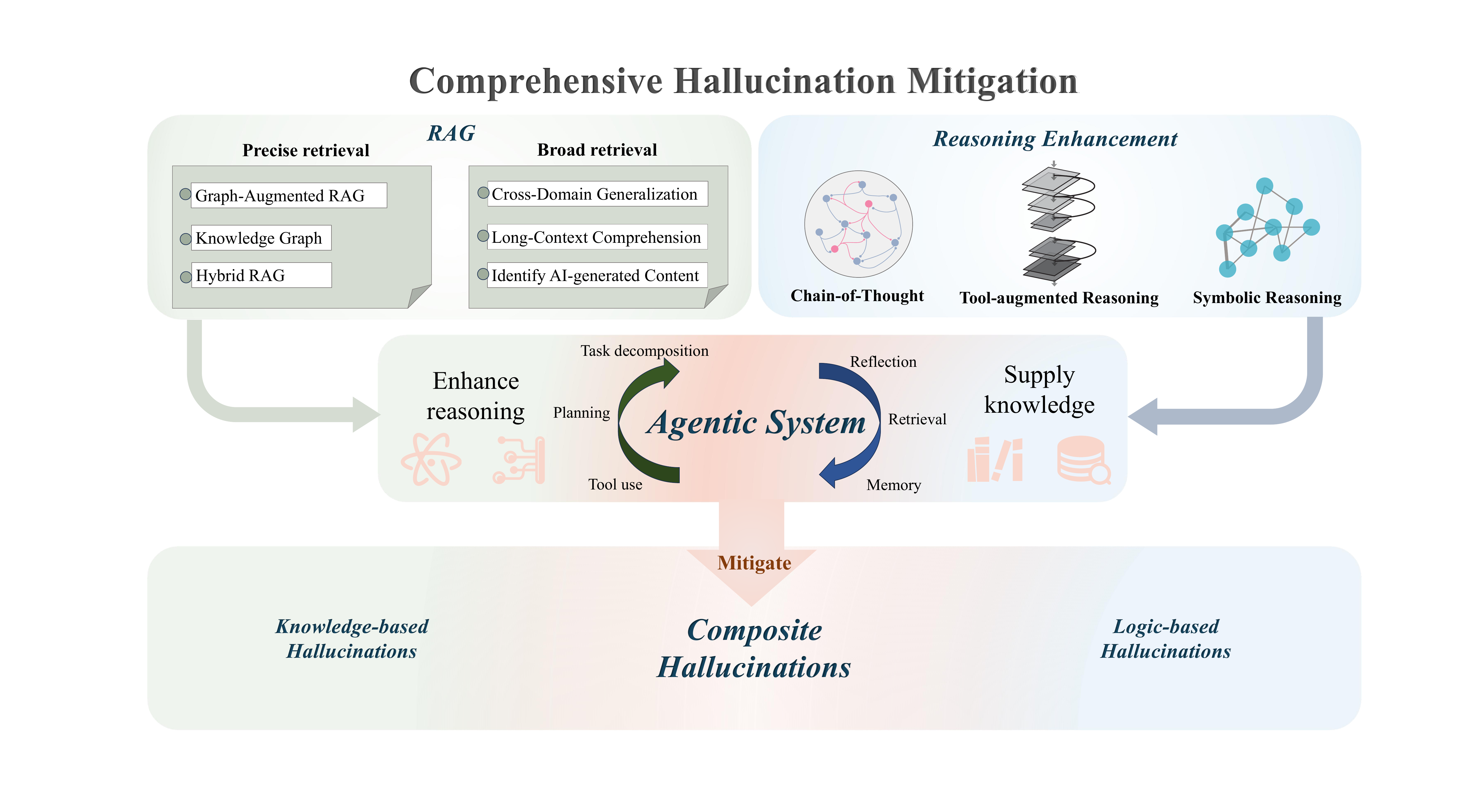}
\caption{Agentic Framework Integrating RAG and Reasoning Enhancement for Comprehensive Hallucination Mitigation}
\label{fig:General Framework}
\Description{An overview of the general framework for hallucination mitigation in large language models, illustrating the interaction between retrieval-augmented generation (RAG), reasoning enhancement, and agentic systems. It highlights how precise and broad retrieval supply knowledge, reasoning and tool-augmented methods enhance inference, and agentic mechanisms such as reflection, planning, and memory jointly contribute to comprehensive hallucination mitigation across knowledge-based, logic-based, and composite types.}
\end{figure}

%% file: Data/6/6.1Motivation.tex
\subsection{Motivation: Why Agentic Systems?}

With the increasing complexity of tasks handled by LLMs, hallucinations have gradually evolved from isolated factual or reasoning errors into more complex forms arising from the interaction of multiple factors. Previous sections have separately discussed mitigation approaches for knowledge-based and logic-based hallucinations. However, complex tasks often require models to simultaneously possess reliable knowledge acquisition capabilities and multi-step reasoning abilities to avoid hallucination generation, which calls for effective integration of retrieval and reasoning processes (Fig.~\ref{fig:General Framework}). Nevertheless, achieving effective integration between retrieval and reasoning still faces several key challenges, including dynamically determining when to retrieve information, selecting effective evidence, and leveraging feedback mechanisms to optimize the reasoning process. These requirements have driven the development of agentic systems, enabling models to achieve more flexible coordination among multiple processes, including retrieval, reasoning, planning, and feedback.

In this work, agentic systems are defined as system frameworks that are built upon LLMs, integrate reasoning capabilities with external information acquisition mechanisms, and can further incorporate components such as self-reflection, feedback mechanisms, and tool interaction. This section introduces representative agentic paradigms that integrate retrieval and reasoning capabilities and analyzes their roles in mitigating complex hallucinations.

It should be noted that this survey focuses on agentic systems primarily as architectural frameworks for enhancing the knowledge acquisition and reasoning capabilities of LLMs, rather than investigating the reliability issues of agent collaboration itself. Therefore, system-level issues introduced by agent collaboration and execution processes, such as communication failures, task planning errors, and tool invocation anomalies, are not discussed separately. 

%% file: Data/6/6.2Agentic_Systems.tex
\subsection{Representative Agentic Frameworks Integrating Retrieval and Reasoning}
\label{sec:6.2}

The integration of retrieval and reasoning has evolved from simple capability combination toward increasingly autonomous and adaptive frameworks~\cite{liang2025reasoning}. Early approaches mainly focus on enabling LLMs to invoke retrieval during the reasoning process, allowing external knowledge acquisition to be dynamically coordinated with intermediate reasoning steps~\cite{li2025survey}. Subsequently, multi-agent frameworks further extend this paradigm by decomposing retrieval, reasoning, and verification into specialized modules, providing greater flexibility for solving complex tasks. More recent autonomous agentic systems introduce self-reflection and feedback mechanisms, enabling continuous evaluation and refinement of their own behaviors. Although these frameworks differ in architectural design, they share a common objective: mitigating composite hallucinations by jointly enhancing external knowledge grounding, logical reasoning capability, and error correction ability.

\subsubsection{Reasoning-Augmented Retrieval Frameworks}
\label{sec:6.2.1}

Early forms of integrating retrieval and reasoning enable LLMs to dynamically acquire external knowledge during the reasoning process. Unlike conventional RAG pipelines that perform retrieval only once based on the initial query and directly incorporate retrieved information into the generation process, reasoning-augmented retrieval frameworks allow retrieval decisions to be influenced by intermediate reasoning states. This design enables models to identify knowledge gaps during reasoning, proactively acquire additional evidence when necessary, and further incorporate retrieved information into subsequent reasoning steps, thereby strengthening the interaction between knowledge acquisition and logical reasoning processes.

ReAct~\cite{yaoReActSynergizingReasoning2023} and IRCoT~\cite{trivediInterleavingRetrievalChainofThought2023} represent two representative approaches within this paradigm. ReAct introduces an interactive “reasoning-action-observation” loop, allowing LLMs to generate intermediate reasoning traces and determine when to invoke external tools based on the current reasoning state. Through continuous interaction with external environments, ReAct enables models to actively acquire additional factual information during problem solving, rather than relying on a fixed retrieval step before generation.
IRCoT further extends this idea by explicitly interleaving retrieval with Chain-of-Thought reasoning. Instead of performing retrieval only at the beginning of the reasoning process, IRCoT dynamically inserts retrieval operations between reasoning steps, allowing intermediate reasoning outcomes to guide subsequent information acquisition.
Both approaches overcome the conventional “retrieve-then-generate” paradigm in RAG and lay the foundation for subsequent Agentic Systems that integrate retrieval, reasoning, and autonomous decision-making capabilities.

\subsubsection{Multi-Agent Retrieval and Reasoning Frameworks}
\label{sec:6.2.2}

Although reasoning-augmented retrieval frameworks enable the interaction between retrieval and reasoning processes within a single model, they still mainly rely on a single LLM as the central controller. For complex tasks, different subtasks may involve heterogeneous knowledge acquisition requirements and diverse reasoning processes, making it difficult for a single-model architecture to flexibly coordinate various information acquisition and reasoning procedures~\cite{suresh2026agenticrag}. Recent studies have therefore explored decomposing capabilities such as retrieval, planning, reasoning, and evidence processing into multiple specialized agents. Through communication and collaboration among agents, multi-agent systems enable the coordination and integration of heterogeneous capabilities.

AutoGen~\cite{wu2024autogen} and MetaGPT~\cite{hong2024metagpt} are representative early frameworks following this paradigm. AutoGen establishes a conversational multi-agent collaboration mechanism, allowing agents with different roles to exchange information, invoke tools, and perform collaborative reasoning toward shared task objectives. MetaGPT further introduces role-specific agents with explicitly defined responsibilities. By simulating professional divisions in software development workflows, it decomposes complex tasks into multiple interdependent subtasks completed through collaboration among agents. These general-purpose multi-agent frameworks demonstrate that modularly organizing LLM capabilities can effectively improve flexibility in complex task solving, providing foundations for the subsequent development of retrieval-augmented multi-agent systems.

Building upon these advances, recent studies have further incorporated multi-agent collaboration mechanisms into retrieval-augmented generation scenarios, enabling knowledge acquisition and reasoning processes to be jointly performed by different agents. MA-RAG~\cite{nguyenMARAGMultiAgentRetrievalAugmented2025} and AgentUniRAG~\cite{phamAgentUniRAGTrainableOpenSource2025} are representative examples of this paradigm. Specifically, MA-RAG introduces multiple specialized agents to collaboratively perform task planning, knowledge retrieval, and answer generation, enabling dynamic knowledge acquisition and complex task reasoning. AgentUniRAG further explores a trainable multi-agent RAG architecture, improving the system’s adaptability to complex tasks and diverse knowledge requirements by optimizing the collaboration mechanisms among agents. Meanwhile, recent studies have also investigated the integration of agents with external tools to further enhance the reasoning and execution capabilities of models in complex tasks. For example, Wu et al.~\cite{wuAgenticReasoningStreamlined2025} introduce tool-use mechanisms that enable agents to dynamically leverage external resources according to task requirements, thereby extending the reasoning capabilities of LLMs and improving reliability during complex problem-solving processes.
Compared with single-agent retrieval-augmented reasoning approaches, multi-agent systems improve flexibility in handling complex problems by explicitly separating knowledge acquisition, reasoning, and task coordination processes.

\subsubsection{Autonomous Agentic Systems with Iterative Feedback Loops}
\label{sec:6.2.3}

Beyond modular collaboration, recent autonomous agentic systems further introduce self-reflection, feedback, and iterative optimization mechanisms~\cite{dong2025rag}. Unlike previous frameworks that mainly focus on integrating retrieval and reasoning capabilities, autonomous agentic systems emphasize continuous interaction among planning, retrieval, reasoning, and evaluation processes~\cite{li2025dora}. Through an “execution–evaluation–feedback–optimization” closed loop, these systems can dynamically adjust subsequent behaviors during task execution. Specifically, agents can evaluate the reliability of intermediate results, identify potential errors or insufficient evidence, and utilize feedback information to update subsequent retrieval strategies, reasoning trajectories, or task plans. Therefore, hallucination mitigation is no longer limited to error prevention before generation but is further extended to dynamic detection and correction during execution.

Self-Refine~\cite{madaan2023self} and Reflexion~\cite{shinn2023reflexion} are representative early approaches that explore feedback-driven optimization mechanisms. Self-Refine introduces an iterative “generation–feedback–refinement” process, enabling models to continuously improve their outputs based on self-generated feedback. Reflexion further extends this idea to language agent scenarios, allowing agents to perform self-reflection based on feedback collected during task execution and optimize subsequent behaviors using reflective information. These approaches demonstrate the importance of feedback mechanisms in improving model reliability and reducing error propagation.

In retrieval-augmented scenarios, Reflective RAG~\cite{wu2026reflective} further incorporates self-reflection mechanisms into the retrieval-augmented generation process, enabling agents to evaluate the effectiveness of previous retrieval behaviors and dynamically adjust future retrieval strategies based on intermediate feedback. Compared with approaches relying on fixed retrieval procedures, Reflective RAG establishes a closed loop among retrieval, reasoning, and feedback, thereby improving knowledge acquisition and external knowledge grounding in multi-step tasks.

By introducing continuous feedback and adaptive optimization mechanisms, autonomous agentic systems provide a more effective solution for addressing composite hallucinations in complex tasks. Compared with traditional retrieval-augmented frameworks and multi-agent collaboration frameworks, these systems not only leverage external knowledge and modular capabilities to accomplish tasks but also actively identify and correct errors during execution, thereby reducing the risk of hallucination accumulation and propagation in long-horizon tasks.

%% file: Data/6/6.3.tex
\subsection{Capability-Oriented Comparison of Agentic Systems}

From the perspective of hallucination mitigation, the three representative agentic system paradigms reflect different stages in the integration of retrieval and reasoning capabilities. They are not mutually exclusive alternatives, but rather represent progressive extensions targeting the increasing demands for knowledge acquisition, capability coordination, and error correction in complex task solving. The applicability of each paradigm depends not only on task complexity but also on the underlying causes of hallucinations and the capabilities that need to be enhanced.

Reasoning-enhanced retrieval frameworks mainly improve the interaction between retrieval and reasoning processes, enabling models to dynamically acquire knowledge based on the current reasoning state and adjust subsequent reasoning processes with external evidence. This paradigm is suitable for hallucination problems caused by insufficient knowledge, ineffective evidence utilization, or the disconnection between retrieval and reasoning. However, since retrieval and reasoning are still primarily performed within a unified framework, their ability to handle multi-source knowledge integration, complex task decomposition, or accumulated errors remains limited.

Multi-agent retrieval and reasoning frameworks further improve capability coordination through modular task decomposition. By assigning functions such as task planning, retrieval, evidence processing, and reasoning to different components, this paradigm can better handle composite tasks involving heterogeneous knowledge, multi-stage reasoning, and coordination among multiple capabilities, thereby mitigating composite hallucinations caused by the joint effects of knowledge deficiency, reasoning errors, and information conflicts. However, coordination and information exchange among modules may also introduce additional risks of error propagation.

Autonomous agentic systems further introduce self-reflection, feedback, and iterative optimization mechanisms, enabling systems to detect intermediate errors during task execution and dynamically adjust subsequent retrieval and reasoning strategies. This paradigm is particularly suitable for long-horizon tasks and highly uncertain environments, where it can alleviate error accumulation during reasoning processes. However, its effectiveness relies on the reliability of feedback and evaluation mechanisms, while also introducing higher computational costs and system complexity.

%% file: Data/7.Benchmarks.tex
\section{Benchmarks}
\label{sec:section7}

This section introduces representative benchmarks for evaluating hallucinations in LLMs. Unlike the hallucination mitigation approaches discussed above, benchmarks are primarily designed to measure model performance in terms of knowledge reliability, reasoning consistency, and complex task execution. Based on their evaluation objectives, existing benchmarks mainly cover three directions: knowledge-based hallucinations, logic-based hallucinations, and agentic workflows.

\textit{Benchmarks for Knowledge-based Hallucination.}
Benchmarks for knowledge-based hallucinations mainly focus on factual errors, missing knowledge, and biases in external knowledge understanding during model generation. They are designed to evaluate the factual reliability of LLMs and their capabilities in knowledge acquisition and utilization.
Existing benchmarks can be broadly divided into two categories. The first category focuses on the internal knowledge capabilities of models, evaluating whether models can generate truthful and reliable responses based on their parametric knowledge, such as TruthfulQA~\cite{linTruthfulQAMeasuringHow2022} and FreshQA~\cite{vuFreshLLMsRefreshingLarge2023}. The second category focuses on reliability in external knowledge-enhanced scenarios, evaluating whether models can avoid propagating incorrect information when utilizing retrieved or domain-specific knowledge, such as MedHallu~\cite{panditMedHalluComprehensiveBenchmark2025}.
Recently, several studies have further investigated fine-grained hallucination analysis. For example, HalluLens~\cite{bangHalluLensLLMHallucination2025} improves the diagnostic capability of hallucination evaluation by distinguishing between intrinsic hallucination and extrinsic hallucination.

\input{Tables/Benchemarks}
\textit{Benchmarks for Logic-based Hallucination.}
Benchmarks for logic-based hallucinations mainly focus on errors occurring during model reasoning processes, including inconsistent reasoning chains, incorrect intermediate steps, and logically invalid conclusions. These benchmarks are designed to evaluate the process reliability and logical consistency of LLMs in complex reasoning tasks.
According to different evaluation objectives, existing benchmarks mainly cover three aspects. The first category focuses on final answer correctness, such as BIG-bench~\cite{srivastavaImitationGameQuantifying2023}, which evaluates reasoning capability primarily through task-level outcomes. The second category further examines intermediate reasoning processes, including PrOntoQA~\cite{saparovLanguageModelsAre2023}, ProofWriter~\cite{tafjordProofWriterGeneratingImplications2021}, ToolBench~\cite{qinToolLLMFacilitatingLarge2023}, API-Bank~\cite{liAPIBankComprehensiveBenchmark2023}, and LogicBench~\cite{parmarLogicBenchSystematicEvaluation2024}. These benchmarks analyze reasoning steps, tool invocation processes, and logical verification capabilities. The third category focuses on argument structures and logical validity, such as ReClor~\cite{yuReClorReadingComprehension2020} and LogiQA~\cite{liuLogiQAChallengeDataset2020}, which evaluate the ability of models to identify complex argumentative relationships and logical errors.

\textit{Benchmarks for Agentic Workflows.}
With the evolution of LLMs into Agentic Systems equipped with planning, retrieval, reasoning, and tool-use capabilities, hallucination evaluation has gradually expanded from single-turn generation tasks to multi-step task execution processes. Unlike traditional benchmarks, agentic workflow benchmarks evaluate not only final outputs but also system reliability throughout task decomposition, information acquisition, tool interaction, and multi-turn execution.
AgentBench~\cite{liuAgentBenchEvaluatingLLMs2023} provides a comprehensive evaluation framework across multiple interactive environments to assess the execution capabilities of LLM-based agents. For domain-specific scenarios, L-MARS~\cite{wangLMARSLegalMultiAgent2025} and InfoDeepSeek~\cite{xiInfoDeepSeekBenchmarkingAgentic2025} focus on legal analysis and web information tasks, respectively, evaluating agents' knowledge acquisition and reasoning capabilities in complex environments. In addition, safety-oriented benchmarks such as R-Judge~\cite{yuanRJudgeBenchmarkingSafety2024} further investigate potential risks and failure cases arising during agent interactions.
Overall, agentic workflow benchmarks further extend hallucination evaluation from individual model outputs to complex execution processes involving retrieval, reasoning, planning, and tool usage. They provide an important foundation for evaluating the reliability of future agentic systems.

%% file: Tables/Benchemarks.tex
\begin{table}[t!]
\renewcommand{\arraystretch}{1.3}
\centering
\caption{An Overview of Representative Hallucination Benchmarks}
\label{tab:Benchmarks}
\small
\begin{tabular}{@{}
  >{\centering\arraybackslash}p{0.13\textwidth}
  >{\centering\arraybackslash}p{0.18\textwidth}
  >{\centering\arraybackslash}p{0.08\textwidth}
  >{\centering\arraybackslash}p{0.16\textwidth}
  >{\centering\arraybackslash}p{0.17\textwidth}
  >{\centering\arraybackslash}p{0.15\textwidth}
  @{}}
\toprule
\textbf{Benchmark} & \textbf{Hallucination Type} &  \textbf{Data Size} & \textbf{Task} & \textbf{Evaluated Capability} & \textbf{Metrics} \\

\midrule
TruthfulQA~\cite{linTruthfulQAMeasuringHow2022} & \makecell[c]{Knowledege-based \\ Hallucination} &  817 & \makecell[c]{General Question \\ Answering}  & \makecell[c]{Intrinsic Knowledge} & \makecell[c]{Accuracy, \\ Human Evaluation} \\
\midrule
MedHallu~\cite{panditMedHalluComprehensiveBenchmark2025} & \makecell[c]{Knowledege-based \\ Hallucination} & 10,000 & \makecell[c]{Medical Question \\ Answering} & RAG & F1 Score \\
\midrule
RAGTruth~\cite{niuRAGTruthHallucinationCorpus2024} & \makecell[c]{Knowledege-based \\ Hallucination} & 18000 & \makecell[c]{QA, Data-to-Text \\ Summarization} & RAG & Human Evaluation \\
\midrule
BIG-bench~\cite{srivastavaImitationGameQuantifying2023} & \makecell[c]{Logic-based \\ Hallucination} & 200 & Logic Reasoning & Reasoning Results & \makecell[c]{Accuracy, \\ F1 Score} \\
\midrule
PrOntoQA~\cite{saparovLanguageModelsAre2023} & \makecell[c]{Logic-based \\ Hallucination} & 40,000 & Logic Reasoning & CoT & Accuracy \\
\midrule
ToolBench~\cite{qinToolLLMFacilitatingLarge2023} & \makecell[c]{Logic-based \\ Hallucination} & 16,464 & API invocation & \makecell[c]{Tool-Augmented \\ Reasoning} & Accuracy\\
\midrule
LogicBench~\cite{parmarLogicBenchSystematicEvaluation2024} & \makecell[c]{Logic-based \\ Hallucination} & N/A & \makecell[c]{Rule-based \\ Reasoning} & Symbolic Reasoning & Accuracy \\
\midrule
AgentBench~\cite{liuAgentBenchEvaluatingLLMs2023} & \makecell[c]{Composite \\ Hallucination} & 17,000 & \makecell[c]{General \\ Interactive Tasks} & Agentic System & Accuracy \\
\midrule
L-MARS~\cite{wangLMARSLegalMultiAgent2025} & \makecell[c]{Composite \\ Hallucination} & 200 & \makecell[c]{ Legal Multi-Turn \\ Question Answering} & Agentic System & Accuracy \\
\midrule
InfoDeepSeek~\cite{xiInfoDeepSeekBenchmarkingAgentic2025} & \makecell[c]{Composite \\ Hallucination} & 245 & \makecell[c]{Web Environment \\ Retrieval} & Agentic System & \makecell[c]{F1 Score, \\ Human Evaluation} \\

\bottomrule
\end{tabular}
\end{table}

%% file: Data/8.Conclusion_and_Challenges.tex
\section{Conclusion and Challenges}
\label{sec:section8}

From the perspective of capability enhancement, this survey systematically reviews hallucination mitigation approaches for LLMs. Unlike existing taxonomies that categorize methods based on technical forms, this survey focuses on how different approaches enhance key capabilities in the LLM generation process and analyzes the relationship between these capability enhancement mechanisms and different types of hallucinations. Specifically, this survey categorizes hallucinations into knowledge-based hallucinations and logic-based hallucinations, and investigates RAG, reasoning enhancement, and their integration within agentic systems from the perspectives of improving knowledge acquisition capabilities and reasoning reliability, respectively. Through the introduction of external knowledge, optimization of reasoning processes, and dynamic interaction mechanisms, these approaches provide important pathways for understanding and mitigating complex hallucinations caused by insufficient knowledge, unreliable reasoning, and the interplay of multiple capability deficiencies.

\subsection{Key Challenges under the Capability Enhancement Framework}

\subsubsection{Knowledge Acquisition Capability: From Information Retrieval to Reliable Knowledge Utilization}

RAG effectively mitigates hallucinations caused by insufficient knowledge in LLMs by introducing external knowledge. However, it also transforms the problem of internal knowledge deficiency into challenges in external knowledge acquisition and utilization. When retrieved information is incorrect, incomplete, or unreliable, external knowledge may still mislead models into generating factually incorrect responses. Furthermore, conflicts between parametric knowledge and external knowledge, multi-source knowledge integration, and evidence reliability evaluation remain insufficiently addressed.
Future development of RAG requires further enhancement of models' ability to evaluate and utilize external information, including evidence assessment, knowledge integration, and coordination between internal and external knowledge, enabling models to select and exploit external knowledge more reliably.

\subsubsection{Reasoning Capability: From Reasoning Process Expansion to Reliable Reasoning}

CoT, tool-augmented reasoning, and symbolic reasoning methods improve the logical reliability of LLMs from perspectives including reasoning process organization, external interaction, and formal constraints, respectively. However, current LLM reasoning still mainly relies on statistical patterns learned from language data and lacks explicit structured world models and intrinsic verification mechanisms. Therefore, even with reasoning processes or external constraints, models may still generate outputs that appear reasonable but contain factual errors or logical inconsistencies.

Furthermore, existing reasoning augmentation methods typically optimize specific reasoning components and lack a unified mechanism that can dynamically assess reasoning reliability and adjust reasoning strategies according to task states. Different reasoning paradigms also face varying challenges in contextual understanding and reliability assessment when mitigating logic-based hallucinations. CoT-based methods can explicitly expose reasoning processes, but their natural language representations remain susceptible to contextual information, implicit assumptions, and pragmatic phenomena, which may lead to either the misclassification of reasonable expressions or the acceptance of incorrect reasoning. Tool-Augmented Reasoning can reduce certain reasoning errors through external computation, retrieval, or verification; however, the effective utilization of tool outputs still depends on the model’s understanding of task contexts and intermediate results. In contrast, Symbolic Reasoning enhances logical consistency through formal constraints but remains limited by potential errors during the conversion from natural language to symbolic representations. Therefore, future research should further integrate external knowledge, contextual understanding, pragmatic analysis, and formal verification mechanisms to develop more adaptive and reliable reasoning augmentation frameworks. Such frameworks should not only improve models’ reasoning capabilities in complex tasks but also avoid misinterpreting reasonable expressions in natural language as logical errors.

\subsubsection{Capability Coordination: From Module Combination to Dynamic Coordination}

Agentic Systems provide new directions for mitigating composite hallucinations in complex tasks by integrating retrieval, reasoning, and feedback mechanisms. However, current agentic systems typically rely on predefined workflows to connect different modules, lacking dynamic coordination capabilities for critical decisions such as when to retrieve information, how to utilize evidence, when to adjust reasoning strategies, and how to correct errors.
Future research should further develop adaptive coordination mechanisms based on task states, enabling more reliable dynamic integration among knowledge acquisition, reasoning, and feedback processes.

\subsection{General Challenges toward Reliable LLMs}

Beyond the limitations of specific capability enhancement mechanisms, hallucination mitigation research also faces cross-paradigm challenges.

First, achieving a balance between reliability improvement and system efficiency remains an important challenge. Existing approaches often rely on additional retrieval, reasoning expansion, and verification procedures. Although these mechanisms improve output reliability, they also introduce increased computational costs and system complexity. Future research should explore more lightweight and adaptive capability enhancement mechanisms that dynamically adjust knowledge acquisition, reasoning, and verification processes according to task requirements.

Second, capability-oriented evaluation frameworks for hallucination mitigation remain underdeveloped. Existing evaluation methods mainly focus on final outputs, while providing limited analysis of intermediate capabilities such as knowledge acquisition, evidence utilization, reasoning organization, and error correction. Future studies should establish more fine-grained capability evaluation frameworks that extend beyond outcome-based assessment toward process-level capability analysis, enabling a more accurate evaluation of the contributions of different technical paradigms to hallucination mitigation.

Future progress in hallucination mitigation should move beyond the optimization of isolated techniques and toward the coordinated enhancement of complementary capabilities, including knowledge acquisition, reliable reasoning, and autonomous error correction. Through a unified capability-oriented framework, LLMs may gradually evolve into trustworthy intelligent systems that can acquire and manage knowledge, reason reliably across complex tasks, and continuously identify and rectify their own errors.

%% file: References-Pro.bib
@inproceedings{azariaInternalStateLLM2023,
 title = {The Internal State of an LLM Knows When It's Lying},
 author = {Azaria, Amos and Mitchell, Tom},
 booktitle = {Findings of the Association for Computational Linguistics: EMNLP 2023},
 year = {2023},
 pages = {967--976}
}

@inproceedings{asaiSELFRAGLEARNINGRETRIEVE2023,
 title = {Self-RAG: Learning to Retrieve, Generate, and Critique through Self-Reflection},
 author = {Asai, Akari and Wu, Zeqiu and Wang, Yizhong and Sil, Avirup and Hajishirzi, Hannaneh},
 booktitle = {Proceedings of the 12th International Conference on Learning Representations (ICLR)},
 year = {2024}
}

@inproceedings{banerjeeLLMsWillAlways2024,
 title = {LLMs Will Always Hallucinate, and We Need to Live with This},
 author = {Banerjee, Sourav and Agarwal, Ayushi and Singla, Saloni},
 booktitle = {Intelligent Systems Conference},
 pages = {624--648},
 year = {2025}
}

@article{burges2006learning,
 title = {Learning to Rank with Nonsmooth Cost Functions},
 author = {Burges, Christopher and Ragno, Robert and Le, Quoc},
 journal = {Advances in Neural Information Processing Systems},
 volume = {19},
 year = {2006}
}

@inproceedings{burgesLearningRankUsing2005,
author = {Burges, Chris and Shaked, Tal and Renshaw, Erin and Lazier, Ari and Deeds, Matt and Hamilton, Nicole and Hullender, Greg},
title = {Learning to rank using gradient descent},
year = {2005},
isbn = {1595931805},
publisher = {Association for Computing Machinery},
address = {New York, NY, USA},
doi = {10.1145/1102351.1102363},
booktitle = {Proceedings of the 22nd International Conference on Machine Learning},
pages = {89–96},
numpages = {8},
location = {Bonn, Germany},
series = {ICML '05}
}

@inproceedings{caiFoRAGFactualityoptimizedRetrieval2024,
author = {Cai, Tianchi and Tan, Zhiwen and Song, Xierui and Sun, Tao and Jiang, Jiyan and Xu, Yunqi and Zhang, Yinger and Gu, Jinjie},
title = {FoRAG: Factuality-optimized Retrieval Augmented Generation for Web-enhanced Long-form Question Answering},
year = {2024},
isbn = {9798400704901},
publisher = {Association for Computing Machinery},
address = {New York, NY, USA},
doi = {10.1145/3637528.3672065},
booktitle = {Proceedings of the 30th ACM SIGKDD Conference on Knowledge Discovery and Data Mining},
pages = {199–210},
numpages = {12},
location = {Barcelona, Spain},
series = {KDD '24}
}

@inproceedings{caoEditingFactualKnowledge2021,
 title = "Editing Factual Knowledge in Language Models",
 author = "De Cao, Nicola and
 Aziz, Wilker and
 Titov, Ivan",
 editor = "Moens, Marie-Francine and
 Huang, Xuanjing and
 Specia, Lucia and
 Yih, Scott Wen-tau",
 booktitle = "Proceedings of the 2021 Conference on Empirical Methods in Natural Language Processing",
 month = nov,
 year = "2021",
 address = "Online and Punta Cana, Dominican Republic",
 publisher = "Association for Computational Linguistics",
 doi = "10.18653/v1/2021.emnlp-main.522",
 pages = "6491--6506"

}

@inproceedings{chenAdvancingToolAugmentedLarge2024,
 author = {Chen, Sijia and Wang, Yibo and Wu, Yi-Feng and Chen, Qing-Guo and Xu, Zhao and Luo, Weihua and Zhang, Kaifu and Zhang, Lijun},
 booktitle = {Advances in Neural Information Processing Systems},
 editor = {A. Globerson and L. Mackey and D. Belgrave and A. Fan and U. Paquet and J. Tomczak and C. Zhang},
 pages = {106555--106581},
 publisher = {Curran Associates, Inc.},
 title = {Advancing Tool-Augmented Large Language Models: Integrating Insights from Errors in Inference Trees},
 volume = {37},
 year = {2024}
}

@inproceedings{chenChatCoTToolAugmentedChainofThought2023,
 title = {ChatCoT: Tool-Augmented Chain-of-Thought Reasoning on Chat-based Large Language Models},
 author = {Chen, Zhipeng and Zhou, Kun and Zhang, Beichen and Gong, Zheng and Zhao, Xin and Wen, Ji-Rong},
 booktitle = {Findings of the Association for Computational Linguistics: EMNLP 2023},
 year = {2023},
 pages = {14777--14790},
 doi = {10.18653/v1/2023.findings-emnlp.985}
}

@inproceedings{chengEmpoweringLLMsLogical2025,
 title = {Empowering LLMs with Logical Reasoning: A Comprehensive Survey},
 author = {Cheng, Fengxiang and Li, Haoxuan and Liu, Fenrong and van Rooij, Robert and Zhang, Kun and Lin, Zhouchen},
 booktitle = {Proceedings of the 34th International Joint Conference on Artificial Intelligence (IJCAI)},
 year = {2025}
}

@inproceedings{chuangDoLaDecodingContrasting2024,
 title = {DoLa: Decoding by Contrasting Layers Improves Factuality in Large Language Models},
 author = {Chuang, Yung-Sung and Xie, Yujia and Luo, Hongyin and Kim, Yoon and Glass, James R. and He, Pengcheng},
 booktitle = {The Twelfth International Conference on Learning Representations (ICLR)},
 year = {2024}
}

@article{dahlLargeLegalFictions2024,
 title = {Large Legal Fictions: Profiling Legal Hallucinations in Large Language Models},
 author = {Henderson, Peter and Krass, Mark and Zheng, Lucia and Ho, Daniel E.},
 journal = {Journal of Legal Analysis},
 volume = {16},
 number = {1},
 pages = {64--93},
 year = {2024},
 doi = {10.1093/jla/laae003}
}

@inproceedings{devlinBERTPretrainingDeep2019,
 title = {BERT: Pre-training of Deep Bidirectional Transformers for Language Understanding},
 author = {Devlin, Jacob and Chang, Ming-Wei and Lee, Kenton and Toutanova, Kristina},
 booktitle = {Proceedings of the 2019 Conference of the North American Chapter of the Association for Computational Linguistics: Human Language Technologies (NAACL-HLT)},
 pages = {4171--4186},
 year = {2019},
 doi = {10.18653/v1/N19-1423}
}

@inproceedings{fanSurveyRAGMeeting2024,
 title = {A Survey on RAG Meeting LLMs: Towards Retrieval-Augmented Large Language Models},
 author = {Fan, Wenqi and Ding, Yujuan and Ning, Liangbo and Wang, Shijie and Li, Hengyun and Yin, Dawei and Chua, Tat-Seng and Li, Qing},
 booktitle = {Proceedings of the 30th ACM SIGKDD Conference on Knowledge Discovery and Data Mining (KDD)},
 pages = {6491--6501},
 year = {2024},
 doi = {10.1145/3637528.3671470}
}

@inproceedings{fangKiRAGKnowledgeDrivenIterative2025,
 title = {KiRAG: Knowledge-Driven Iterative Retriever for Enhancing Retrieval-Augmented Generation},
 author = {Fang, Jinyuan and Meng, Zaiqiao and Macdonald, Craig},
 booktitle = {Proceedings of the 63rd Annual Meeting of the Association for Computational Linguistics (ACL)},
 pages = {18969--18985},
 year = {2025},
 doi = {10.18653/v1/2025.acl-long.929}
}

@inproceedings{formalSPLADESparseLexical2021,
 title = {SPLADE: Sparse Lexical and Expansion Model for First Stage Ranking},
 author = {Formal, Thibault and Piwowarski, Benjamin and Clinchant, Stéphane},
 booktitle = {Proceedings of the 44th International ACM SIGIR Conference on Research and Development in Information Retrieval (SIGIR)},
 pages = {2288--2292},
 year = {2021},
 doi = {10.1145/3404835.3463098}
}

@inproceedings{fuREASONINGSELFDOUBTMORE2025,
 title = {Reasoning Without Self-Doubt: More Efficient Chain-of-Thought Through Certainty Probing},
 author = {Fu, Yichao and Chen, Junda and Zhuang, Yonghao and Fu, Zheyu and Stoica, Ion and Zhang, Hao},
 booktitle = {ICLR 2025 Workshop on Foundation Models in the Wild},
 year = {2025}
}

@inproceedings{gaoEfficientToolUse2025,
 title = {Efficient Tool Use with Chain-of-Abstraction Reasoning},
 author = {Gao, Silin and Dwivedi-Yu, Jane and Yu, Ping and Tan, Xiaoqing Ellen and Pasunuru, Ramakanth and Golovneva, Olga and Sinha, Koustuv and Celikyilmaz, Asli and Bosselut, Antoine and Wang, Tianlu},
 booktitle = {Proceedings of the 31st International Conference on Computational Linguistics (COLING)},
 pages = {2727--2743},
 year = {2025}
}

@inproceedings{gehrmannGLTRStatisticalDetection2019,
 title = {GLTR: Statistical Detection and Visualization of Generated Text},
 author = {Gehrmann, Sebastian and Strobelt, Hendrik and Rush, Alexander},
 booktitle = {Proceedings of the 57th Annual Meeting of the Association for Computational Linguistics: System Demonstrations (ACL)},
 pages = {111--116},
 year = {2019},
 doi = {10.18653/v1/P19-3019}
}

@inproceedings{guttikondaExplainableAIRetrievalAugmented2025,
 title = {Explainable AI: A Retrieval-Augmented Generation Based Framework for Model Interpretability},
 author = {Tariq, Usman and Shah, Muhammad Abdullah and Khan, Adeel},
 booktitle = {Proceedings of the 17th International Conference on Agents and Artificial Intelligence (ICAART)},
 pages = {948--955},
 year = {2025}
}

@inproceedings{huCGRAGResearchQuestion2025,
 title = {CG-RAG: Research Question Answering by Citation Graph Retrieval-Augmented LLMs},
 author = {Hu, Yuntong and Lei, Zhihan and Dai, Zhongjie and Zhang, Allen and Angirekula, Abhinav and Zhang, Zheng and Zhao, Liang},
 booktitle = {Proceedings of SIGIR 2025},
 pages = {678--687},
 year = {2025},
 doi = {10.1145/3726302.3729920}
}

@inproceedings{huangImprovingContextualFaithfulness2025,
 title = {Improving Contextual Faithfulness of Large Language Models via Retrieval Heads-Induced Optimization},
 author = {Huang, Lei and Feng, Xiaocheng and Ma, Weitao and Fan, Yuchun and Feng, Xiachong and Ye, Yangfan and Zhong, Weihong and Gu, Yuxuan and Wang, Baoxin and Wu, Dayong and Hu, Guoping and Qin, Bing},
 booktitle = {Proceedings of ACL 2025},
 pages = {16896--16913},
 year = {2025},
 doi = {10.18653/v1/2025.acl-long.826}
}

@article{huangSurveyHallucinationLarge2025,
 title = {A Survey on Hallucination in Large Language Models: Principles, Taxonomy, Challenges, and Open Questions},
 author = {Huang, Lei and Yu, Weijiang and Ma, Weitao and Zhong, Weihong and Feng, Zhangyin and Wang, Haotian and Chen, Qianglong and Peng, Weihua and Feng, Xiaocheng and Qin, Bing and Liu, Ting},
 journal = {ACM Transactions on Information Systems},
 volume = {43},
 number = {2},
 articleno = {42},
 pages = {1--55},
 year = {2025},
 doi = {10.1145/3703155}
}

@inproceedings{huangKETRAGCostEfficientMultiGranular2025,
 title = {KET-RAG: A Cost-Efficient Multi-Granular Indexing Framework for Graph-RAG},
 author = {Huang, Yiqian and Zhang, Shiqi and Xiao, Xiaokui},
 booktitle = {Proceedings of KDD 2025},
 pages = {1003--1012},
 year = {2025},
 doi = {10.1145/3711896.3737012}
}

@inproceedings{hwangDSLRDocumentRefinement2024,
 title = {DSLR: Document Refinement with Sentence-Level Re-ranking and Reconstruction to Enhance Retrieval-Augmented Generation},
 author = {Hwang, Taeho and Jeong, Soyeong and Cho, Sukmin and Han, SeungYoon and Park, Jong},
 booktitle = {Proceedings of the 3rd Workshop on Knowledge Augmented Methods for NLP},
 pages = {73--92},
 year = {2024},
 doi = {10.18653/v1/2024.knowledgenlp-1.6}
}

@article{izacardUnsupervisedDenseInformation2022,
 title = {Unsupervised Dense Information Retrieval with Contrastive Learning},
 author = {Izacard, Gautier and Caron, Mathilde and Hosseini, Lucas and Riedel, Sebastian and Bojanowski, Piotr and Joulin, Armand and Grave, Edouard},
 journal = {Transactions on Machine Learning Research},
 year = {2022}
}

@article{jiSurveyKnowledgeGraphs2021,
 title = {A Survey on Knowledge Graphs: Representation, Acquisition, and Applications},
 author = {Ji, Shaoxiong and Pan, Shirui and Cambria, Erik and Marttinen, Pekka and Yu, Philip S.},
 journal = {IEEE Transactions on Neural Networks and Learning Systems},
 volume = {33},
 number = {2},
 pages = {494--514},
 year = {2022},
 doi = {10.1109/TNNLS.2021.3070843}
}

@article{jiSurveyHallucinationNatural2023,
 title = {Survey of Hallucination in Natural Language Generation},
 author = {Ji, Ziwei and Lee, Nayeon and Frieske, Rita and Yu, Tiezheng and Su, Dan and Xu, Yan and Ishii, Etsuko and Bang, Ye Jin and Madotto, Andrea and Fung, Pascale},
 journal = {ACM Computing Surveys},
 volume = {55},
 number = {12},
 articleno = {248},
 pages = {1--38},
 year = {2023},
 doi = {10.1145/3571730}
}

@inproceedings{jiangLongLLMLinguaAcceleratingEnhancing2024,
 title = {LongLLMLingua: Accelerating and Enhancing LLMs in Long Context Scenarios via Prompt Compression},
 author = {Jiang, Huiqiang and Wu, Qianhui and Luo, Xufang and Li, Dongsheng and Lin, Chin-Yew and Yang, Yuqing and Qiu, Lili},
 booktitle = {Proceedings of ACL 2024},
 pages = {1658--1677},
 year = {2024},
 doi = {10.18653/v1/2024.acl-long.91}
}

@inproceedings{kalraHyPARAGHybridParameter2025,
 title = {HyPA-RAG: A Hybrid Parameter Adaptive Retrieval-Augmented Generation System for AI Legal and Policy Applications},
 author = {Kalra, Rishi and Wu, Zekun and Gulley, Ayesha and Hilliard, Airlie and Guan, Xin and Koshiyama, Adriano and Treleaven, Philip Colin},
 booktitle = {Proceedings of the 1st Workshop on Customizable NLP (CustomNLP4U)},
 pages = {237--256},
 year = {2024},
 doi = {10.18653/v1/2024.customnlp4u-1.18}
}

@inproceedings{karpukhinDensePassageRetrieval2020,
 title = {Dense Passage Retrieval for Open-Domain Question Answering},
 author = {Karpukhin, Vladimir and Oguz, Barlas and Min, Sewon and Lewis, Patrick and Wu, Ledell and Edunov, Sergey and Chen, Danqi and Yih, Wen-tau},
 booktitle = {Proceedings of EMNLP 2020},
 pages = {6769--6781},
 year = {2020},
 doi = {10.18653/v1/2020.emnlp-main.550}
}

@inproceedings{kimRERAGImprovingOpenDomain2024,
 title = {RE-RAG: Improving Open-Domain QA Performance and Interpretability with Relevance Estimator in Retrieval-Augmented Generation},
 author = {Kim, Kiseung and Lee, Jay-Yoon},
 booktitle = {Proceedings of EMNLP 2024},
 pages = {22149--22161},
 year = {2024},
 doi = {10.18653/v1/2024.emnlp-main.1236}
}

@inproceedings{kojimaLargeLanguageModels2023,
 title = {Large Language Models are Zero-Shot Reasoners},
 author = {Kojima, Takeshi and Gu, Shixiang and Reid, Machel and Matsuo, Yutaka and Iwasawa, Yusuke},
 booktitle = {Proceedings of NeurIPS 2022},
 pages = {22199--22213},
 volume = {35},
 year = {2022}
}

@inproceedings{leeHumanInspiredReadingAgent2024,
 title = {A Human-Inspired Reading Agent with Gist Memory of Very Long Contexts},
 author = {Lee, Kuang-Huei and Chen, Xinyun and Furuta, Hiroki and Canny, John and Fischer, Ian},
 booktitle = {Proceedings of ICML 2024},
 pages = {1054--1074},
 year = {2024}
}

@inproceedings{leeFactualityEnhancedLanguage2023,
 title = {Factuality Enhanced Language Models for Open-Ended Text Generation},
 author = {Lee, Nayeon and Ping, Wei and Xu, Peng and Patwary, Mostofa and Fung, Pascale N and Shoeybi, Mohammad and Catanzaro, Bryan},
 booktitle = {Proceedings of NeurIPS 2022},
 pages = {34586--34599},
 volume = {35},
 year = {2022}
}

@inproceedings{lewisRetrievalAugmentedGenerationKnowledgeIntensive2021,
 title = {Retrieval-Augmented Generation for Knowledge-Intensive NLP Tasks},
 author = {Lewis, Patrick and Perez, Ethan and Piktus, Aleksandra and Petroni, Fabio and Karpukhin, Vladimir and Goyal, Naman and Küttler, Heinrich and Lewis, Mike and Yih, Wen-tau and Rocktäschel, Tim and Riedel, Sebastian and Kiela, Douwe},
 booktitle = {Proceedings of NeurIPS 2020},
 pages = {9459--9474},
 volume = {33},
 year = {2020}
}

@inproceedings{liDALKDynamicCoAugmentation2024,
 title = {DALK: Dynamic Co-Augmentation of LLMs and Knowledge Graphs to Answer Alzheimer's Disease Questions with Scientific Literature},
 author = {Li, Dawei and Yang, Shu and Tan, Zhen and Baik, Jae Young and Yun, Sukwon and Lee, Joseph and Chacko, Aaron and Hou, Bojian and Duong-Tran, Duy and Ding, Ying and Liu, Huan and Shen, Li and Chen, Tianlong},
 booktitle = {Findings of EMNLP 2024},
 pages = {2187--2205},
 year = {2024},
 doi = {10.18653/v1/2024.findings-emnlp.119}
}

@article{liEmpoweringMoleculeDiscovery2024,
 title = {Empowering Molecule Discovery for Molecule-Caption Translation with Large Language Models: A ChatGPT Perspective},
 author = {Li, Jiatong and Liu, Yunqing and Fan, Wenqi and Wei, Xiao-Yong and Liu, Hui and Tang, Jiliang and Li, Qing},
 journal = {IEEE Transactions on Knowledge and Data Engineering},
 volume = {36},
 number = {11},
 pages = {6071--6083},
 year = {2024},
 doi = {10.1109/TKDE.2024.3393356}
}

@inproceedings{liAPIBankComprehensiveBenchmark2023,
 title = {API-Bank: A Comprehensive Benchmark for Tool-Augmented LLMs},
 author = {Li, Minghao and Zhao, Yingxiu and Yu, Bowen and Song, Feifan and Li, Hangyu and Yu, Haiyang and Li, Zhoujun and Huang, Fei and Li, Yongbin},
 booktitle = {Proceedings of EMNLP 2023},
 pages = {3102--3116},
 year = {2023},
 doi = {10.18653/v1/2023.emnlp-main.187}
}

@inproceedings{liGraphReaderBuildingGraphbased2024,
 title = {GraphReader: Building Graph-based Agent to Enhance Long-Context Abilities of Large Language Models},
 author = {Li, Shilong and He, Yancheng and Guo, Hangyu and Bu, Xingyuan and Bai, Ge and Liu, Jie and Liu, Jiaheng and Qu, Xingwei and Li, Yangguang and Ouyang, Wanli and Su, Wenbo and Zheng, Bo},
 booktitle = {Findings of EMNLP 2024},
 pages = {12758--12786},
 year = {2024},
 doi = {10.18653/v1/2024.findings-emnlp.746}
}

@inproceedings{liContrastiveDecodingOpenended2023,
 title = {Contrastive Decoding: Open-ended Text Generation as Optimization},
 author = {Li, Xiang Lisa and Holtzman, Ari and Fried, Daniel and Liang, Percy and Eisner, Jason and Hashimoto, Tatsunori and Zettlemoyer, Luke and Lewis, Mike},
 booktitle = {Proceedings of ACL 2023},
 pages = {12286--12312},
 year = {2023},
 doi = {10.18653/v1/2023.acl-long.687}
}

@article{liBatGPTBidirectionalAutoregessive2023,
 title = {BatGPT: A Bidirectional Autoregressive Talker from Generative Pre-trained Transformer},
 author = {Li, Zuchao and Zhang, Shitou and Zhao, Hai and Yang, Yifei and Yang, Dongjie},
 journal = {arXiv preprint arXiv:2307.00360},
 year = {2023}
}

@inproceedings{linTruthfulQAMeasuringHow2022,
 title = {TruthfulQA: Measuring How Models Mimic Human Falsehoods},
 author = {Lin, Stephanie and Hilton, Jacob and Evans, Owain},
 booktitle = {Proceedings of ACL 2022},
 pages = {3214--3252},
 year = {2022},
 doi = {10.18653/v1/2022.acl-long.229}
}

@inproceedings{lingDeductiveVerificationChainofThought2023,
 title = {Deductive Verification of Chain-of-Thought Reasoning},
 author = {Ling, Zhan and Fang, Yunhao and Li, Xuanlin and Huang, Zhiao and Lee, Mingu and Memisevic, Roland and Su, Hao},
 booktitle = {Proceedings of NeurIPS 2023},
 pages = {36407--36433},
 volume = {36},
 year = {2023}
}

@inproceedings{liuExposingAttentionGlitches2023,
 title = {Exposing Attention Glitches with Flip-Flop Language Modeling},
 author = {Liu, Bingbin and Ash, Jordan and Goel, Surbhi and Krishnamurthy, Akshay and Zhang, Cyril},
 booktitle = {Proceedings of NeurIPS 2023},
 pages = {25549--25583},
 volume = {36},
 year = {2023}
}

@inproceedings{liuLogiQAChallengeDataset2020,
 title = {LogiQA: A Challenge Dataset for Machine Reading Comprehension with Logical Reasoning},
 author = {Liu, Jian and Cui, Leyang and Liu, Hanmeng and Huang, Dandan and Wang, Yile and Zhang, Yue},
 booktitle = {Proceedings of IJCAI 2021},
 pages = {501--507},
 year = {2021}
}

@inproceedings{liuLARALinguisticAdaptiveRetrievalAugmentation2024,
 title = {LARA: Linguistic-Adaptive Retrieval-Augmentation for Multi-Turn Intent Classification},
 author = {Liu, Junhua and Tan, Yong Keat and Fu, Bin and Lim, Kwan Hui},
 booktitle = {Proceedings of EMNLP 2024 (Industry Track)},
 pages = {1096--1106},
 year = {2024},
 doi = {10.18653/v1/2024.emnlp-industry.82}
}

@inproceedings{liuMindYourStep2025,
 title = {Mind Your Step (by Step): Chain-of-Thought Can Reduce Performance on Tasks Where Thinking Makes Humans Worse},
 author = {Liu, Ryan and Geng, Jiayi and Wu, Addison J. and Sucholutsky, Ilia and Lombrozo, Tania and Griffiths, Thomas L.},
 booktitle = {Proceedings of ICML 2025},
 year = {2025}
}

@article{liu2009learning,
 title = {Learning to Rank for Information Retrieval},
 author = {Liu, Tie-Yan},
 journal = {Foundations and Trends in Information Retrieval},
 volume = {3},
 number = {3},
 pages = {225--331},
 year = {2009},
 doi = {10.1561/1500000016}
}

@inproceedings{liuWebGLMEfficientWebEnhanced2023,
 title = {WebGLM: Towards an Efficient Web-Enhanced Question Answering System with Human Preferences},
 author = {Liu, Xiao and Lai, Hanyu and Yu, Hao and Xu, Yifan and Zeng, Aohan and Du, Zhengxiao and Zhang, Peng and Dong, Yuxiao and Tang, Jie},
 booktitle = {Proceedings of KDD 2023},
 pages = {4549--4560},
 year = {2023},
 doi = {10.1145/3580305.3599931}
}

@inproceedings{liuAgentBenchEvaluatingLLMs2023,
 title = {AgentBench: Evaluating LLMs as Agents},
 author = {Liu, Xiao and Yu, Hao and Zhang, Hanchen and Xu, Yifan and Lei, Xuanyu and Lai, Hanyu and Gu, Yu and Ding, Hangliang and Men, Kaiwen and Yang, Kejuan and Zhang, Shudan and Deng, Xiang and Zeng, Aohan and Du, Zhengxiao and Zhang, Chenhui and Shen, Sheng and Zhang, Tianjun and Su, Yu and Sun, Huan and Huang, Minlie and Dong, Yuxiao and Tang, Jie},
 booktitle = {Proceedings of ICLR 2024},
 year = {2024}
}

@inproceedings{liuRAISFLearningAnswer2024,
 title = {RA-ISF: Learning to Answer and Understand from Retrieval Augmentation via Iterative Self-Feedback},
 author = {Liu, Yanming and Peng, Xinyue and Zhang, Xuhong and Liu, Weihao and Yin, Jianwei and Cao, Jiannan and Du, Tianyu},
 booktitle = {Findings of ACL 2024},
 pages = {4730--4749},
 year = {2024},
 doi = {10.18653/v1/2024.findings-acl.281}
}

@inproceedings{luToolFiVeEnhancingToolAugmented2025,
 title = {ToolFiVe: Enhancing Tool-Augmented LLMs via Tool Filtering and Verification},
 author = {Lu, Hailun and Li, Xingming and Ji, Xuanyu and Kan, Zhigang and Hu, Qingyong},
 booktitle = {Proceedings of ICASSP 2025},
 pages = {1--5},
 year = {2025},
 doi = {10.1109/ICASSP49660.2025.10887544}
}

@inproceedings{luTARTOpenSourceToolAugmented2025,
 title = {TART: An Open-Source Tool-Augmented Framework for Explainable Table-Based Reasoning},
 author = {Lu, Xinyuan and Pan, Liangming and Ma, Yubo and Nakov, Preslav and Kan, Min-Yen},
 booktitle = {Findings of NAACL 2025},
 pages = {4323--4339},
 year = {2025},
 doi = {10.18653/v1/2025.findings-naacl.244}
}

@inproceedings{lyuFaithfulChainofThoughtReasoning2023,
 title = {Faithful Chain-of-Thought Reasoning},
 author = {Lyu, Qing and Havaldar, Shreya and Stein, Adam and Zhang, Li and Rao, Delip and Wong, Eric and Apidianaki, Marianna and Callison-Burch, Chris},
 booktitle = {Proceedings of IJCNLP 2023},
 pages = {305--329},
 year = {2023},
 doi = {10.18653/v1/2023.ijcnlp-main.20}
}

@inproceedings{maThinkonGraph20Deep2025,
 title = {Think-on-Graph 2.0: Deep and Faithful Large Language Model Reasoning with Knowledge-Guided Retrieval Augmented Generation},
 author = {Ma, Shengjie and Xu, Chengjin and Jiang, Xuhui and Li, Muzhi and Qu, Huaren and Yang, Cehao and Mao, Jiaxin and Guo, Jian},
 booktitle = {Proceedings of ICLR 2025},
 year = {2025}
}

@inproceedings{maQueryRewritingRetrievalAugmented2023,
 title = {Query Rewriting in Retrieval-Augmented Large Language Models},
 author = {Ma, Xinbei and Gong, Yeyun and He, Pengcheng and Zhao, Hai and Duan, Nan},
 booktitle = {Proceedings of EMNLP 2023},
 year = {2023}
}

@inproceedings{maSciAgentToolaugmentedLanguage2024,
 title = {SciAgent: Tool-Augmented Language Models for Scientific Reasoning},
 author = {Ma, Yubo and Gou, Zhibin and Hao, Junheng and Xu, Ruochen and Wang, Shuohang and Pan, Liangming and Yang, Yujiu and Cao, Yixin and Sun, Aixin},
 booktitle = {Proceedings of EMNLP 2024},
 pages = {15701--15736},
 year = {2024},
 doi = {10.18653/v1/2024.emnlp-main.880}
}

@inproceedings{mavromatisGNNRAGGraphNeural2024,
 title = {GNN-RAG: Graph Neural Retrieval for Efficient Large Language Model Reasoning on Knowledge Graphs},
 author = {Mavromatis, Costas and Karypis, George},
 booktitle = {Findings of ACL 2025},
 pages = {16682--16699},
 year = {2025},
 doi = {10.18653/v1/2025.findings-acl.856}
}

@inproceedings{maynezFaithfulnessFactualityAbstractive2020,
 title = {On Faithfulness and Factuality in Abstractive Summarization},
 author = {Maynez, Joshua and Narayan, Shashi and Bohnet, Bernd and McDonald, Ryan},
 booktitle = {Proceedings of ACL 2020},
 pages = {1906--1919},
 year = {2020},
 doi = {10.18653/v1/2020.acl-main.173}
}

@inproceedings{mirzadehGSMSymbolicUnderstandingLimitations2025,
 title = {GSM-Symbolic: Understanding the Limitations of Mathematical Reasoning in Large Language Models},
 author = {Mirzadeh, Seyed Iman and Alizadeh, Keivan and Shahrokhi, Hooman and Tuzel, Oncel and Bengio, Samy and Farajtabar, Mehrdad},
 booktitle = {Proceedings of ICLR 2025},
 year = {2025}
}

@inproceedings{mishraFinegrainedHallucinationDetection2024,
 title = {Fine-Grained Hallucination Detection and Editing for Language Models},
 author = {Mishra, Abhika and Asai, Akari and Balachandran, Vidhisha and Wang, Yizhong and Neubig, Graham and Tsvetkov, Yulia and Hajishirzi, Hannaneh},
 booktitle = {Proceedings of the First Conference on Language Modeling},
 year = {2024}
}

@inproceedings{mitchellDetectGPTZeroShotMachineGenerated2023,
 title = {DetectGPT: Zero-Shot Machine-Generated Text Detection Using Probability Curvature},
 author = {Mitchell, Eric and Lee, Yoonho and Khazatsky, Alexander and Manning, Christopher D. and Finn, Chelsea},
 booktitle = {Proceedings of ICML 2023},
 year = {2023}
}

@inproceedings{mitchellFastModelEditing2022,
 title = {Fast Model Editing at Scale},
 author = {Mitchell, Eric and Lin, Charles and Bosselut, Antoine and Finn, Chelsea and Manning, Christopher D.},
 booktitle = {Proceedings of ICLR 2022},
 year = {2022}
}

@inproceedings{muennighoffS1SimpleTesttime2025,
 title = {s1: Simple Test-Time Scaling},
 author = {Muennighoff, Niklas and Yang, Zitong and Shi, Weijia and Li, Xiang Lisa and Fei-Fei, Li and Hajishirzi, Hannaneh and Zettlemoyer, Luke and Liang, Percy and Candes, Emmanuel and Hashimoto, Tatsunori},
 booktitle = {Workshop on Reasoning and Planning for Large Language Models},
 year = {2025}
}

@article{ouyangLLMBoxChocolates2023,
 title = {An Empirical Study of the Non-Determinism of ChatGPT in Code Generation},
 author = {Ouyang, Shuyin and Zhang, Jie M. and Harman, Mark and Wang, Meng},
 journal = {ACM Transactions on Software Engineering and Methodology},
 volume = {34},
 number = {2},
 articleno = {42},
 pages = {1--28},
 year = {2025},
 doi = {10.1145/3697010}
}

@inproceedings{panLogicLMEmpoweringLarge2023,
 title = {Logic-LM: Empowering Large Language Models with Symbolic Solvers for Faithful Logical Reasoning},
 author = {Pan, Liangming and Albalak, Alon and Wang, Xinyi and Wang, William},
 booktitle = {Findings of EMNLP 2023},
 pages = {3806--3824},
 year = {2023},
 doi = {10.18653/v1/2023.findings-emnlp.248}
}

@inproceedings{panANSWEREDAdaptiveToolAugmented2024,
 title = {ANSWERED: Adaptive Tool-Augmented LLMs with Strategic Error Feedback for Compositional Reasoning},
 author = {Pan, Yuchen and Zhang, Xichuan and Zhang, Haoyu and Liu, Yangyang and Wang, Haiyang and Huang, Qiangjuan and Wang, Zhiyuan},
 booktitle = {Advanced Intelligent Computing Technology and Applications},
 pages = {263--280},
 year = {2024}
}

@inproceedings{parmarLogicBenchSystematicEvaluation2024,
 title = {LogicBench: Towards Systematic Evaluation of Logical Reasoning Ability of Large Language Models},
 author = {Parmar, Mihir and Patel, Nisarg and Varshney, Neeraj and Nakamura, Mutsumi and Luo, Man and Mashetty, Santosh and Mitra, Arindam and Baral, Chitta},
 booktitle = {Proceedings of ACL 2024},
 pages = {13679--13707},
 year = {2024},
 doi = {10.18653/v1/2024.acl-long.739}
}

@inproceedings{paulREFINERReasoningFeedback2024,
 title = {REFINER: Reasoning Feedback on Intermediate Representations},
 author = {Paul, Debjit and Ismayilzada, Mete and Peyrard, Maxime and Borges, Beatriz and Bosselut, Antoine and West, Robert and Faltings, Boi},
 booktitle = {Proceedings of EACL 2024},
 pages = {1100--1126},
 year = {2024},
 doi = {10.18653/v1/2024.eacl-long.67}
}

@inproceedings{qinToolLLMFacilitatingLarge2023,
 title = {ToolLLM: Facilitating Large Language Models to Master 16000+ Real-World APIs},
 author = {Qin, Yujia and Liang, Shihao and Ye, Yining and Zhu, Kunlun and Yan, Lan and Lu, Yaxi and Lin, Yankai and Cong, Xin and Tang, Xiangru and Qian, Bill and Zhao, Sihan and Hong, Lauren and Tian, Runchu and Xie, Ruobing and Zhou, Jie and Li, Dahai and Liu, Zhiyuan and Sun, Maosong and Gerstein, Mark},
 booktitle = {Proceedings of ICLR 2024},
 year = {2024}
}

@inproceedings{rakibmollahDetectionFakeNews2023,
 title = {Detection of Fake News with RoBERTa Based Embedding and Modified Deep Neural Network Architecture},
 author = {Rakib Mollah, Md. Abdur and Kabir, Mir Md. Jahangir and Kabir, Monika and Reza, Md. Sazid},
 booktitle = {Proceedings of ICCIT 2023},
 pages = {1--6},
 year = {2023},
 doi = {10.1109/ICCIT60459.2023.10441206}
}

@article{ramInContextRetrievalAugmentedLanguage2023,
 title = {In-Context Retrieval-Augmented Language Models},
 author = {Ram, Ori and Levine, Yoav and Dalmedigos, Itay and Muhlgay, Dor and Shashua, Amnon and Leyton-Brown, Kevin and Shoham, Yoav},
 journal = {Transactions of the Association for Computational Linguistics},
 volume = {11},
 pages = {1316--1331},
 year = {2023},
 doi = {10.1162/tacl_a_00605}
}

@inproceedings{ranaldiElicitingCriticalReasoning2024,
 title = {Eliciting Critical Reasoning in Retrieval-Augmented Generation via Contrastive Explanations},
 author = {Ranaldi, Leonardo and Valentino, Marco and Freitas, Andre},
 booktitle = {Proceedings of NAACL 2025},
 pages = {11168--11183},
 year = {2025},
 doi = {10.18653/v1/2025.naacl-long.557}
}

@article{sahaYouBelieveYour2025,
 title = {You Believe Your LLM Is Not Delusional? Think Again! A Study of LLM Hallucination on Foundation Models under Perturbation},
 author = {Saha, Anirban and Gupta, Binay and Chatterjee, Anirban and Banerjee, Kunal},
 journal = {Discover Data},
 volume = {3},
 articleno = {20},
 year = {2025},
 doi = {10.1007/s44248-025-00041-7}
}

@inproceedings{sahooComprehensiveSurveyHallucination2024,
 title = {A Comprehensive Survey of Hallucination in Large Language, Image, Video and Audio Foundation Models},
 author = {Sahoo, Pranab and Meharia, Prabhash and Ghosh, Akash and Saha, Sriparna and Jain, Vinija and Chadha, Aman},
 booktitle = {Findings of EMNLP 2024},
 pages = {11709--11724},
 year = {2024},
 doi = {10.18653/v1/2024.findings-emnlp.685}
}

@article{saltonTermweightingApproachesAutomatic1988,
 title = {Term-Weighting Approaches in Automatic Text Retrieval},
 author = {Salton, Gerard and Buckley, Chris},
 journal = {Information Processing \& Management},
 volume = {24},
 number = {5},
 pages = {513--523},
 year = {1988},
 doi = {10.1016/0306-4573(88)90021-0}
}

@inproceedings{santhanamColBERTv2EffectiveEfficient2022,
 title = {ColBERTv2: Effective and Efficient Retrieval via Lightweight Late Interaction},
 author = {Santhanam, Keshav and Khattab, Omar and Saad-Falcon, Jon and Potts, Christopher and Zaharia, Matei},
 booktitle = {Proceedings of NAACL 2022},
 pages = {3715--3734},
 year = {2022},
 doi = {10.18653/v1/2022.naacl-main.272}
}

@inproceedings{saparovLanguageModelsAre2023,
 title = {Language Models Are Greedy Reasoners: A Systematic Formal Analysis of Chain-of-Thought},
 author = {Saparov, Abulhair and He, He},
 booktitle = {Proceedings of ICLR 2023},
 year = {2023}
}

@inproceedings{sarmahHybridRAGIntegratingKnowledge2024,
 title = {HybridRAG: Integrating Knowledge Graphs and Vector Retrieval Augmented Generation for Efficient Information Extraction},
 author = {Sarmah, Bhaskarjit and Mehta, Dhagash and Hall, Benika and Rao, Rohan and Patel, Sunil and Pasquali, Stefano},
 booktitle = {Proceedings of ICAIF 2024},
 pages = {608--616},
 year = {2024},
 doi = {10.1145/3677052.3698671}
}

@inproceedings{sawarkarBlendedRAGImproving2024,
 title = {Blended RAG: Improving Retriever-Augmented Generation Accuracy with Semantic Search and Hybrid Query-Based Retrievers},
 author = {Sawarkar, Kunal and Mangal, Abhilasha and Solanki, Shivam Raj},
 booktitle = {Proceedings of MIPR 2024},
 pages = {155--161},
 year = {2024},
 doi = {10.1109/MIPR62202.2024.00031}
}

@inproceedings{schickToolformerLanguageModels2023,
 title = {Toolformer: Language Models Can Teach Themselves to Use Tools},
 author = {Schick, Timo and Dwivedi-Yu, Jane and Dessi, Roberto and Raileanu, Roberta and Lomeli, Maria and Hambro, Eric and Zettlemoyer, Luke and Cancedda, Nicola and Scialom, Thomas},
 booktitle = {Advances in Neural Information Processing Systems},
 volume = {36},
 pages = {68539--68551},
 year = {2023}
}

@article{shumailovAIModelsCollapse2024,
 title = {AI Models Collapse When Trained on Recursively Generated Data},
 author = {Shumailov, Ilia and Shumaylov, Zakhar and Zhao, Yiren and Papernot, Nicolas and Anderson, Ross and Gal, Yarin},
 journal = {Nature},
 volume = {631},
 number = {8022},
 pages = {755--759},
 year = {2024},
 doi = {10.1038/s41586-024-07566-y}
}

@article{srivastavaImitationGameQuantifying2023,
 title = {Beyond the Imitation Game: Quantifying and Extrapolating the Capabilities of Language Models},
 author = {Srivastava, Aarohi and Rastogi, Abhinav and Rao, Abhishek and Shoeb, Abu Awal Md and Abid, Abubakar and Fisch, Adam and Brown, Adam R. and Santoro, Adam and Gupta, Aditya and Garriga-Alonso, Adrià and others},
 journal = {Transactions on Machine Learning Research},
 year = {2023}
}

@inproceedings{sunThinkonGraphDeepResponsible2024,
 title = {Think-on-Graph: Deep and Responsible Reasoning of Large Language Models on Knowledge Graphs},
 author = {Sun, Jiashuo and Xu, Chengjin and Tang, Lumingyuan and Wang, Saizhuo and Lin, Chen and Gong, Yeyun and Ni, Lionel and Shum, Heung-Yeung and Guo, Jian},
 booktitle = {Proceedings of ICLR 2024},
 year = {2024}
}

@inproceedings{tafjordProofWriterGeneratingImplications2021,
 title = {ProofWriter: Generating Implications, Proofs, and Abductive Statements over Natural Language},
 author = {Tafjord, Oyvind and Dalvi Mishra, Bhavana and Clark, Peter},
 booktitle = {Findings of ACL–IJCNLP 2021},
 pages = {3621--3634},
 year = {2021},
 doi = {10.18653/v1/2021.findings-acl.317}
}

@inproceedings{tanHtmlRAGHTMLBetter2025,
 title = {HTMLRAG: HTML Is Better Than Plain Text for Modeling Retrieved Knowledge in RAG Systems},
 author = {Tan, Jiejun and Dou, Zhicheng and Wang, Wen and Wang, Mang and Chen, Weipeng and Wen, Ji-Rong},
 booktitle = {Proceedings of WWW 2025},
 pages = {1733--1746},
 year = {2025},
 doi = {10.1145/3696410.3714546}
}

@inproceedings{tanSmallModelsBig2024,
 title = {Small Models, Big Insights: Leveraging Slim Proxy Models to Decide When and What to Retrieve for LLMs},
 author = {Tan, Jiejun and Dou, Zhicheng and Zhu, Yutao and Guo, Peidong and Fang, Kun and Wen, Ji-Rong},
 booktitle = {Proceedings of ACL 2024},
 pages = {4420--4436},
 year = {2024},
 doi = {10.18653/v1/2024.acl-long.242}
}

@inproceedings{tangMitigatingHallucinatedTranslations2025,
 title = {Mitigating Hallucinated Translations in Large Language Models with Hallucination-Focused Preference Optimization},
 author = {Tang, Zilu and Chatterjee, Rajen and Garg, Sarthak},
 booktitle = {Proceedings of NAACL 2025},
 pages = {3410--3433},
 year = {2025},
 doi = {10.18653/v1/2025.naacl-long.175}
}

@inproceedings{trivediInterleavingRetrievalChainofThought2023,
 title = {Interleaving Retrieval with Chain-of-Thought Reasoning for Knowledge-Intensive Multi-Step Questions},
 author = {Trivedi, Harsh and Balasubramanian, Niranjan and Khot, Tushar and Sabharwal, Ashish},
 booktitle = {Proceedings of ACL 2023},
 pages = {10014--10037},
 year = {2023},
 doi = {10.18653/v1/2023.acl-long.557}
}

@inproceedings{vaswaniAttentionAllYou,
 title = {Attention Is All You Need},
 author = {Vaswani, Ashish and Shazeer, Noam and Parmar, Niki and Uszkoreit, Jakob and Jones, Llion and Gomez, Aidan N. and Kaiser, Lukasz and Polosukhin, Illia},
 booktitle = {Advances in Neural Information Processing Systems},
 pages = {5998--6008},
 year = {2017}
}

@inproceedings{vuFreshLLMsRefreshingLarge2023,
 title = {FreshLLMs: Refreshing Large Language Models with Search Engine Augmentation},
 author = {Vu, Tu and Iyyer, Mohit and Wang, Xuezhi and Constant, Noah and Wei, Jerry and Wei, Jason and Tar, Chris and Sung, Yun-Hsuan and Zhou, Denny and Le, Quoc and Luong, Thang},
 booktitle = {Findings of ACL 2024},
 pages = {13697--13720},
 year = {2024},
 doi = {10.18653/v1/2024.findings-acl.813}
}

@inproceedings{wangDynamicChainofThoughtAdaptive2025,
 title = {Dynamic Chain-of-Thought: Towards Adaptive Deep Reasoning},
 author = {Wang, Libo},
 booktitle = {Proceedings of Greeks in AI Symposium 2025},
 year = {2025}
}

@inproceedings{wangSelfConsistencyImprovesChain2023,
 title = {Self-Consistency Improves Chain of Thought Reasoning in Language Models},
 author = {Wang, Xuezhi and Wei, Jason and Schuurmans, Dale and Le, Quoc V. and Chi, Ed H. and Narang, Sharan and Chowdhery, Aakanksha and Zhou, Denny},
 booktitle = {Proceedings of ICLR 2023},
 year = {2023}
}

@inproceedings{wangChatLogicIntegratingLogic2024,
 title = {ChatLogic: Integrating Logic Programming with Large Language Models for Multi-Step Reasoning},
 author = {Wang, Zhongsheng and Liu, Jiamou and Bao, Qiming and Rong, Hongfei and Zhang, Jingfeng},
 booktitle = {Proceedings of IJCNN 2024},
 pages = {1--8},
 year = {2024},
 doi = {10.1109/IJCNN60899.2024.10650138}
}

@article{weiEmergentAbilitiesLarge2022,
 title = {Emergent Abilities of Large Language Models},
 author = {Wei, Jason and Tay, Yi and Bommasani, Rishi and Raffel, Colin and Zoph, Barret and Borgeaud, Sebastian and Yogatama, Dani and Bosma, Maarten and Zhou, Denny and Metzler, Donald and Chi, Ed H. and Hashimoto, Tatsunori and Vinyals, Oriol and Liang, Percy and Dean, Jeff and Fedus, William},
 journal = {Transactions on Machine Learning Research},
 year = {2022}
}

@inproceedings{weiChainofThoughtPromptingElicits2023,
 title = {Chain-of-Thought Prompting Elicits Reasoning in Large Language Models},
 author = {Wei, Jason and Wang, Xuezhi and Schuurmans, Dale and Bosma, Maarten and Ichter, Brian and Xia, Fei and Chi, Ed and Le, Quoc V. and Zhou, Denny},
 booktitle = {Advances in Neural Information Processing Systems},
 volume = {35},
 pages = {24824--24837},
 year = {2022}
}

@inproceedings{wuWebWalkerBenchmarkingLLMs2025,
 title = {WebWalker: Benchmarking LLMs in Web Traversal},
 author = {Wu, Jialong and Yin, Wenbiao and Jiang, Yong and Wang, Zhenglin and Xi, Zekun and Fang, Runnan and Zhang, Linhai and He, Yulan and Zhou, Deyu and Xie, Pengjun and Huang, Fei},
 booktitle = {Workshop on Reasoning and Planning for Large Language Models},
 year = {2025}
}

@inproceedings{wuAgenticReasoningStreamlined2025,
 title = {Agentic Reasoning: A Streamlined Framework for Enhancing LLM Reasoning with Agentic Tools},
 author = {Wu, Junde and Zhu, Jiayuan and Liu, Yuyuan and Xu, Min and Jin, Yueming},
 booktitle = {Proceedings of ACL 2025},
 pages = {28489--28503},
 year = {2025},
 doi = {10.18653/v1/2025.acl-long.1383}
}

@inproceedings{wuReFusionImprovingNatural2024,
 title = {ReFusion: Improving Natural Language Understanding with Computation-Efficient Retrieval Representation Fusion},
 author = {Wu, Shangyu and Xiong, Ying and Cui, Yufei and Liu, Xue and Tang, Buzhou and Kuo, Tei-Wei and Xue, Chun Jason},
 booktitle = {Proceedings of ICLR 2024},
 year = {2024}
}

@inproceedings{xuFaithfulLogicalReasoning2024,
 title = {Faithful Logical Reasoning via Symbolic Chain-of-Thought},
 author = {Xu, Jundong and Fei, Hao and Pan, Liangming and Liu, Qian and Lee, Mong-Li and Hsu, Wynne},
 booktitle = {Proceedings of ACL 2024},
 pages = {13326--13365},
 year = {2024},
 doi = {10.18653/v1/2024.acl-long.720}
}

@inproceedings{xuXAttentionBlockSparse2025,
 title = {XAttention: Block Sparse Attention with Antidiagonal Scoring},
 author = {Xu, Ruyi and Xiao, Guangxuan and Huang, Haofeng and Guo, Junxian and Han, Song},
 booktitle = {Proceedings of ICML 2025},
 year = {2025}
}

@inproceedings{yaoReActSynergizingReasoning2023,
 title = {ReAct: Synergizing Reasoning and Acting in Language Models},
 author = {Yao, Shunyu and Zhao, Jeffrey and Yu, Dian and Du, Nan and Shafran, Izhak and Narasimhan, Karthik R. and Cao, Yuan},
 booktitle = {Proceedings of ICLR 2023},
 year = {2023}
}

@inproceedings{yuVisRAGVisionbasedRetrievalaugmented2025,
 title = {VisRAG: Vision-Based Retrieval-Augmented Generation on Multi-Modality Documents},
 author = {Yu, Shi and Tang, Chaoyue and Xu, Bokai and Cui, Junbo and Ran, Junhao and Yan, Yukun and Liu, Zhenghao and Wang, Shuo and Han, Xu and Liu, Zhiyuan and Sun, Maosong},
 booktitle = {Proceedings of ICLR 2025},
 year = {2025}
}

@inproceedings{yuReClorReadingComprehension2020,
 title = {ReClor: A Reading Comprehension Dataset Requiring Logical Reasoning},
 author = {Yu, Weihao and Jiang, Zihang and Dong, Yanfei and Feng, Jiashi},
 booktitle = {Proceedings of ICLR 2020},
 year = {2020}
}

@inproceedings{yuanNativeSparseAttention2025,
 title = {Native Sparse Attention: Hardware-Aligned and Natively Trainable Sparse Attention},
 author = {Yuan, Jingyang and Gao, Huazuo and Dai, Damai and Luo, Junyu and Zhao, Liang and Zhang, Zhengyan and Xie, Zhenda and Wei, Yuxing and Wang, Lean and Xiao, Zhiping and Wang, Yuqing and Ruan, Chong and Zhang, Ming and Liang, Wenfeng and Zeng, Wangding},
 booktitle = {Proceedings of ACL 2025},
 pages = {23078--23097},
 year = {2025},
 doi = {10.18653/v1/2025.acl-long.1126}
}

@inproceedings{yuanRJudgeBenchmarkingSafety2024,
 title = {R-Judge: Benchmarking Safety Risk Awareness for LLM Agents},
 author = {Yuan, Tongxin and He, Zhiwei and Dong, Lingzhong and Wang, Yiming and Zhao, Ruijie and Xia, Tian and Xu, Lizhen and Zhou, Binglin and Li, Fangqi and Zhang, Zhuosheng and Wang, Rui and Liu, Gongshen},
 booktitle = {Findings of EMNLP 2024},
 pages = {1467--1490},
 year = {2024},
 doi = {10.18653/v1/2024.findings-emnlp.79}
}

@inproceedings{zhangEvaluatingImprovingToolAugmented2023,
 title = {Evaluating and Improving Tool-Augmented Computation-Intensive Math Reasoning},
 author = {Zhang, Beichen and Zhou, Kun and Wei, Xilin and Zhao, Xin and Sha, Jing and Wang, Shijin and Wen, Ji-Rong},
 booktitle = {Advances in Neural Information Processing Systems},
 volume = {36},
 pages = {23570--23589},
 year = {2023}
}

@article{zhangHallucinationMitigationRetrievalAugmented2025,
 title = {Hallucination Mitigation for Retrieval-Augmented Large Language Models: A Review},
 author = {Zhang, Wan and Zhang, Jing},
 journal = {Mathematics},
 volume = {13},
 number = {5},
 articleno = {856},
 year = {2025},
 doi = {10.3390/math13050856}
}

@article{zhangSirensSongAI2023,
 title = {Siren’s Song in the AI Ocean: A Survey on Hallucination in Large Language Models},
 author = {Zhang, Yue and Li, Yafu and Cui, Leyang and Cai, Deng and Liu, Lemao and Fu, Tingchen and Huang, Xinting and Zhao, Enbo and Zhang, Yu and Chen, Yulong and others},
 journal = {Computational Linguistics},
 pages = {1--46},
 year = {2025}
}

@inproceedings{zhaoSoKWatermarkingAIGenerated2025,
 title = {SoK: Watermarking for AI-Generated Content},
 author = {Zhao, Xuandong and Gunn, Sam and Christ, Miranda and Fairoze, Jaiden and Fabrega, Andres and Carlini, Nicholas and Garg, Sanjam and Hong, Sanghyun and Nasr, Milad and Tramer, Florian and Jha, Somesh and Li, Lei and Wang, Yu-Xiang and Song, Dawn},
 booktitle = {Proceedings of IEEE Symposium on Security and Privacy (SP 2025)},
 pages = {2621--2639},
 year = {2025},
 doi = {10.1109/SP61157.2025.00178}
}

@article{zhongComprehensiveSurveyAutomatic2023,
 title = {A Comprehensive Survey on Automatic Knowledge Graph Construction},
 author = {Zhong, Lingfeng and Wu, Jia and Li, Qian and Peng, Hao and Wu, Xindong},
 journal = {ACM Computing Surveys},
 volume = {56},
 number = {4},
 articleno = {94},
 year = {2023},
 doi = {10.1145/3618295}
}

@inproceedings{zhongMixofGranularityOptimizeChunking2025,
 title = {Mix-of-Granularity: Optimize the Chunking Granularity for Retrieval-Augmented Generation},
 author = {Zhong, Zijie and Liu, Hanwen and Cui, Xiaoya and Zhang, Xiaofan and Qin, Zengchang},
 booktitle = {Proceedings of COLING 2025},
 pages = {5756--5774},
 year = {2025}
}

@article{liu2024deepseek,
  title={Deepseek-v3 technical report},
  author={Liu, Aixin and Feng, Bei and Xue, Bing and Wang, Bingxuan and Wu, Bochao and Lu, Chengda and Zhao, Chenggang and Deng, Chengqi and Zhang, Chenyu and Ruan, Chong and others},
  journal={arXiv preprint arXiv:2412.19437},
  year={2024}
}

@article{Jiang2023Mistral7B,
  title = {Mistral 7B},
  author = {Jiang, Albert Q. and Sablayrolles, Alexandre and Mensch, Arthur and Bamford, Chris and Chaplot, Devendra Singh and de Las Casas, Diego and Bressand, Florian and Lengyel, Gianna and Lample, Guillaume and Saulnier, Lucile and Lavaud, Lélio Renard and Lachaux, Marie-Anne and Stock, Pierre and Le Scao, Teven and Lavril, Thibaut and Wang, Thomas and Lacroix, Timothée and El Sayed, William},
  journal = {CoRR},
  volume = {abs/2310.06825},
  year = {2023},
  doi = {10.48550/arXiv.2310.06825},
  url = {https://doi.org/10.48550/arXiv.2310.06825}
}

@article{team2023gemini,
  title={Gemini: a family of highly capable multimodal models},
  author={Team, Gemini and Anil, Rohan and Borgeaud, Sebastian and Alayrac, Jean-Baptiste and Yu, Jiahui and Soricut, Radu and Schalkwyk, Johan and Dai, Andrew M and Hauth, Anja and Millican, Katie and others},
  journal={arXiv preprint arXiv:2312.11805},
  year={2023}
}

@article{robertson2009probabilistic,
  title={The probabilistic relevance framework: BM25 and beyond},
  author={Robertson, Stephen and Zaragoza, Hugo and others},
  journal={Foundations and Trends{\textregistered} in Information Retrieval},
  volume={3},
  number={4},
  pages={333--389},
  year={2009},
  publisher={Now Publishers, Inc.}
}

@misc{touvronLLaMAOpenEfficient2023,
  title = {{{LLaMA}}: {{Open}} and {{Efficient Foundation Language Models}}},
  shorttitle = {{{LLaMA}}},
  author = {Touvron, Hugo and Lavril, Thibaut and Izacard, Gautier and Martinet, Xavier and Lachaux, Marie-Anne and Lacroix, Timoth{\'e}e and Rozi{\`e}re, Baptiste and Goyal, Naman and Hambro, Eric and Azhar, Faisal and Rodriguez, Aurelien and Joulin, Armand and Grave, Edouard and Lample, Guillaume},
  year = {2023},
  month = feb
}

@misc{openaiGPT4TechnicalReport2024,
  title = {{{GPT-4 Technical Report}}},
  author = {OpenAI and Achiam, Josh and Adler, Steven and Agarwal, Sandhini and Ahmad, Lama and Akkaya, Ilge and others},
  year = {2024},
  month = mar,
  number = {arXiv:2303.08774},
  eprint = {2303.08774},
  primaryclass = {cs},
  publisher = {arXiv},
  doi = {10.48550/arXiv.2303.08774},
  url = {http://arxiv.org/abs/2303.08774},
  urldate = {2025-07-25},
  archiveprefix = {arXiv}
}

@inproceedings{chanRQRAGLearningRefine2024,
title={{RQ}-{RAG}: Learning to Refine Queries for Retrieval Augmented Generation},
author={Chi-Min Chan and Chunpu Xu and Ruibin Yuan and Hongyin Luo and Wei Xue and Yike Guo and Jie Fu},
booktitle={First Conference on Language Modeling},
year={2024},
url={https://openreview.net/forum?id=tzE7VqsaJ4}
}

@inproceedings{niuRAGTruthHallucinationCorpus2024,
 title = {RAGTruth: A Hallucination Corpus for Developing Trustworthy Retrieval-Augmented Language Models},
 author = {Niu, Cheng and Wu, Yuanhao and Zhu, Juno and Xu, Siliang and Shum, KaShun and Zhong, Randy and Song, Juntong and Zhang, Tong},
 booktitle = {Proceedings of ACL 2024},
 pages = {10862--10878},
 year = {2024},
 doi = {10.18653/v1/2024.acl-long.585}
}

@article{kumar2024llmsurvey,
  title = {Large Language Models (LLMs): Survey, Technical Frameworks, and Future Challenges},
  author = {Kumar, Pranjal},
  journal = {Artificial Intelligence Review},
  volume = {57},
  year = {2024},
  doi = {10.1007/s10462-024-10888-y}
}

@inproceedings{chuNavigateEnigmaticLabyrinth2024,
 title = {Navigate through Enigmatic Labyrinth: A Survey of Chain of Thought Reasoning—Advances, Frontiers and Future},
 author = {Chu, Zheng and Chen, Jingchang and Chen, Qianglong and Yu, Weijiang and He, Tao and Wang, Haotian and Peng, Weihua and Liu, Ming and Qin, Bing and Liu, Ting},
 booktitle = {Proceedings of the 62nd Annual Meeting of the Association for Computational Linguistics (Volume 1: Long Papers)},
 year = {2024},
 pages = {1173--1203},
 doi = {10.18653/v1/2024.acl-long.65}
}

@inproceedings{xiongApproximateNearestNeighbor2020a,
title={Approximate Nearest Neighbor Negative Contrastive Learning for Dense Text Retrieval},
author={Lee Xiong and Chenyan Xiong and Ye Li and Kwok-Fung Tang and Jialin Liu and Paul N. Bennett and Junaid Ahmed and Arnold Overwijk},
booktitle={International Conference on Learning Representations},
year={2021},
url={https://openreview.net/forum?id=zeFrfgyZln}
}

@inproceedings{khattabColBERTEfficientEffective2020,
 title = {ColBERT: Efficient and Effective Passage Search via Contextualized Late Interaction over BERT},
 author = {Khattab, Omar and Zaharia, Matei},
 booktitle = {Proceedings of the 43rd International ACM SIGIR Conference on Research and Development in Information Retrieval},
 year = {2020},
 pages = {39--48},
 doi = {10.1145/3397271.3401075}
}

@article{farquhar2024detecting,
  title={Detecting hallucinations in large language models using semantic entropy},
  author={Farquhar, Sebastian and Kossen, Jannik and Kuhn, Lorenz and Gal, Yarin},
  journal={Nature},
  volume={630},
  number={8017},
  pages={625--630},
  year={2024},
  publisher={Nature Publishing Group UK London}
}

@inproceedings{watson2025there,
  title={Is There No Such Thing as a Bad Question? H4R: HalluciBot For Ratiocination, Rewriting, Ranking, and Routing},
  author={Watson, William and Cho, Nicole and Srishankar, Nishan},
  booktitle={Proceedings of the AAAI Conference on Artificial Intelligence},
  volume={39},
  number={24},
  pages={25470--25478},
  year={2025}
}

@inproceedings{qianExplicitQueryRewriting2022,
  title = {Explicit {{Query Rewriting}} for {{Conversational Dense Retrieval}}},
  booktitle = {Proceedings of the 2022 {{Conference}} on {{Empirical Methods}} in {{Natural Language Processing}}},
  author = {Qian, Hongjin and Dou, Zhicheng},
  year = {2022},
  pages = {4725--4737},
  publisher = {Association for Computational Linguistics},
  address = {Abu Dhabi, United Arab Emirates},
  doi = {10.18653/v1/2022.emnlp-main.311},
  urldate = {2025-10-27},
  langid = {english}
}

@article{liuLostMiddleHow2023,
  title = {Lost in the Middle: How Language Models Use Long Contexts},
  author = {Liu, Nelson F. and Lin, Kevin and Hewitt, John and Paranjape, Ashwin and Bevilacqua, Michele and Petroni, Fabio and Liang, Percy},
  journal = {Transactions of the Association for Computational Linguistics},
  volume = {12},
  pages = {157--173},
  year = {2024},
  doi = {10.1162/tacl_a_00638}
}

@inproceedings{changMAINRAGMultiAgentFiltering,
  title = {MAIN-RAG: Multi-Agent Filtering Retrieval-Augmented Generation},
  author = {Chang, Chia-Yuan and Jiang, Zhimeng and Rakesh, Vineeth and Pan, Menghai and Yeh, Chin-Chia Michael and Wang, Guanchu and Hu, Mingzhi and Xu, Zhichao and Zheng, Yan and Das, Mahashweta and Zou, Na},
  booktitle = {Proceedings of the 63rd Annual Meeting of the Association for Computational Linguistics (ACL 2025)},
  pages = {2607--2622},
  year = {2025},
  doi = {10.18653/v1/2025.acl-long.131}
}

@inproceedings{wassermanREALMMRAGRealWorldMultiModal2025,
  title = {REAL-MM-RAG: A Real-World Multi-Modal Retrieval Benchmark},
  author = {Wasserman, Navve and Pony, Roi and Naparstek, Oshri and Goldfarb, Adi Raz and Schwartz, Eli and Barzelay, Udi and Karlinsky, Leonid},
  booktitle = {Proceedings of the 63rd Annual Meeting of the Association for Computational Linguistics (ACL 2025)},
  pages = {31660--31683},
  year = {2025},
  doi = {10.18653/v1/2025.acl-long.1528}
}

@inproceedings{zhaoRetrievingMultimodalInformation2023,
  title = {Retrieving Multimodal Information for Augmented Generation: A Survey},
  author = {Zhao, Ruochen and Chen, Hailin and Wang, Weishi and Jiao, Fangkai and Do, Xuan Long and Qin, Chengwei and Ding, Bosheng and Guo, Xiaobao and Li, Minzhi and Li, Xingxuan and Joty, Shafiq},
  booktitle = {Findings of the Association for Computational Linguistics: EMNLP 2023},
  pages = {4736--4756},
  year = {2023},
  doi = {10.18653/v1/2023.findings-emnlp.314}
}

@inproceedings{chenMuRAGMultimodalRetrievalAugmented2022,
  title = {MuRAG: Multimodal Retrieval-Augmented Generator for Open Question Answering over Images and Text},
  author = {Chen, Wenhu and Hu, Hexiang and Chen, Xi and Verga, Pat and Cohen, William},
  booktitle = {Proceedings of the 2022 Conference on Empirical Methods in Natural Language Processing (EMNLP 2022)},
  pages = {5558--5570},
  year = {2022},
  doi = {10.18653/v1/2022.emnlp-main.375}
}

@article{nawaz2025review,
  title={A review of neuro-symbolic AI integrating reasoning and learning for advanced cognitive systems},
  author={Nawaz, Uzma and Anees-ur-Rahaman, Mufti and Saeed, Zubair},
  journal={Intelligent Systems with Applications},
  pages={200541},
  year={2025},
  publisher={Elsevier}
}

@inproceedings{deraedt2021neural,
  author    = {Luc De Raedt and Sebastijan Duman\v{c}i\'{c} and Robin Manhaeve and Giuseppe Marra},
  title     = {From Statistical Relational to Neural-Symbolic Artificial Intelligence},
  booktitle = {Proceedings of the Twenty-Ninth International Joint Conference on Artificial Intelligence},
  year      = {2021},
  pages     = {688--695}
}

@article{chenProgramThoughtsPrompting2023,
title={Program of Thoughts Prompting: Disentangling Computation from Reasoning for Numerical Reasoning Tasks},
author={Wenhu Chen and Xueguang Ma and Xinyi Wang and William W. Cohen},
journal={Transactions on Machine Learning Research},
issn={2835-8856},
year={2023},
url={https://openreview.net/forum?id=YfZ4ZPt8zd},
note={}
}

@inproceedings{arivazhaganHybridHierarchicalRetrieval2023,
  title = {Hybrid Hierarchical Retrieval for Open-Domain Question Answering},
  author = {Arivazhagan, Manoj Ghuhan and Liu, Lan and Qi, Peng and Chen, Xinchi and Wang, William Yang and Huang, Zhiheng},
  booktitle = {Findings of the Association for Computational Linguistics: ACL},
  year = {2023},
  pages = {10680--10689},
  doi = {10.18653/v1/2023.findings-acl.679}
}

@article{peng2025graph,
  title={Graph retrieval-augmented generation: A survey},
  author={Peng, Boci and Zhu, Yun and Liu, Yongchao and Bo, Xiaohe and Shi, Haizhou and Hong, Chuntao and Zhang, Yan and Tang, Siliang},
  journal={ACM Transactions on Information Systems},
  volume={44},
  number={2},
  pages={1--52},
  year={2025},
  publisher={ACM New York, NY}
}

@inproceedings{wu2026reflective,
  title={Reflective RAG: Self-Evaluation Driven Strategy Optimization in Agentic Retrieval-Augmented Generation},
  author={Wu, Haiyan and Wang, Chenchen and Sun, Chaoqun and Lu, Chengxiong and Chen, Yan-Hong and Zhang, Zhiqiang and Feng, Xiaoqing},
  booktitle={Findings of the Association for Computational Linguistics: ACL 2026},
  pages={13252--13267},
  year={2026}
}

@inproceedings{dong2025rag,
  title={Rag-critic: Leveraging automated critic-guided agentic workflow for retrieval augmented generation},
  author={Dong, Guanting and Jin, Jiajie and Li, Xiaoxi and Zhu, Yutao and Dou, Zhicheng and Wen, Ji-Rong},
  booktitle={Proceedings of the 63rd Annual Meeting of the Association for Computational Linguistics (Volume 1: Long Papers)},
  pages={3551--3578},
  year={2025}
}

@inproceedings{li2025survey,
  title={A Survey of RAG-Reasoning Systems in Large Language Models},
  author={Li, Yangning and Zhang, Weizhi and Yang, Yuyao and Huang, Wei-Chieh and Wu, Yaozu and Luo, Junyu and Bei, Yuanchen and Zou, Henry Peng and Luo, Xiao and Zhao, Yusheng and others},
  booktitle={Findings of the Association for Computational Linguistics: EMNLP 2025},
  pages={12120--12145},
  year={2025}
}

@inproceedings{liang2025reasoning,
  title={Reasoning rag via system 1 or system 2: A survey on reasoning agentic retrieval-augmented generation for industry challenges},
  author={Liang, Jintao and Lin, Huifeng and Wu, You and Zhao, Rui and Li, Ziyue and others},
  booktitle={Proceedings of the 14th International Joint Conference on Natural Language Processing and the 4th Conference of the Asia-Pacific Chapter of the Association for Computational Linguistics},
  pages={1954--1966},
  year={2025}
}

@inproceedings{li2025dora,
  title={DORA: Dynamic optimization prompt for continuous reflection of LLM-based agent},
  author={Li, Kun and Zhao, Tingzhang and Zhou, Wei and Hu, Songlin},
  booktitle={Proceedings of the 31st International Conference on Computational Linguistics},
  pages={7546--7557},
  year={2025}
}

@inproceedings{wu2024autogen,
  title={AutoGen: Enabling Next-Gen LLM Applications via Multi-Agent Conversation},
  author={Wu, Qingyun and Bansal, Gagan and Zhang, Jieyu and Wu, Yiran and Li, Beibin and Zhu, Erkang and Jiang, Li and Zhang, Xiaoyun and Zhang, Shaokun and Liu, Jiale and others},
  booktitle={ICLR 2024 Workshop on Large Language Model (LLM) Agents}
}

@inproceedings{hong2024metagpt,
  title={MetaGPT: Meta programming for a multi-agent collaborative framework},
  author={Hong, Sirui and Zhuge, Mingchen and Chen, Jonathan and Zheng, Xiawu and Cheng, Yuheng and Wang, Jinlin and Zhang, Ceyao and Yau, Steven and Lin, Zijuan and Zhou, Liyang and others},
  booktitle={International Conference on Learning Representations},
  volume={2024},
  pages={23247--23275},
  year={2024}
}

@article{shinn2023reflexion,
  title={Reflexion: Language agents with verbal reinforcement learning},
  author={Shinn, Noah and Cassano, Federico and Gopinath, Ashwin and Narasimhan, Karthik and Yao, Shunyu},
  journal={Advances in neural information processing systems},
  volume={36},
  pages={8634--8652},
  year={2023}
}

@article{madaan2023self,
  title={Self-refine: Iterative refinement with self-feedback},
  author={Madaan, Aman and Tandon, Niket and Gupta, Prakhar and Hallinan, Skyler and Gao, Luyu and Wiegreffe, Sarah and Alon, Uri and Dziri, Nouha and Prabhumoye, Shrimai and Yang, Yiming and others},
  journal={Advances in neural information processing systems},
  volume={36},
  pages={46534--46594},
  year={2023}
}

@inproceedings{li2025review,
  title={A review of prominent paradigms for llm-based agents: Tool use, planning (including rag), and feedback learning},
  author={Li, Xinzhe},
  booktitle={Proceedings of the 31st international conference on computational linguistics},
  pages={9760--9779},
  year={2025}
}

@incollection{conte2012rational,
  author    = {Conte, Rosaria},
  title     = {Rational, Goal-Oriented Agents},
  booktitle = {Encyclopedia of Complexity and Systems Science},
  editor    = {Meyers, Robert A.},
  publisher = {Springer},
  address   = {New York, NY},
  year      = {2012},
  doi       = {10.1007/978-1-4614-1800-9_158}
}

@article{martinelli2025taxonomy,
  author  = {Martinelli, Emanuele},
  title   = {A taxonomy of agents},
  journal = {Philosophical Explorations},
  volume  = {28},
  number  = {3},
  pages   = {285--302},
  year    = {2025},
  doi     = {10.1080/13869795.2025.2521296}
}

@inproceedings{ouyang2022training,
  title={Training language models to follow instructions with human feedback},
  author={Ouyang, Long and Wu, Jeff and Jiang, Xu and Almeida, Diogo and Wainwright, Carroll L and Mishkin, Pamela and Zhang, Chong and Agarwal, Sandhini and Slama, Katarina and Ray, Alex and others},
  booktitle={Proceedings of the 36th International Conference on Neural Information Processing Systems},
  pages={27730--27744},
  year={2022}
}

@misc{wei2026agentic,
  title={Agentic reasoning for large language models},
  author={Wei, Tianxin and Li, Ting-Wei and Liu, Zhining and Ning, Xuying and Yang, Ze and Zou, Jiaru and Zeng, Zhichen and Qiu, Ruizhong and Lin, Xiao and Fu, Dongqi and others},
  journal={arXiv preprint arXiv:2601.12538},
  year={2026}
}

@misc{zhang2025survey,
  title={A survey of graph retrieval-augmented generation for customized large language models},
  author={Zhang, Qinggang and Chen, Shengyuan and Bei, Yuanchen and Yuan, Zheng and Zhou, Huachi and Hong, Zijin and Chen, Hao and Xiao, Yilin and Zhou, Chuang and Dong, Junnan and others},
  journal={arXiv preprint arXiv:2501.13958},
  year={2025}
}

@misc{ke2025survey,
  title={A survey of frontiers in llm reasoning: Inference scaling, learning to reason, and agentic systems},
  author={Ke, Zixuan and Jiao, Fangkai and Ming, Yifei and Nguyen, Xuan-Phi and Xu, Austin and Long, Do Xuan and Li, Minzhi and Qin, Chengwei and Wang, Peifeng and Savarese, Silvio and others},
  journal={arXiv preprint arXiv:2504.09037},
  year={2025}
}

@misc{suresh2026agenticrag,
  title={AgenticRAG: Agentic Retrieval for Enterprise Knowledge Bases},
  author={Suresh, Susheel and Mak, Hazel and Chou, Shangpo and Kroon, Fred and Bhatnagar, Sahil},
  journal={arXiv preprint arXiv:2605.05538},
  year={2026}
}

@misc{singh2025agentic,
  title={Agentic retrieval-augmented generation: A survey on agentic rag},
  author={Singh, Aditi and Ehtesham, Abul and Kumar, Saket and Khoei, Tala Talaei and Vasilakos, Athanasios V},
  journal={arXiv preprint arXiv:2501.09136},
  year={2025}
}

@misc{sharma2025retrieval,
  title={Retrieval-augmented generation: A comprehensive survey of architectures, enhancements, and robustness frontiers},
  author={Sharma, Chaitanya},
  journal={arXiv preprint arXiv:2506.00054},
  year={2025}
}

@misc{linLLMbasedAgentsSuffer2025,
  title = {{{LLM-based Agents Suffer}} from {{Hallucinations}}: {{A Survey}} of {{Taxonomy}}, {{Methods}}, and {{Directions}}},
  shorttitle = {{{LLM-based Agents Suffer}} from {{Hallucinations}}},
  author = {Lin, Xixun and Ning, Yucheng and Zhang, Jingwen and Dong, Yan and Liu, Yilong and Wu, Yongxuan and Qi, Xiaohua and Sun, Nan and Shang, Yanmin and Cao, Pengfei and Zou, Lixin and Chen, Xu and Zhou, Chuan and Wu, Jia and Pan, Shirui and Wang, Bin and Cao, Yanan and Chen, Kai and Hu, Songlin and Guo, Li},
  year = {2025},
  month = sep,
  number = {arXiv:2509.18970},
  eprint = {2509.18970},
  primaryclass = {cs},
  publisher = {arXiv},
  doi = {10.48550/arXiv.2509.18970},
  urldate = {2025-10-18},
  archiveprefix = {arXiv}
}

@misc{bangHalluLensLLMHallucination2025,
  title = {{{HalluLens}}: {{LLM Hallucination Benchmark}}},
  shorttitle = {{{HalluLens}}},
  author = {Bang, Yejin and Ji, Ziwei and Schelten, Alan and Hartshorn, Anthony and Fowler, Tara and Zhang, Cheng and Cancedda, Nicola and Fung, Pascale},
  year = {2025},
  month = apr,
  number = {arXiv:2504.17550},
  eprint = {2504.17550},
  primaryclass = {cs},
  publisher = {arXiv},
  doi = {10.48550/arXiv.2504.17550},
  urldate = {2025-06-23},
  archiveprefix = {arXiv}
}

@misc{lazaridouInternetaugmentedLanguageModels2022,
  title = {Internet-Augmented Language Models through Few-Shot Prompting for Open-Domain Question Answering},
  author = {Lazaridou, Angeliki and Gribovskaya, Elena and Stokowiec, Wojciech and Grigorev, Nikolai},
  year = {2022},
  month = may,
  number = {arXiv:2203.05115},
  eprint = {2203.05115},
  primaryclass = {cs},
  publisher = {arXiv},
  doi = {10.48550/arXiv.2203.05115},
  urldate = {2025-07-21},
  archiveprefix = {arXiv}
}

@misc{nakanoWebGPTBrowserassistedQuestionanswering2022,
  title = {{{WebGPT}}: {{Browser-assisted}} Question-Answering with Human Feedback},
  shorttitle = {{{WebGPT}}},
  author = {Nakano, Reiichiro and Hilton, Jacob and Balaji, Suchir and Wu, Jeff and Ouyang, Long and Kim, Christina and Hesse, Christopher and Jain, Shantanu and Kosaraju, Vineet and Saunders, William and Jiang, Xu and Cobbe, Karl and Eloundou, Tyna and Krueger, Gretchen and Button, Kevin and Knight, Matthew and Chess, Benjamin and Schulman, John},
  year = {2022},
  month = jun,
  number = {arXiv:2112.09332},
  eprint = {2112.09332},
  primaryclass = {cs},
  publisher = {arXiv},
  doi = {10.48550/arXiv.2112.09332},
  urldate = {2025-05-21},
  archiveprefix = {arXiv},
  langid = {english}
}

@misc{xuNodeRAGStructuringGraphbased2025,
  title = {{{NodeRAG}}: {{Structuring Graph-based RAG}} with {{Heterogeneous Nodes}}},
  shorttitle = {{{NodeRAG}}},
  author = {Xu, Tianyang and Zheng, Haojie and Li, Chengze and Chen, Haoxiang and Liu, Yixin and Chen, Ruoxi and Sun, Lichao},
  year = {2025},
  month = apr,
  number = {arXiv:2504.11544},
  eprint = {2504.11544},
  primaryclass = {cs},
  publisher = {arXiv},
  doi = {10.48550/arXiv.2504.11544},
  urldate = {2025-10-24},
  archiveprefix = {arXiv}
}

@article{zhouTrustRAGEnhancingRobustness2025,
  title = {{{TrustRAG}}: {{Enhancing Robustness}} and {{Trustworthiness}} in {{RAG}}},
  author = {Zhou, Huichi and Lee, Kin-Hei and Zhan, Zhonghao and Chen, Yue and Li, Zhenhao and Wang, Zhaoyang and Haddadi, Hamed and Yilmaz, Emine},
  year = {2025},
  month = jan,
  eprint = {2501.00879},
  archiveprefix = {arXiv}
}

@misc{zhuangRankR1EnhancingReasoning2025,
  title = {Rank-{{R1}}: {{Enhancing Reasoning}} in {{LLM-based Document Rerankers}} via {{Reinforcement Learning}}},
  shorttitle = {Rank-{{R1}}},
  author = {Zhuang, Shengyao and Ma, Xueguang and Koopman, Bevan and Lin, Jimmy and Zuccon, Guido},
  year = {2025},
  month = mar,
  number = {arXiv:2503.06034},
  eprint = {2503.06034},
  primaryclass = {cs},
  publisher = {arXiv},
  doi = {10.48550/arXiv.2503.06034},
  urldate = {2025-07-29},
  archiveprefix = {arXiv}
}

@misc{liuLLM4RankingEasytouseFramework2025,
  title = {{{LLM4Ranking}}: {{An Easy-to-use Framework}} of {{Utilizing Large Language Models}} for {{Document Reranking}}},
  shorttitle = {{{LLM4Ranking}}},
  author = {Liu, Qi and Duan, Haozhe and Chen, Yiqun and Lu, Quanfeng and Sun, Weiwei and Mao, Jiaxin},
  year = {2025},
  month = apr,
  number = {arXiv:2504.07439},
  eprint = {2504.07439},
  primaryclass = {cs},
  publisher = {arXiv},
  doi = {10.48550/arXiv.2504.07439},
  urldate = {2025-07-28},
  archiveprefix = {arXiv}
}

@misc{baiHallucinationMultimodalLarge2025,
  title = {Hallucination of {{Multimodal Large Language Models}}: {{A Survey}}},
  shorttitle = {Hallucination of {{Multimodal Large Language Models}}},
  author = {Bai, Zechen and Wang, Pichao and Xiao, Tianjun and He, Tong and Han, Zongbo and Zhang, Zheng and Shou, Mike Zheng},
  year = {2025},
  month = apr,
  number = {arXiv:2404.18930},
  eprint = {2404.18930},
  primaryclass = {cs},
  publisher = {arXiv},
  doi = {10.48550/arXiv.2404.18930},
  urldate = {2025-08-20},
  archiveprefix = {arXiv}
}

@misc{zhaoSurveyLargeLanguage2025,
  title = {A {{Survey}} of {{Large Language Models}}},
  author = {Zhao, Wayne Xin and Zhou, Kun and Li, Junyi and Tang, Tianyi and Wang, Xiaolei and Hou, Yupeng and Min, Yingqian and Zhang, Beichen and Zhang, Junjie and Dong, Zican and Du, Yifan and Yang, Chen and Chen, Yushuo and Chen, Zhipeng and Jiang, Jinhao and Ren, Ruiyang and Li, Yifan and Tang, Xinyu and Liu, Zikang and Liu, Peiyu and Nie, Jian-Yun and Wen, Ji-Rong},
  year = {2025},
  month = mar,
  number = {arXiv:2303.18223},
  eprint = {2303.18223},
  primaryclass = {cs},
  publisher = {arXiv},
  doi = {10.48550/arXiv.2303.18223},
  urldate = {2025-05-30},
  archiveprefix = {arXiv}
}

@misc{edgeLocalGlobalGraph2025,
  title = {From {{Local}} to {{Global}}: {{A Graph RAG Approach}} to {{Query-Focused Summarization}}},
  shorttitle = {From {{Local}} to {{Global}}},
  author = {Edge, Darren and Trinh, Ha and Cheng, Newman and Bradley, Joshua and Chao, Alex and Mody, Apurva and Truitt, Steven and Metropolitansky, Dasha and Ness, Robert Osazuwa and Larson, Jonathan},
  year = {2025},
  month = feb,
  number = {arXiv:2404.16130},
  eprint = {2404.16130},
  primaryclass = {cs},
  publisher = {arXiv},
  doi = {10.48550/arXiv.2404.16130},
  urldate = {2025-07-24},
  archiveprefix = {arXiv}
}

@misc{zhengDeepResearcherScalingDeep2025,
  title = {{{DeepResearcher}}: {{Scaling Deep Research}} via {{Reinforcement Learning}} in {{Real-world Environments}}},
  shorttitle = {{{DeepResearcher}}},
  author = {Zheng, Yuxiang and Fu, Dayuan and Hu, Xiangkun and Cai, Xiaojie and Ye, Lyumanshan and Lu, Pengrui and Liu, Pengfei},
  year = {2025},
  month = apr,
  number = {arXiv:2504.03160},
  eprint = {2504.03160},
  primaryclass = {cs},
  publisher = {arXiv},
  doi = {10.48550/arXiv.2504.03160},
  url = {http://arxiv.org/abs/2504.03160},
  urldate = {2025-06-07},
  archiveprefix = {arXiv}
}

@misc{louAdaCoTParetoOptimalAdaptive2025,
  title = {{{AdaCoT}}: {{Pareto-Optimal Adaptive Chain-of-Thought Triggering}} via {{Reinforcement Learning}}},
  shorttitle = {{{AdaCoT}}},
  author = {Lou, Chenwei and Sun, Zewei and Liang, Xinnian and Qu, Meng and Shen, Wei and Wang, Wenqi and Li, Yuntao and Yang, Qingping and Wu, Shuangzhi},
  year = {2025},
  month = may,
  number = {arXiv:2505.11896},
  eprint = {2505.11896},
  primaryclass = {cs},
  publisher = {arXiv},
  doi = {10.48550/arXiv.2505.11896},
  url = {http://arxiv.org/abs/2505.11896},
  urldate = {2025-07-24},
  archiveprefix = {arXiv}
}

@misc{zhuChainofThoughtMattersImproving2025,
  title = {Chain-of-{{Thought Matters}}: {{Improving Long-Context Language Models}} with {{Reasoning Path Supervision}}},
  shorttitle = {Chain-of-{{Thought Matters}}},
  author = {Zhu, Dawei and Wei, Xiyu and Zhao, Guangxiang and Wu, Wenhao and Zou, Haosheng and Ran, Junfeng and Wang, Xun and Sun, Lin and Zhang, Xiangzheng and Li, Sujian},
  year = {2025},
  month = feb,
  number = {arXiv:2502.20790},
  eprint = {2502.20790},
  primaryclass = {cs},
  publisher = {arXiv},
  doi = {10.48550/arXiv.2502.20790},
  url = {http://arxiv.org/abs/2502.20790},
  urldate = {2025-07-24},
  archiveprefix = {arXiv}
}

@misc{caoComprehensiveSurveyAIGenerated2023,
  title = {A {{Comprehensive Survey}} of {{AI-Generated Content}} ({{AIGC}}): {{A History}} of {{Generative AI}} from {{GAN}} to {{ChatGPT}}},
  shorttitle = {A {{Comprehensive Survey}} of {{AI-Generated Content}} ({{AIGC}})},
  author = {Cao, Yihan and Li, Siyu and Liu, Yixin and Yan, Zhiling and Dai, Yutong and Yu, Philip S. and Sun, Lichao},
  year = {2023},
  month = mar,
  number = {arXiv:2303.04226},
  eprint = {2303.04226},
  primaryclass = {cs},
  publisher = {arXiv},
  doi = {10.48550/arXiv.2303.04226},
  url = {http://arxiv.org/abs/2303.04226},
  urldate = {2025-08-12},
  archiveprefix = {arXiv}
}

@misc{kimMedicalHallucinationsFoundation2025,
  title = {Medical {{Hallucinations}} in {{Foundation Models}} and {{Their Impact}} on {{Healthcare}}},
  author = {Kim, Yubin and Jeong, Hyewon and Chen, Shan and Li, Shuyue Stella and Lu, Mingyu and Alhamoud, Kumail and Mun, Jimin and Grau, Cristina and Jung, Minseok and Gameiro, Rodrigo and Fan, Lizhou and Park, Eugene and Lin, Tristan and Yoon, Joonsik and Yoon, Wonjin and Sap, Maarten and Tsvetkov, Yulia and Liang, Paul and Xu, Xuhai and Liu, Xin and McDuff, Daniel and Lee, Hyeonhoon and Park, Hae Won and Tulebaev, Samir and Breazeal, Cynthia},
  year = {2025},
  month = feb,
  number = {arXiv:2503.05777},
  eprint = {2503.05777},
  primaryclass = {cs},
  publisher = {arXiv},
  doi = {10.48550/arXiv.2503.05777},
  url = {http://arxiv.org/abs/2503.05777},
  urldate = {2025-08-11},
  archiveprefix = {arXiv}
}

@misc{jiangSurveyLargeLanguage2024,
  title = {A {{Survey}} on {{Large Language Model Hallucination}} via a {{Creativity Perspective}}},
  author = {Jiang, Xuhui and Tian, Yuxing and Hua, Fengrui and Xu, Chengjin and Wang, Yuanzhuo and Guo, Jian},
  year = {2024},
  month = feb,
  number = {arXiv:2402.06647},
  eprint = {2402.06647},
  primaryclass = {cs},
  publisher = {arXiv},
  doi = {10.48550/arXiv.2402.06647},
  url = {http://arxiv.org/abs/2402.06647},
  urldate = {2025-08-11},
  archiveprefix = {arXiv}
}

@misc{panditMedHalluComprehensiveBenchmark2025,
  title = {{{MedHallu}}: {{A Comprehensive Benchmark}} for {{Detecting Medical Hallucinations}} in {{Large Language Models}}},
  shorttitle = {{{MedHallu}}},
  author = {Pandit, Shrey and Xu, Jiawei and Hong, Junyuan and Wang, Zhangyang and Chen, Tianlong and Xu, Kaidi and Ding, Ying},
  year = {2025},
  month = feb,
  number = {arXiv:2502.14302},
  eprint = {2502.14302},
  primaryclass = {cs},
  publisher = {arXiv},
  doi = {10.48550/arXiv.2502.14302},
  url = {http://arxiv.org/abs/2502.14302},
  urldate = {2025-09-28},
  archiveprefix = {arXiv}
}

@misc{wangLMARSLegalMultiAgent2025,
  title = {L-{{MARS}}: {{Legal Multi-Agent Workflow}} with {{Orchestrated Reasoning}} and {{Agentic Search}}},
  shorttitle = {L-{{MARS}}},
  author = {Wang, Ziqi and Yuan, Boqin},
  year = {2025},
  month = sep,
  number = {arXiv:2509.00761},
  eprint = {2509.00761},
  primaryclass = {cs},
  publisher = {arXiv},
  doi = {10.48550/arXiv.2509.00761},
  url = {http://arxiv.org/abs/2509.00761},
  urldate = {2025-09-28},
  archiveprefix = {arXiv}
}

@misc{xiInfoDeepSeekBenchmarkingAgentic2025,
  title = {{{InfoDeepSeek}}: {{Benchmarking Agentic Information Seeking}} for {{Retrieval-Augmented Generation}}},
  shorttitle = {{{InfoDeepSeek}}},
  author = {Xi, Yunjia and Lin, Jianghao and Zhu, Menghui and Xiao, Yongzhao and Ou, Zhuoying and Liu, Jiaqi and Wan, Tong and Chen, Bo and Liu, Weiwen and Wang, Yasheng and Tang, Ruiming and Zhang, Weinan and Yu, Yong},
  year = {2025},
  month = may,
  number = {arXiv:2505.15872},
  eprint = {2505.15872},
  primaryclass = {cs},
  publisher = {arXiv},
  doi = {10.48550/arXiv.2505.15872},
  url = {http://arxiv.org/abs/2505.15872},
  urldate = {2025-09-28},
  archiveprefix = {arXiv}
}

@misc{nguyenMARAGMultiAgentRetrievalAugmented2025,
  title = {{{MA-RAG}}: {{Multi-Agent Retrieval-Augmented Generation}} via {{Collaborative Chain-of-Thought Reasoning}}},
  shorttitle = {{{MA-RAG}}},
  author = {Nguyen, Thang and Chin, Peter and Tai, Yu-Wing},
  year = {2025},
  month = may,
  number = {arXiv:2505.20096},
  eprint = {2505.20096},
  primaryclass = {cs},
  publisher = {arXiv},
  doi = {10.48550/arXiv.2505.20096},
  url = {http://arxiv.org/abs/2505.20096},
  urldate = {2025-07-27},
  archiveprefix = {arXiv}
}

@misc{phamAgentUniRAGTrainableOpenSource2025,
  title = {Agent-{{UniRAG}}: {{A Trainable Open-Source LLM Agent Framework}} for {{Unified Retrieval-Augmented Generation Systems}}},
  shorttitle = {Agent-{{UniRAG}}},
  author = {Pham, Hoang and Nguyen, Thuy-Duong and Bui, Khac-Hoai Nam},
  year = {2025},
  month = may,
  number = {arXiv:2505.22571},
  eprint = {2505.22571},
  primaryclass = {cs},
  publisher = {arXiv},
  doi = {10.48550/arXiv.2505.22571},
  url = {http://arxiv.org/abs/2505.22571},
  urldate = {2025-07-27},
  archiveprefix = {arXiv}
}
